\documentclass[10pt,twocolumn]{article}
\usepackage{amsmath,amssymb,mathtools,amsthm}
\usepackage[margin=0.75in]{geometry}
\usepackage{microtype}
\usepackage{graphicx}
\usepackage{subcaption}
\usepackage{booktabs}
\usepackage{multirow}
\usepackage{tabularx}
\usepackage[table]{xcolor}
\usepackage{placeins}
\usepackage{stfloats}
\usepackage{float}
\usepackage{enumitem}
\usepackage[T1]{fontenc}
\usepackage{newtxtext,newtxmath}
\usepackage{pifont}  

\usepackage[numbers,sort&compress]{natbib}

\usepackage[textsize=tiny]{todonotes}
\usepackage[
  colorlinks=true,
  linkcolor=cyan!60!blue,
  citecolor=cyan!60!blue,
  urlcolor=cyan!70!blue
]{hyperref}
\usepackage[capitalize,noabbrev]{cleveref}
\usepackage{fontawesome5}
\usepackage{pdfpages}
\theoremstyle{plain}

\theoremstyle{definition}

\theoremstyle{remark}

\newcommand{\cmark}{\textcolor{green!60!black}{\ding{51}}}
\newcommand{\xmark}{\textcolor{red!70!black}{\ding{55}}}
\title{\bf UniG2U-Bench: Do Unified Models Advance Multimodal Understanding?}

\author{
\begin{tabular}{c}
Zimo Wen$^{1,2}$ 
Boxiu Li$^{1,2}$ 
Wanbo Zhang$^{4}$ 
Junxiang Lei$^{4}$ 
Xiaoyu Chen$^{2}$ 
Yijia Fan$^{1}$ 
Qi Zhang$^{}$ \\
Yujiang Wang$^{5}$ 
Lili Qiu$^{1}$ 
Bo Li$^{3}$ 
Ziwei Liu$^{3}$ 
Caihua Shan$^{1}$
Yifan Yang$^{1}$ 
Yifei Shen$^{1*}$ \\
\\
$^{1}$Microsoft Research Asia \\
$^{2}$Shanghai Jiao Tong University \\
$^{3}$Nanyang Technological University \\
$^{4}$Fudan University \\
$^{5}$University of Oxford \\
\\
\large
\href{https://nssmd.github.io/unig2u.github.io/}
{\textbf{Project Page: https://nssmd.github.io/unig2u.github.io/}}
\end{tabular}
}

\date{} 

\begin{document}
\maketitle
\renewcommand{\thefootnote}{\fnsymbol{footnote}}
\footnotetext[1]{Corresponding author.}

% ...

\begin{abstract}
\textit{Unified multimodal models have recently demonstrated strong generative capabilities, yet whether and when generation improves understanding remains unclear. Existing benchmarks lack a systematic exploration of the specific tasks where generation facilitates understanding. To this end, we introduce UniG2U-Bench, a comprehensive benchmark categorizing generation-to-understanding (G2U) evaluation into 7 regimes and 30 subtasks, requiring varying degrees of implicit or explicit visual transformations. Extensive evaluation of over 30 models reveals three core findings:
1) Unified models generally underperform their base Vision-Language Models (VLMs), and Generate-then-Answer (GtA) inference typically degrades performance relative to direct inference.
2) Consistent enhancements emerge in spatial intelligence, visual illusions, or multi-round reasoning subtasks, where enhanced spatial and shape perception, as well as multi-step intermediate image states, prove beneficial.
3) Tasks with similar reasoning structures and models sharing architectures exhibit correlated behaviors, suggesting that generation-understanding coupling induces class-consistent inductive biases over tasks, pretraining data, and model architectures.
These findings highlight the necessity for more diverse training data and novel paradigms to fully unlock the potential of unified multimodal modeling.}
\end{abstract}

\section{Introduction}

Recent advances in multimodal foundation models have led to a growing interest in unifying understanding and generation within a single architecture~\cite{yang2025survey}. Unified models have demonstrated impressive performance in image synthesis, editing, and cross-modal generation, suggesting that strong perceptual and semantic understanding serves as a critical foundation for high-quality generation~\cite{zou2025unimmmu,xie2025mmeunify,li2025unieval,tri_marf_2026}.

Despite this progress, the interaction between understanding and generation has been studied predominantly in a single direction: understanding improves generation. Whether generation itself can enhance understanding by serving as an external medium for reasoning, verification, or hypothesis construction, remains largely unexplored. 

A key reason is the absence of benchmarks that explicitly evaluate understanding \emph{through} generation. Most existing benchmarks for multimodal understanding emphasize recognition, classification, or verbal reasoning over static visual inputs. As a result, strong performance can often be achieved by first converting images into dense textual descriptions and then applying a text-only large language model~\cite{xing2025caprl,qiao2024prism}. Consequently, many questions in popular understanding benchmarks can be answered without relying on visual information that is inherently non-linguistic. However, in many complex scenarios—such as drawing auxiliary constructions in geometry, reconstructing spatial layouts, or externalizing intermediate states in puzzles and games—effective understanding cannot be reduced to linguistic abstraction alone. In such cases, generation must serve as an internal or external reasoning mechanism that enables access to visual transformations not captured by textual descriptions.

The rapid evolution of unified multimodal models—such as Bagel, OmniGen, Show-o, and Janus—has proven that generation and understanding can coexist within a shared parameter space. While recent benchmarks like UniMMU~\cite{zou2025unimmmu} and MME-Unify~\cite{xie2025mmeunify} evaluate these models, they primarily assess understanding and generation in isolation (i.e., asking ``can it answer?'' and ``can it draw?'' independently). They fail to address whether the generative capacity actively \emph{aids} the understanding process. Furthermore, evaluating these models is challenging due to the confounding variables of model scaling and backbone differences. 

To systematically study this, we introduce \textbf{UniG2U}, which stands as the most comprehensive understanding benchmark to date for unified models—featuring the largest sample volume, the most diverse task categories, the widest model coverage, and the most complete evaluation framework. Unlike previous benchmarks, UniG2U strictly pairs cutting-edge unified models with their purely discriminative base VLMs under matched inference protocols. By doing so, we cleanly isolate the Generation-to-Understanding (G2U) capability, allowing us to accurately assess whether explicit or implicit visual generation facilitates multimodal reasoning.

\begin{figure}
    \centering
    \includegraphics[width=1\linewidth]{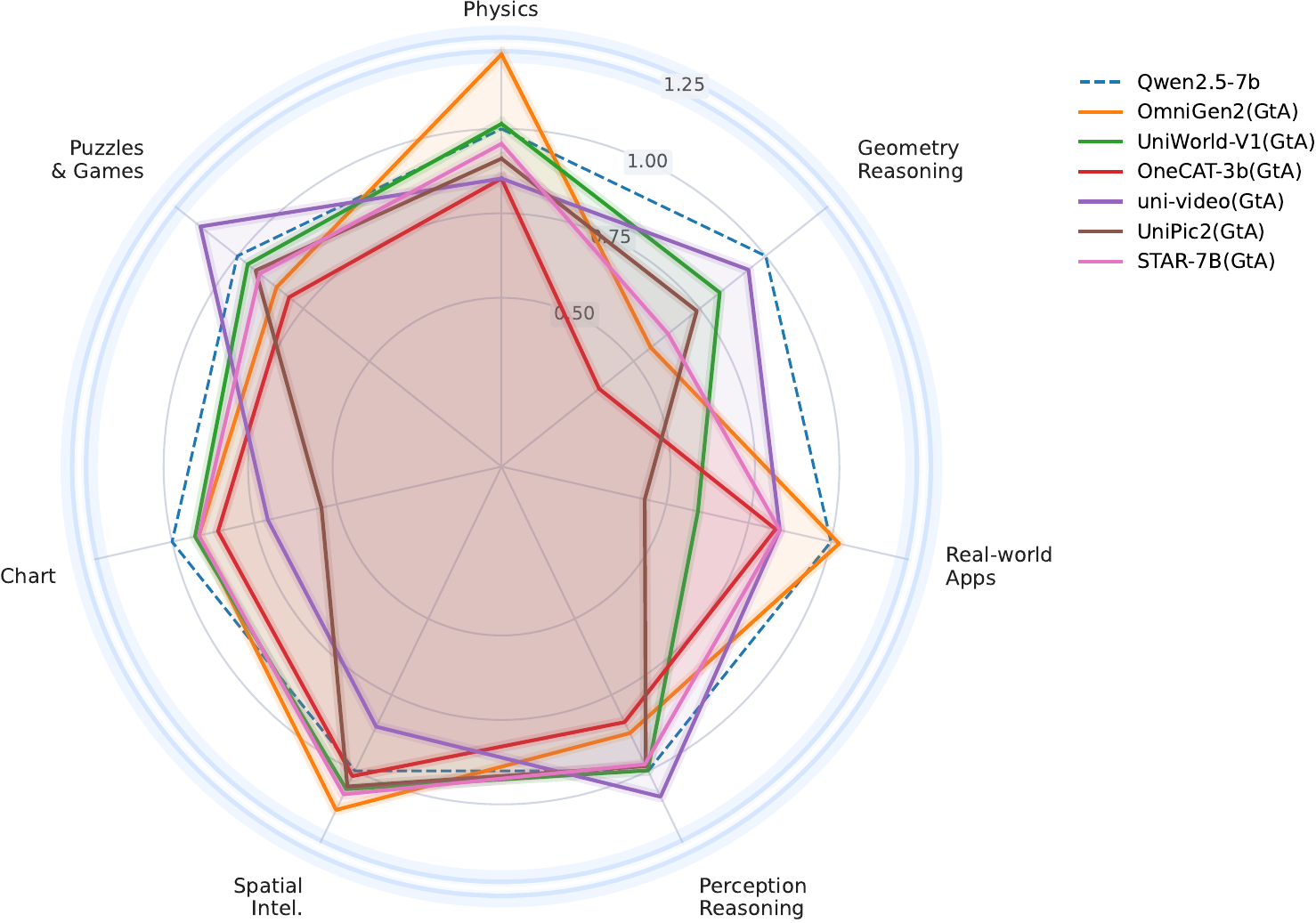}
    \caption{Model Performance Radar Chart}
    \label{fig:radar}
\end{figure}
We benchmarked over 30 models—encompassing base visual understanding models (VLMs), unified models, and agentic models (which we also define as generalized unified models)—across seven reasoning regimes. Each unified model is evaluated under two distinct inference modes: Direct and Generate-then-Answer (GtA). By rigorously comparing unified models with their base VLMs, we observe a nuanced Generation-to-Understanding (G2U) dynamic, characterized by varying degrees of performance improvements and degradations across different task categories, as vividly illustrated by the relative performance radar chart for a representative subset of the evaluated models(see Fig. \ref{fig:radar}). To deeply investigate this dynamic, we formulate four key research questions (RQs) and design corresponding experiments. Furthermore, we introduce two novel metrics—Reasoning-Alignment (RA) and Answer-Alignment (AL)—to quantitatively measure the quality of intermediate images generated during the GtA stage. Through this comprehensive evaluation, we distill our observations into three core findings:

First, we note an overall performance degradation. Integrating generative capabilities does not yield universal gains; rather, unified models generally underperform their corresponding base VLMs on standard understanding tasks. Furthermore, the GtA paradigm—generating intermediate images prior to answering—typically degrades performance relative to direct inference, often propagating visual errors in structurally constrained domains.

Second, despite this general decline, structured improvements consistently emerge. In specific spatial, illusion-sensitive, and multi-round reasoning subtasks, externalizing visual transformations transitions into a significant advantage. In these transformation-intensive scenarios, enhanced spatial and shape perception, as well as multi-step intermediate image states, prove highly beneficial.

Third, we identify a structured task-model correlation. At the task level, G2U gains are not random but cluster by cognitive demands: perception-oriented tasks correlate strongly with one another, as do logic- and reasoning-intensive tasks. At the model level, unified models built upon the same base model demonstrate strong behavioral correlations, whereas those sharing only architectural similarities exhibit much weaker correlations. These highly predictable trends suggest that generation-understanding coupling induces class-consistent inductive biases, which are fundamentally shaped by the underlying pretraining data and, to a lesser extent, the model architectures.

Our main contributions are summarized as follows:
\begin{enumerate}
    \item \textbf{A Novel Testbed:} We introduce UniG2U, the largest unified-model evaluation testbed to date designed specifically for the Generation-to-Understanding (G2U) paradigm. It covers 3,000 carefully curated instances across seven reasoning categories, providing standardized data, protocols, and scripts for reproducible evaluation.
    
    \item \textbf{Comprehensive Experiments:} We conduct the largest large-scale study of unified models so far, benchmarking over 30 models across autoregressive, diffusion-based, and hybrid architectures. We systematically isolate G2U gains by strictly comparing unified models against their foundational VLM baselines.
    
    \item \textbf{Deep Mechanistic Insights:} Our analyses demystify when generation helps or harms understanding. We expose the inherent architectural trade-offs between generative and discriminative objectives, the vulnerabilities of the Generate-then-Answer  visual reasoning paradigm, and the critical role of intermediate alignment fidelity.
\end{enumerate}
\section{Related Works}

\subsection{Unified Multimodal Models}
Early multimodal systems largely followed a \emph{disjoint} design, coupling a vision--language understanding model with a separate image generator in a pipeline, which made end-to-end credit assignment across modalities difficult \citep{alayrac2022flamingo,li2023blip2,danet_2025,past_2025}.
The rise of large multimodal LMs strengthened the trend toward a \emph{single interaction interface}---using language to control multimodal perception and reasoning---yet image generation often remained externalized as a separate module \citep{huang2023kosmos1}.
In parallel, diffusion-style text-to-image synthesis made ``understanding+generation'' pipelines practical at scale \citep{rombach2022ldm}.
More recently, research has pushed toward \emph{unified multimodal models} that can both \emph{understand} and \emph{generate} within one interface, motivated by shared representation learning and tighter coupling between perception and synthesis \citep{team2024chameleon,xie2024showo}.
Token-based early-fusion models suggest that a single next-token objective over mixed-modal sequences can support both behaviors \citep{team2024chameleon}, while hybrid formulations combine autoregressive decoding with discrete diffusion/flow-style generation inside a single transformer to broaden modality patterns \citep{xie2024showo}.
Scaling these designs toward broader modality coverage (e.g., image/video) has made unified interfaces increasingly competitive as general-purpose multimodal backbones \citep{xie2025showo2,deng2025bagel,chen2025januspro,ipr1_2025}.
Surveys have begun to systematize this design space and repeatedly emphasize that \emph{evaluation} remains a key bottleneck for understanding when unification yields \emph{genuine} capability gains \citep{yang2025survey,zhao2025unifiedsurvey}.

\subsection{Benchmarks for Unified Models}
Benchmarks have begun to operationalize ``unification'' in different ways, including capability coverage, mixed-modality output competence, and explicit bidirectional synergy protocols \citep{xie2025mmeunify,shi2025realunify,zou2025unimmmu,li2026ueval,tirbench_2025}.
MME-Unify (MME-U) provides a scalable audit via three domains (understanding, generation, and mixed/unified tasks) and makes unified tasks \emph{quantifiable} by casting them into \emph{text-option + image-option} selection, judging generated images through CLIP-style similarity \citep{xie2025mmeunify,radford2021clip}.
RealUnify defines unification as \emph{bidirectional synergy} (UEG/GEU) and introduces paired \emph{direct} and \emph{stepwise} protocols to localize whether failures stem from core skills or integration \citep{shi2025realunify}.
Uni-MMMU centers multi-discipline reasoning with Gen$\rightarrow$Und / Und$\rightarrow$Gen paradigms and emphasizes verifiable intermediates with dual-channel (text+image) scoring \citep{zou2025unimmmu,yue2023mmmu}.
ROVER targets reciprocal cross-modal reasoning, introducing a specific split (ROVER-TG) to evaluate visually-augmented verbal generation, where models produce intermediate visual rationales before answering\citep{liang2025rover}.
UEval evaluates real-world requests requiring both text and images with item-level rubrics and a runnable pipeline/leaderboard \citep{li2026ueval,li2025unieval}.

Despite these advances, existing benchmarks still fall short of diagnosing \emph{generation-to-understanding} (G2U) as a capability phenomenon: (i) what constitutes a \emph{G2U task} where intermediate visuals are intrinsically useful under matched budgets, and (ii) how results vary across fine-grained \emph{task regimes} and \emph{model archetypes}.
MME-U offers scale but does not isolate synergy or define G2U task structure \citep{xie2025mmeunify}. RealUnify and Uni-MMMU introduce synergy-aware protocols but cover a narrower slice of task types and unified-model designs \citep{shi2025realunify,zou2025unimmmu}. While ROVER explores interleaved visual reasoning, its G2U evaluation is limited in scale (404 tasks across 6 subtasks) and does not explicitly pair unified models with their base VLMs to strictly isolate the synergistic gains. UEval focuses on mixed-output quality rather than causal G2U effects \citep{li2026ueval}.

To bridge this critical gap, we introduce \textbf{UniG2U}. As explicitly compared in Table \ref{tab:benchmark_compare}, UniG2U stands as the most comprehensive benchmark to date for evaluating unified models. It features the largest sample volume (3,000 understanding tasks), the most diverse taxonomy (30 fine-grained subtasks across 7 reasoning regimes), and the widest model coverage (evaluating over 30 models). More importantly, UniG2U is the first benchmark to systematically and quantitatively analyze the synergistic effect of generation on understanding. By curating G2U-centric regimes (e.g., spatial manipulation, state externalization, visual illusions) and strictly pairing unified models with their purely discriminative base VLMs, we cleanly isolate the G2U capability. Through this unprecedented scale and rigorous framework, UniG2U provides key mechanistic insights into the generation-understanding coupling, revealing class-consistent inductive biases and identifying specific scenarios where generation effectively serves multimodal reasoning.

\begin{table}[ht]
\centering
\caption{Comparison of unified multimodal benchmarks.}
\label{tab:benchmark_compare}
\footnotesize
\setlength{\tabcolsep}{5pt}
\begin{tabular}{lccc}
\toprule
Benchmark & \# Understanding Samples & \# Subtasks & Synergy \\
\midrule
MME-Unify & 1964 & 3 & \xmark \\
Uni-MMMU & 524 & 4 & \cmark \\
ROVER & 404 & 6 & \cmark \\
RealUnify & 400 & 4 & \cmark \\
\textbf{UniG2U (ours)} & \textbf{3000} & \textbf{30} & \textbf{\cmark} \\
\bottomrule
\end{tabular}
\end{table}


\section{Formulations and Definitions}
\label{sec:formulation}

This section formalizes the core objects studied in UniG2U.
We (i) define \emph{unified multimodal models} (UMMs) and the capability-level notion of \emph{Generation Helps Understanding} (G2U),
(ii) introduce a taxonomy of UMMs used throughout the paper, and
(iii) distinguish \textbf{Direct} vs.\textbf{ Generate-then-Answer (GtA)}  inference under a unified evaluation.
\subsection{Definitions: UMMs and G2U}
\label{sec:unified_model_def}

\paragraph{Task setting.}
Each instance is formulated as a tuple $x=(V,T,Q)$, where $V$ denotes the input visual context (e.g., images), $T$ represents the text instructions or context, and $Q$ is the specific query to be answered. The target answer is denoted as $y\in\mathcal{Y}$.

\paragraph{Pure understanding models.}
A \emph{pure understanding} model maps the given input directly to an answer,
\begin{equation}
\hat{y}=M_U(x), \qquad M_U\in\mathcal{M}_U,
\end{equation}
and does not expose an interface for producing intermediate \emph{visual} artifacts intended to be re-consumed for the same instance.

\paragraph{Unified multimodal models (UMMs).}
A \emph{unified multimodal model} ($M_{UM}$) supports multimodal understanding together with a generative interface that can produce an intermediate visual artifact $G$ (e.g., image, sketch, annotation, reconstruction, edit) conditioned on the multimodal input $x$. Conceptually, UMMs are motivated by the hypothesis that generation can (i) shape internal multimodal representations through unified training, and/or (ii) act as a controllable reasoning resource via these intermediate artifacts.

\paragraph{Defining ``Generation Helps Understanding'' (G2U).}
\label{sec:g2u_def}
Let $\mathcal{A}(\cdot,\cdot)$ be the task evaluation metric. We define the performance of a model $M$ on a dataset $\mathcal{D}$ as:
\begin{equation}
\mathrm{Perf}(M;\mathcal{D})
= \mathbb{E}_{(x,y)\sim\mathcal{D}}\big[\mathcal{A}(M(x),y)\big].
\end{equation}
To rigorously isolate the effect of generative capabilities, we must mitigate confounders stemming from different architectural designs or pretraining data. Therefore, for a given unified system $M_{UM}$, let $\mathcal{B}(M_{UM}) \in \mathcal{M}_U$ denote its corresponding \emph{purely discriminative base VLM} (i.e., the foundational understanding model upon which the unified model is built, evaluated under a matched computation budget). 

We define the \textbf{G2U gain} for the unified system $M_{UM}$ as:
\begin{equation}
\Delta_{\mathrm{G2U}}(M_{UM};\mathcal{D})
= \mathrm{Perf}(M_{UM};\mathcal{D}) - \mathrm{Perf}(\mathcal{B}(M_{UM});\mathcal{D}).
\label{eq:g2u_gain}
\end{equation}
By enforcing strict \textbf{capability comparability} (base-unified pairing) and \textbf{budget matching}, we ensure that any observed $\Delta_{\mathrm{G2U}}$ intrinsically reflects the synergistic effect of generation-to-understanding, rather than mere variations in base model scale or empirical noise.

\subsection{Taxonomy of Unified Multimodal Models}
\label{sec:umm_taxonomy}

We group evaluated UMMs into three categories based on their architectural designs and training paradigms: \textbf{end-to-end unified models} (E2E), \textbf{decoupled unified systems} (Decoupled), and \textbf{agentic unified models} (UM-Ag).

\begin{figure*}[ht]
  \centering
  \includegraphics[width=0.8\linewidth]{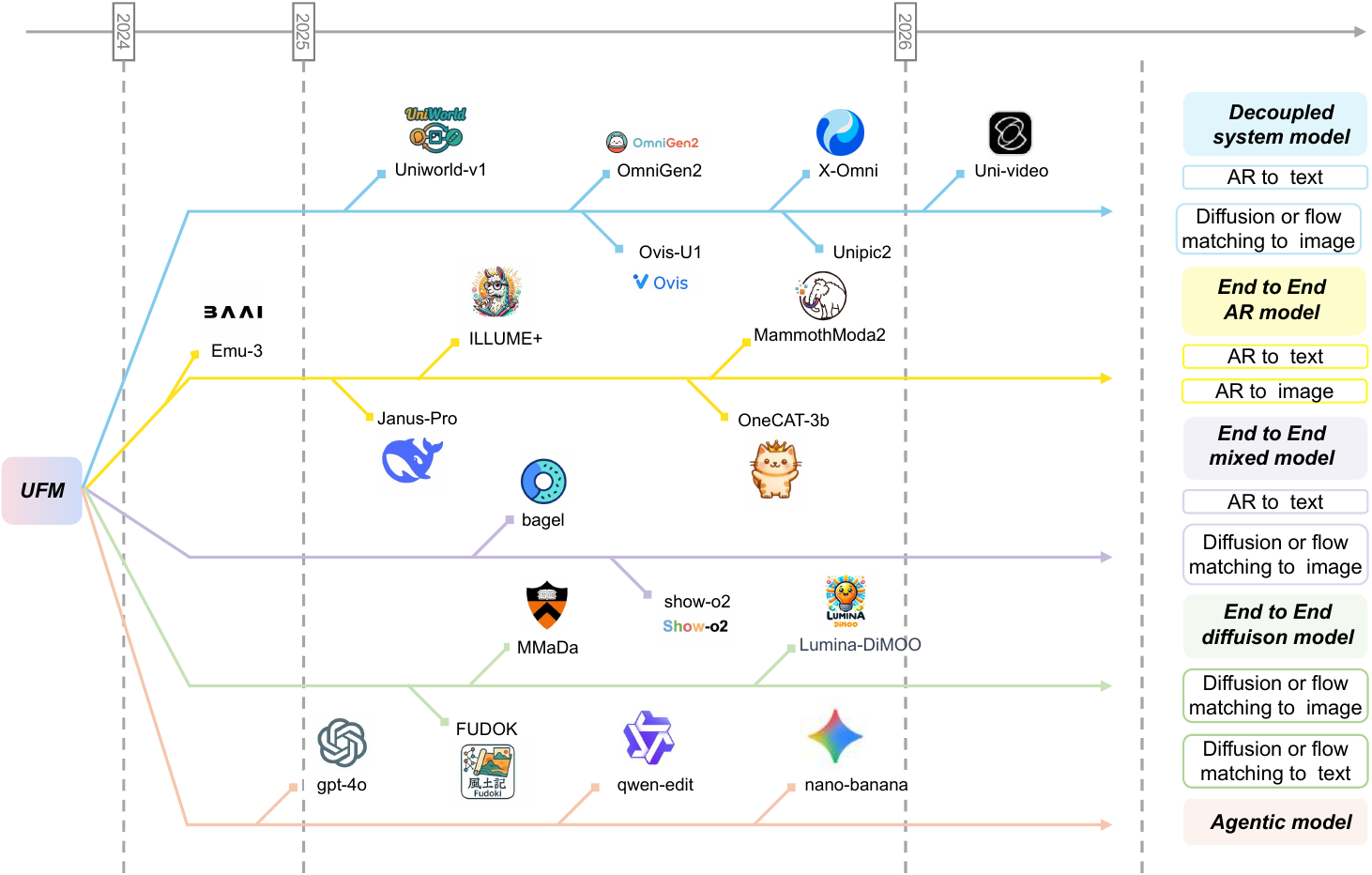}
  \caption{Taxonomy of unified multimodal models (UMMs). All models annotated in the figure are benchmarked in this work.}
  \label{fig:umm-taxonomy}
\end{figure*}

All categories can be conceptually expressed via a shared factorization:
\begin{equation}
G=\mathrm{Gen}(x), \qquad \hat{y}=\mathrm{Ans}(x,G),
\label{eq:unified_factorization}
\end{equation}
where $G=\varnothing$ recovers direct answering without intermediate visual artifacts. The taxonomy fundamentally differs in how $\mathrm{Gen}$ and $\mathrm{Ans}$ are implemented and trained.

\paragraph{(1) End-to-end unified models (E2E).}
E2E models are \emph{parameter-coupled}: $\mathrm{Gen}$ and $\mathrm{Ans}$ are realized by a single model with shared parameters $\theta$. Crucially, the generation and understanding capabilities are \textbf{jointly trained} (e.g., via interleaved multimodal data) within a unified parameter space. A defining property is that both capabilities emerge from the end-to-end training of one unified architecture, rather than from separated components. Representative instances in our study include \emph{Bagel} and \emph{UniPic2-Metaquery-9B}.

\paragraph{(2) Decoupled unified systems (Decoupled).}
Decoupled systems are \emph{module-separated}. In contrast to E2E models, the generation and understanding components are \textbf{not jointly trained}. Instead, $\mathrm{Gen}$ and $\mathrm{Ans}$ are implemented by independent modules (with parameters $\phi$ and $\psi$) that are stitched together through modular orchestration. Concretely, a generation module produces an explicit intermediate artifact $G$, which is subsequently consumed by a frozen answering module. This modularity enables flexible component-level splicing. Typical examples include \emph{Bagel two-stage} and \emph{OneCAT}.

\paragraph{(3) Agentic unified models (UM-Ag).}
In agentic systems, generation and understanding are performed by \textbf{completely different models}. They achieve unification merely at the protocol level (via a control policy or bounded tool-use interaction) rather than through architectural integration or joint training. For instance, a pure language model might invoke an external, black-box image generation API. In our study, tool-driven visual exploration settings such as \emph{GPT-4o + image tool} instantiate this category.

Conceptually, we regard UM-Ag as a generalized variant of UMMs. However, in our experimental analysis, we explicitly distinguish agentic models from standard unified models (E2E and Decoupled). This separation is necessitated by the fact that most agentic systems consist of distinct, often closed-source proprietary models that rely on external tool-use rather than undergoing any form of multimodal unified training. Consequently, they inherently lack a strictly matched base VLM, rendering them incompatible with the rigorous base-unified comparative framework defined in Section \ref{sec:unified_model_def} for calculating exact G2U gains.

Fig.~\ref{fig:umm-taxonomy} summarizes this taxonomy and highlights how the three classes differ in their degree of unification—ranging from jointly trained parameters (E2E) to stitched modules (Decoupled), and finally to tool-invoked separate models (UM-Ag).

\subsection{Direct vs.\ Generate-then-Answer (GtA) Inference}
\label{sec:direct_vs_gta}

The Generation-to-Understanding (G2U) paradigm can be instantiated through two distinct inference protocols: Direct and Generate-then-Answer (GtA).
In the \emph{Direct} setting, the system answers the query without producing any external intermediate visual artifacts, formalized as $\hat{y}=\mathrm{Ans}(x,\varnothing)$.
In the \emph{GtA} setting, the system explicitly generates an intermediate visual rationale $G$ and re-consumes it to derive the final answer, as defined in \eqref{eq:unified_factorization} (or via an agentic rollout with multiple $G_t$).

\paragraph{Decomposing G2U gains.}
Define the best GtA performance for model $M$ as:
\begin{equation}
\begin{aligned}
\mathrm{Perf}_{\mathrm{GtA}}(M;\mathcal{D})
\triangleq
\max\Big\{
&\mathrm{Perf}_{\mathrm{1shot\_GtA}}(M;\mathcal{D}), \\
&\mathrm{Perf}_{\mathrm{agentic\_GtA}}(M;\mathcal{D})
\Big\},
\end{aligned}
\label{eq:perf_gta_def}
\end{equation}
where $\mathrm{Perf}_{\mathrm{1shot\_GtA}}$ instantiates \eqref{eq:unified_factorization} with a single generated artifact $G$, and $\mathrm{Perf}_{\mathrm{agentic\_GtA}}$ instantiates it with a closed-loop rollout.

We can then decompose the overall G2U performance shift into two specific components:
\begin{equation}
\begin{aligned}
\Delta_{\mathrm{Direct}}
&= \mathrm{Perf}_{\mathrm{Direct}}(M_{UM};\mathcal{D})-\mathrm{Perf}(\mathcal{B}(M_{UM});\mathcal{D}), \\
\Delta_{\mathrm{GtA}}
&= \mathrm{Perf}_{\mathrm{GtA}}(M_{UM};\mathcal{D})-\mathrm{Perf}_{\mathrm{Direct}}(M_{UM};\mathcal{D}).
\end{aligned}
\label{eq:g2u_decomp}
\end{equation}
Here, $\Delta_{\mathrm{Direct}}$ captures the intrinsic capability shift resulting from unified multimodal training (without explicit generation), while $\Delta_{\mathrm{GtA}}$ isolates the specific performance impact of explicitly externalizing the visual reasoning process prior to answering.

While externalizing visual transformations can be beneficial, the GtA protocol can also degrade performance if the intermediate generated artifacts are inaccurate, physically implausible, or misaligned with the intended reasoning plan, thereby misleading the subsequent answering module.

\section{The Setup of UniG2U}
\subsection{Dataset Construction and Task Taxonomy}

\begin{figure*}[tbp]
  \centering
  \includegraphics[width=\linewidth]{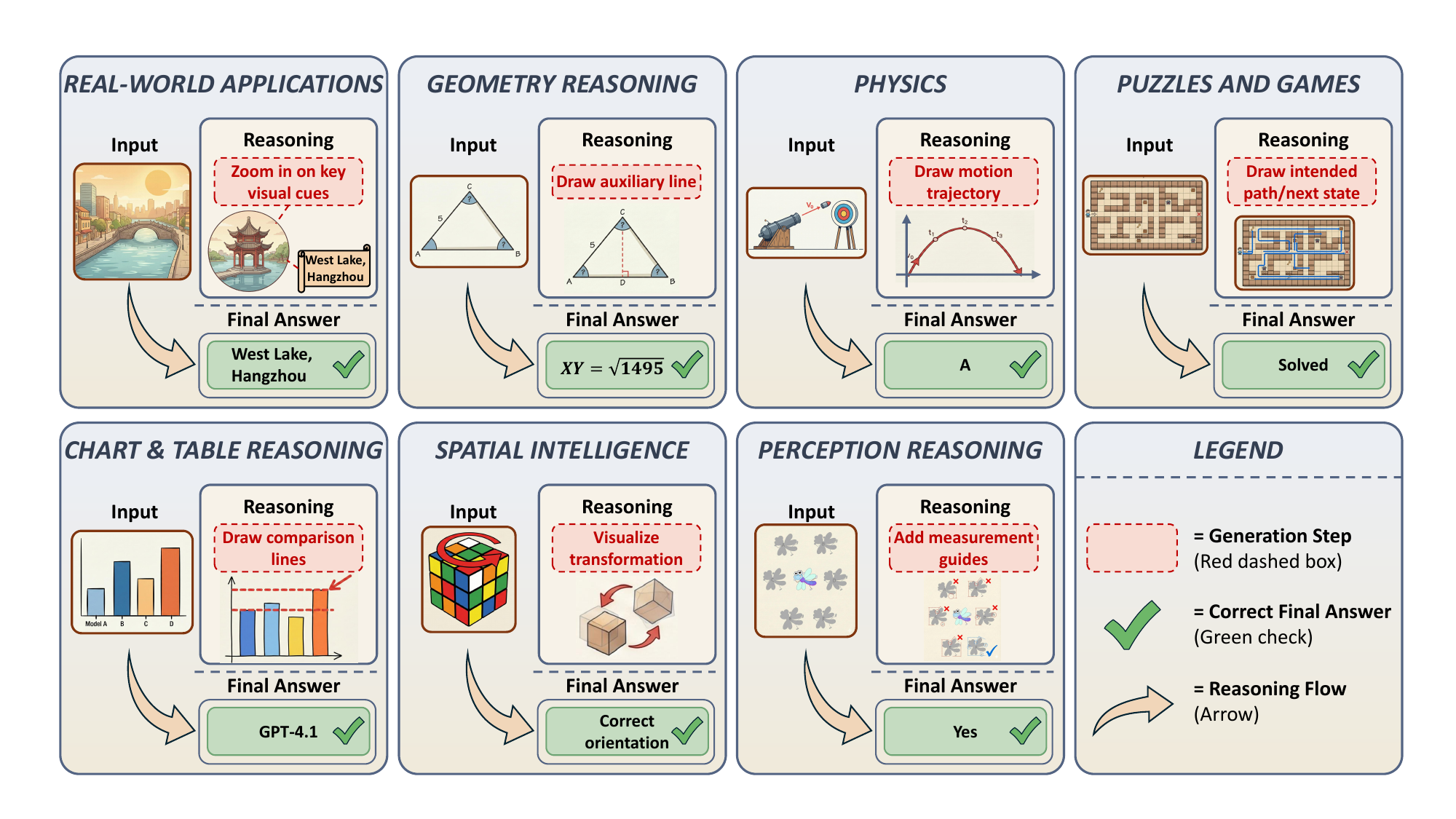}
  \caption{Task taxonomy overview of UniG2U.}
  \label{fig:task-taxonomy}
\end{figure*}

UniG2U is designed as a diagnostic benchmark to study when and why generation benefits multimodal understanding. Rather than focusing on a single task family, we construct a diverse dataset covering a wide spectrum of perceptual and cognitive structures. In total, UniG2U contains approximately \textbf{3,000 samples}, organized into \textbf{7 high-level categories} and \textbf{30 fine-grained subcategories}. This section details the taxonomy design and the motivation behind each category.

\subsubsection{Design Principles}
The dataset taxonomy is guided by two core principles, fundamentally echoing Richard Feynman's profound adage: \emph{``What I cannot create, I do not understand.''} 

First, categories should reflect \emph{distinct underlying cognitive structures} rather than superficial differences in dataset sources. 

Second, all selected categories must theoretically exhibit the potential for Generation-to-Understanding (G2U) synergy. Rather than aggregating arbitrary multimodal tasks, we intentionally curate scenarios where visual externalization is intrinsically coupled with comprehension. Whether through drawing geometric auxiliary lines, simulating physical dynamics, tracking spatial transformations, or even reconstructing fine-grained perceptual features.

\begin{enumerate}
    \item \textbf{Real-world Applications (200 samples).} Visually grounded tasks resembling practical applications, such as attentional focusing and visual shortest-path reasoning (VSP). These tasks typically require identifying relevant regions, tracking spatial relations, or following visual constraints across complex scenes. Data are primarily sourced from VSP-style benchmarks and attentional reasoning datasets.

    \item \textbf{Geometry Reasoning (200 samples).} Visual mathematics and geometry problems, including both 2D and 3D geometric reasoning. Subtasks include planar and solid geometry, curated from datasets such as Geometry3K and AuxSolidMath. These problems often benefit from structured intermediates (e.g., drawing auxiliary lines, reconstructing geometric structure), making them ideal for evaluating explicit generation-augmented understanding.

    \item \textbf{Physics Reasoning (200 samples).} Known-answer scientific reasoning grounded in diagrams, including physics and optics problems. We include curated subsets from datasets such as Cloudriver and PhyX, where solving the problem requires interpreting diagrams, understanding physical processes, or reasoning over scientific illustrations.

    \item \textbf{Puzzles and Games (537 samples).} Structured reasoning with state tracking and rule-based transformations, including mental reconstruction/tracking, visual tracking, and classical puzzles such as maze navigation, jigsaw puzzles, and sliding puzzles. Data are drawn from multiple sources including Uni-MMMU, RealUnify, and BabyVision-style datasets. These tasks often admit explicit intermediate visual states, making them suitable for studying both implicit and explicit generation mechanisms.

    \item \textbf{Chart \& Table Reasoning (100 samples).} Reasoning over structured visual data (charts/tables). We use a curated subset of ChartQA, where models must interpret bar charts, line plots, and other visualized representations; intermediate generation may help by re-expressing or simplifying the chart.

    \item \textbf{Spatial Intelligence (500 samples).} Reasoning about spatial relations, geometric transformations, and object dynamics (e.g., multi-step spatial reasoning, attribute measurement/approximation, camera/object motion). Samples are drawn primarily from spatial reasoning datasets such as MMSI-Bench.

    \item \textbf{Perception Reasoning (1263 samples).} Fine-grained visual perception and low-level reasoning, including scene/shape-based reasoning, in-scene vs. in-shape discrimination, logo understanding, algorithmic and deductive visual reasoning, spatial pattern recognition, and analogical/inductive reasoning. Data are curated from IllusionBench, visual puzzle datasets, and BabyVision-style benchmarks; compared to other categories, many tasks here are primarily perceptual and may benefit less from explicit generation.
\end{enumerate}

\begin{table*}[t]
\centering
\small
\caption{\textbf{Overview of the UniG2U dataset.} The benchmark contains 3,000 samples organized into 7 reasoning categories and 30 fine-grained subtasks, thoughtfully curated from multiple established multimodal reasoning datasets.}
\label{tab:dataset}
\begin{tabular}{@{} l p{6.5cm} p{3.5cm} c @{}}
\toprule
\textbf{Category} & \textbf{Subtasks} & \textbf{Data Sources} & \textbf{\# Samples} \\
\midrule
Real-world Applications 
& Attentional Focusing; Visual Shortest-Path 
& VSP \cite{vsp2025}, RealUnify \cite{shi2025realunify}
& 200 \\ \addlinespace

Geometry Reasoning 
& Planar Geometry; Solid Geometry 
& Geometry3K \cite{lu2023mathvista}, AuxSolidMath-Easy \cite{guo2026geovlmath} 
& 200 \\ \addlinespace

Physics Reasoning 
& Mechanics; Optics 
& PhyX \cite{phyx2025} 
& 200 \\ \addlinespace

Puzzles and Games 
& Mental Reconstruction; Mental Tracking; Visual Tracking; Maze; Jigsaw; Sliding 
& Uni-MMMU \cite{zou2025unimmmu}, RealUnify \cite{shi2025realunify}, BabyVision \cite{chen2026babyvision} 
& 537 \\ \addlinespace

Chart \& Table Reasoning 
& Chart Question Answering 
& ChartQA \cite{masry2022chartqa} 
& 100 \\ \addlinespace

Spatial Intelligence 
& MSR; Attribute (Measurement); Attribute (Approximation); Motion (Camera); Motion (Object) 
& MMSI-Bench \cite{yang2025mmsibench} 
& 500 \\ \addlinespace

Perception Reasoning 
& Icon-Scene; Icon-Shape; In-scene; In-shape; Logo-Scene; Logo-Shape; Algorithmic; Deductive; Spatial; Analogical; Inductive 
& IllusionBench \cite{zhang2025illusionbench}, Visual Puzzles \cite{visualpuzzles2025}, BabyVision \cite{chen2026babyvision} 
& 1263 \\ 

\midrule
\textbf{Total} & \textbf{30 fine-grained subtasks} & \textbf{Multiple sources} & \textbf{3,000} \\
\bottomrule
\end{tabular}
\end{table*}

By structuring the benchmark across seven categories with distinct reasoning characteristics, UniG2U enables fine-grained analysis of the relationship between generation and understanding. As illustrated in Fig.~\ref{fig:task-taxonomy}, the taxonomy spans both generation-friendly categories (e.g., spatial reasoning, puzzles) and more perception-heavy categories, allowing us to probe when intermediate visual generation is helpful, neutral, or harmful. Instead of merely reporting aggregate accuracy, this design allows us to investigate how the effectiveness of generation varies across cognitive regimes, providing deeper insight into the mechanisms of unified multimodal models.

\subsubsection{Task Selection and Coverage}
\label{sec:task-selection}

\begin{figure}[htbp]
  \centering
  \includegraphics[width=0.75\linewidth]{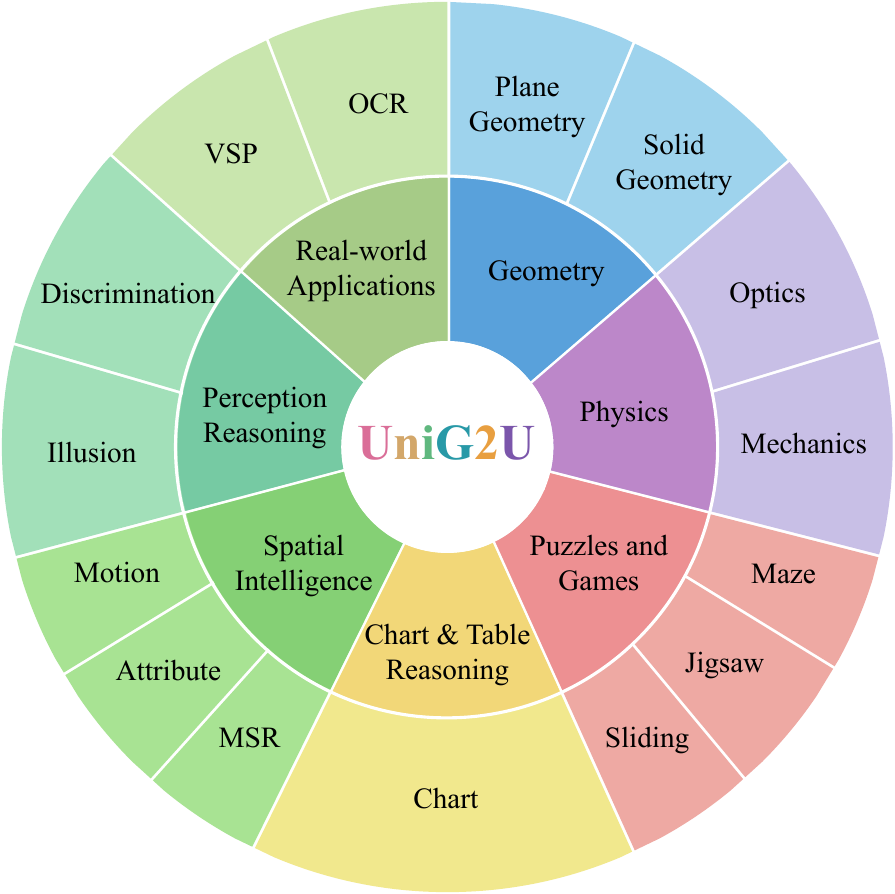}
  \caption{Task selection and coverage in UniG2U.}
  \label{fig:task-selection-coverage}
\end{figure}
\label{sec:task-format}

UniG2U is constructed around a single guiding hypothesis: there exists a family of multimodal understanding tasks for which intermediate visual generation can serve as an effective reasoning aid. Such tasks are common in practice, including (i) \textbf{spatial reasoning} where models benefit from externalizing transformations or layouts, (ii) \textbf{chart \& table reasoning} where re-expression or simplification of plots can reduce cognitive load, (iii) \textbf{puzzles and games} where intermediate states can be visualized to support state tracking, and (iv) \textbf{Physics/geometry} problems where auxiliary constructions and diagram manipulation often enable correct solutions.

To ensure both \emph{comprehensive coverage} and \emph{strong relevance} to the G2U objective, we curate \textbf{3,000} instances across \textbf{7} high-level categories and \textbf{30} subtasks (Table~\ref{tab:dataset}). As shown in Fig.~\ref{fig:task-selection-coverage}, our candidate pool is collected from diverse public sources (e.g., IllusionBench, MMSI-Bench, Visual Puzzles, Geometry3K, AuxSolidMath, Uni-MMMU, ChartQA, RealUnify, BabyVision, PhyX, and VSP-style benchmarks), from which we further \emph{select and filter} instances that best reflect settings where intermediate generation is likely to assist understanding.

In particular, the benchmark includes 200 samples each for Real-world Applications, Geometry Reasoning, and Physics Reasoning; 537 samples for Puzzles and Games; 100 samples for Chart \& Table Reasoning; 500 samples for Spatial Intelligence; and 1,263 samples for Perception Reasoning. This distribution allows us to analyze generation benefits across cognitive regimes ranging from generation-friendly spatial manipulation to primarily perceptual recognition.

It is worth noting that while the sample size per fine-grained subtask may appear relatively modest, this is a deliberate consequence of our rigorous selection criteria. Tasks that genuinely exhibit G2U synergy---where intermediate visual externalization provides a tangible reasoning advantage---are intrinsically scarce in existing datasets, which are historically skewed towards direct perception or text-only reasoning. Despite this inherent scarcity, UniG2U's collection of 3,000 meticulously curated samples renders it the largest benchmark of its kind compared to existing unified evaluations (\textbf{as quantitatively compared in Table~\ref{tab:benchmark_compare}}). Coupled with our extensive evaluation across more than 30 distinct unified models and baselines, UniG2U represents a truly large-scale, comprehensive, and highly persuasive investigation into the multimodal unification landscape.

Each instance in UniG2U is organized around a \emph{question-answering} format with an image (or a small set of images) as the primary input. In addition to the final answer, we annotate (when applicable) auxiliary information to facilitate diagnostic evaluation:
\begin{itemize}
    \item \textbf{Category / Subtask labels} for fine-grained analysis across the taxonomy.
    \item \textbf{Answer type} (e.g., multiple-choice, short text, numeric), which determines post-processing and matching rules.
    \item \textbf{Generation requirement tags} indicating whether a task naturally admits helpful intermediate visual representations (e.g., geometric construction, state transition visualization, chart re-expression).
\end{itemize}

Importantly, UniG2U does \emph{not} force a single canonical intermediate image for every problem. Instead, we treat intermediate generation as a \emph{reasoning tool} whose usefulness should be measured by whether it improves final task performance.

\subsection{Evaluation Protocol}
\label{sec:eval-protocol}

Our evaluation protocol is strictly designed to ensure \emph{fair comparison} across a diverse suite of models and to rigorously isolate the \emph{causal contribution of intermediate generation} (i.e., the G2U effect).

\begin{figure*}[t]
  \centering
  \includegraphics[width=\linewidth]{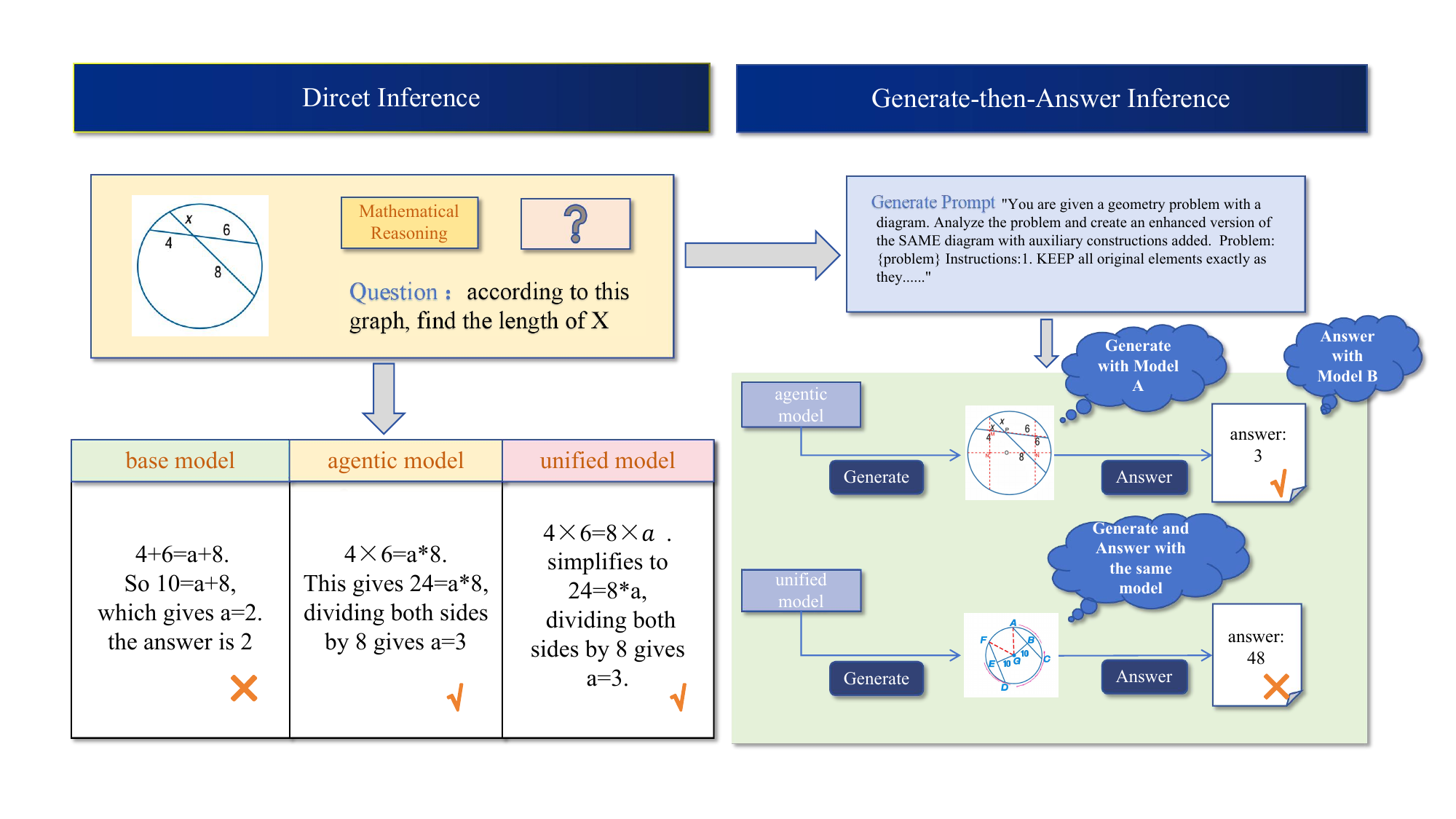}
  \caption{Overview of our two evaluation prompt formats: Direct and Generate-then-Answer (GtA).}
  \label{fig:stepwise-protocol}
\end{figure*}

\paragraph{Unified evaluation environment.}
To minimize implementation biases, we execute all evaluations under a unified architecture using the \texttt{lmms-eval} framework whenever possible. For proprietary models or specific architectures incompatible with this environment, we fallback to their native evaluation stacks while strictly preserving the rest of our protocol (i.e., prompt templates, budget constraints, and decoding strategies).

\paragraph{Dual-protocol prompting (Direct vs.\ GtA).}
To localize the effect of generation, we evaluate all models under two paired inference protocols, as illustrated in Fig.~\ref{fig:stepwise-protocol}:
\begin{itemize}
    \item \textbf{Direct}: The model processes the task instruction, input image(s), and query to directly output the final answer in a single end-to-end forward pass.
    \item \textbf{Generate-then-Answer (GtA)}: The instance is explicitly decomposed. The model is first prompted to generate an intermediate visual artifact $G$ (e.g., drawing auxiliary lines or tracking states). Subsequently, it generates the final answer using both the original input and the newly generated $G$ as expanded context.
\end{itemize}
For inherently multi-step reasoning problems (e.g., maze navigation, jigsaw puzzles), we employ an \emph{iterative GtA} variant: if an instance requires $K$ steps, the model alternates between producing a step-$k$ intermediate visual state and the corresponding step-$k$ decision, until the final answer is reached. Models unable to support the GtA mode due to architectural constraints receive specific fallback treatments, detailed in Appendix \ref{sec:appendix_ood_ablation}.

\paragraph{Budget matching and interaction constraints.}
To prevent performance gains from being confounded by disparate compute budgets, we enforce strict bounds. Across all models and task categories, we cap the maximum token generation length. Input images are processed at the model's default resolution; if out-of-memory (OOM) exceptions occur, we downscale to the nearest supported resolution.

\paragraph{Unified decoding settings.}
To ensure deterministic and reproducible evaluations, we standardize the text decoding hyperparameters across all models using greedy decoding ($\text{temperature} = 0$, $\text{top-}p = 1.0$). For intermediate image generation, we generally adopt the default configurations of the respective models.

\subsection{Metrics}
\label{sec:metrics}

\paragraph{Accuracy.}
We report accuracy (or exact match) as the primary baseline metric. Depending on the task format, we compute correctness via (i) \emph{rule-based evaluation} using text normalization followed by regex matching, or (ii) \emph{LLM-as-a-judge evaluation} by invoking the GPT-4o API with task-specific grading prompts. Unless otherwise noted, GPT-4o is strictly used to adjudicate non-trivial free-form outputs and visual-dependent judgments.

\paragraph{G2U Gain.}
Our core metric is the \textbf{G2U Gain} ($\Delta$), defined as the accuracy improvement of a unified model $M_{UM}$ relative to its rigorously paired purely discriminative base model $\mathcal{B}(M_{UM})$. Recalling Eq.~\ref{eq:g2u_decomp}, we report two distinct variants:
\begin{align}
\Delta_{\mathrm{Direct}} \;&=\; \mathrm{Acc}^{\text{Direct}}_{M_{UM}} \; -\; \mathrm{Acc}_{\mathcal{B}(M_{UM})}, \label{eq:direct_g2u_gain}\\
\Delta_{\mathrm{GtA}} \;&=\; \mathrm{Acc}^{\text{GtA}}_{M_{UM}} \; -\; \mathrm{Acc}_{\mathcal{B}(M_{UM})}. \label{eq:gta_g2u_gain}
\end{align}
This relative formulation effectively isolates the net capability shift introduced by the unified architecture and the generation process. We compute the G2U Gain both overall and per subtask to quantitatively diagnose when generation helps, remains neutral, or harms multimodal understanding.

\paragraph{Diagnostic metrics for GtA intermediates: RA and AL.}
To mechanically understand the effectiveness (or failure) of intermediate images produced during the GtA protocol, we introduce two novel alignment metrics evaluated by GPT-4o on a 1-to-5 scale, using customized rubrics for each task family.

\begin{figure}[t]
    \centering
    \includegraphics[width=\linewidth]{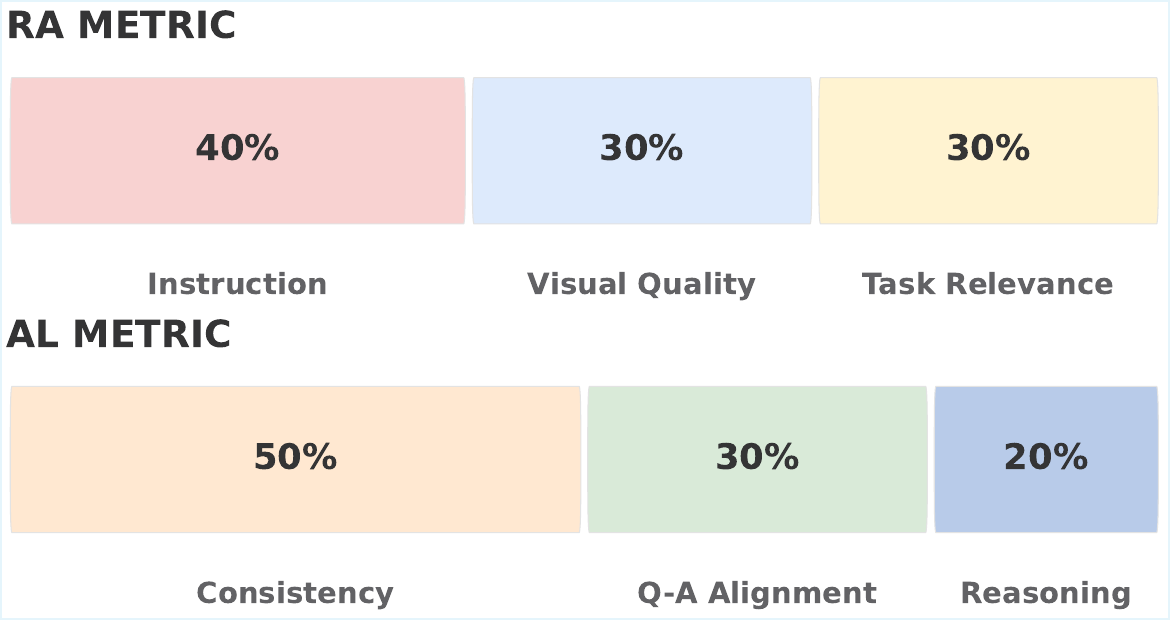}
    \caption{Rubric weights for GtA intermediate evaluation. RA measures the adherence and fidelity of the generated intermediate image w.r.t.\ the generation prompt, while AL measures the logical consistency between the final answer, the generated image, and the original question.}
    \label{fig:Rubric weights}
\end{figure}

\textbf{Reasoning-to-Visual Alignment (RA).}
RA evaluates whether the generated intermediate image correctly materializes the intended reasoning scaffold. It aggregates instruction adherence, visual quality, and task relevance (weighting schema detailed in Fig.~\ref{fig:Rubric weights}). Essentially, RA measures the \emph{fidelity of visual externalization}.

\textbf{Answer-to-Visual Alignment (AL).}
AL assesses whether the model's final answer is logically consistent with both the generated intermediate image and the original query. It aggregates visual-answer consistency, question-answer alignment, and reasoning coherence (see Fig.~\ref{fig:Rubric weights}). AL effectively measures the \emph{re-consumption capability} of the generated visual context.

\paragraph{Scoring schema.}
For both RA and AL, GPT-4o produces a final score on a 5-point scale by aggregating rubric-specific sub-scores. We report the mean RA/AL scores across relevant instances, providing granular breakdowns for each model and subtask to correlate intermediate generation quality with ultimate task accuracy.

\subsection{Models and Implementation Details}
\label{sec:models-impl}

To comprehensively study the G2U effect under realistic model ecosystems, we systematically evaluate a massive suite of \textbf{35 distinct models}. These are categorized into three families: \textbf{11} base VLMs, \textbf{21} native unified models, and \textbf{3} agentic models.

\paragraph{Base VLMs (understanding-only).}
We benchmark 11 strong vision-language models (VLMs) that perform \emph{pure understanding} without generating intermediate images. This set comprises the exact corresponding base models for all tested unified models, alongside leading VLM models GPT-4o and Ovis2.5. These base VLMs serve as the strict reference points for computing the G2U Gain (\cref{eq:direct_g2u_gain,eq:gta_g2u_gain}).

\paragraph{Native unified models.}
We benchmark 21 representative open-weight unified models that natively integrate multimodal understanding and image generation within a single model interface (i.e., the model can directly follow the \textbf{Direct} and \textbf{GtA} protocols without external tool wiring). These models serve as our primary subjects for evaluating intrinsic understanding-through-generation. Crucially, as detailed in Table~\ref{tab:unified-models}, the \textbf{Backbone} column explicitly denotes the corresponding Base VLM used for baseline comparisons.

\paragraph{Agentic models }
Finally, we evaluate 3 agentic models by composing a planner LLM with an external image generation tool. As established in our taxonomy, agentic systems are conceptually regarded as a generalized variant of unified models. Concretely, the planner produces a \texttt{generation\_prompt}, invokes the generation API to obtain an intermediate image, and then answers the query conditioned on this newly generated visual state. Notably, these agentic systems are typically driven by the most powerful, closed-source proprietary models available (e.g., \textbf{GPT-4o + GPT-image} and \textbf{Gemini + Nano Banana Pro}). Evaluating this category serves a critical dual purpose: on one hand, it benchmarks the performance upper-bound (ceiling) of generalized unified models on our tasks; on the other hand, it allows us to investigate how non-end-to-end, purely tool-driven visual externalization performs.

When a unified model does not have an official corresponding VLM base model (or its base model is not a vision-language model), we select the closest available VLM baseline in terms of training data and architecture, and use it as the strictly paired base model for comparison.

\paragraph{Models evaluated in this work.}
We exhaustively evaluate all 21 unified models listed in Table~\ref{tab:unified-models}, including: Emu3 \cite{wang2024emu3}, X-Omni \cite{geng2025xomni}, UniWorld-V1 \cite{lin2025uniworld}, MIO \cite{hu2025mio}, Bagel \cite{deng2025bagel}, OmniGen2 \cite{xiao2024omnigen}, Lumina-DiMOO \cite{xin2025luminadimoo}, Ovis-U1 \cite{wang2025ovisu1}, OneCAT-3B \cite{peng2025onecat}, JavisGPT \cite{liu2025javisgpt}, UAE \cite{yuan2025uae}, TokenFlow-XL \cite{li2024tokenflow}, UniPic2 \cite{peng2025unipic2}, Janus-Pro \cite{chen2025januspro}, Show-o2 \cite{xie2025showo2}, ILLUME+ \cite{zhang2025illume}, STAR-7B \cite{li2025star}, MMaDA \cite{yang2025mmada}, FUDOKI \cite{wang2025fudoki}, AIA \cite{zheng2025aia}, Uni-Video \cite{wei2025univideo}, and MammothModa2 \cite{shen2025mammothmoda2}.

An overview of these models and their key design choices is provided in Table~\ref{tab:unified-models}, detailing their backbones (Base VLMs), supported modalities, image generation paradigms, training recipes, i2i capability, and whether the unification is end-to-end or decoupled.

\begin{table*}[t]
\centering
\caption{\textbf{Comparison of Unified Multimodal Models.} \textcolor{cyan}{\faSnowflake}~indicates frozen backbone, \textcolor{orange}{\faUnlock}~indicates unfrozen (trainable) backbone. The \textbf{Backbone} column represents the corresponding Base VLM used for G2U evaluations.}
\label{tab:unified-models}
\resizebox{\textwidth}{!}{%
\setlength{\tabcolsep}{4pt}
\renewcommand{\arraystretch}{1.0}
\begin{tabular}{@{}llp{3.2cm}llllll@{}}
\toprule
\textbf{Model} & \textbf{Date} & \textbf{Backbone} & \textbf{Modality} & \textbf{Image Paradigm} & \textbf{Training} & \textbf{i2i} & \textbf{Unify} & \textbf{Text Paradigm} \\
\midrule
Emu3 & 2024.09 & LLaMA-2 \textcolor{orange}{\faUnlock} & image, audio & diffusion & UP+ST+AT & no (OOD) & end-to-end & AR \\
X-Omni & 2025.07 & Qwen-2.5 \textcolor{orange}{\faUnlock} & image & AR & UP+ST+AT & no (OOD) & decoupled & AR \\
UniWorld-V1 & 2025.06 & Qwen-2.5-VL \textcolor{cyan}{\faSnowflake} & image & diffusion & UP+ST & yes & decoupled & AR \\
MIO & 2025.10 & Yi-6B-Base \textcolor{orange}{\faUnlock} & image, audio & diffusion & UP+ST & yes & decoupled & AR \\
Bagel & 2025.05 & Qwen-2.5 \textcolor{orange}{\faUnlock} & image, video & Flow & UP+ST & yes & end-to-end & AR \\
OmniGen2 & 2025.06 & Qwen-2.5-VL \textcolor{orange}{\faUnlock} & image & Flow & UP+ST & yes & decoupled & AR \\
Lumina-DiMOO & 2025.04 & LLaDA \textcolor{orange}{\faUnlock} & image & Discrete Diffusion & UP+ST+AT & yes & end-to-end & Discrete Diffusion \\
Ovis-U1 & 2025.06 & Qwen3-1.7B \textcolor{orange}{\faUnlock} & image & diffusion & UP+ST & yes & decoupled & AR \\
OneCAT-3B & 2025.09 & Qwen-2.5-VL \textcolor{orange}{\faUnlock} & image & AR & UP+ST & yes & end-to-end & AR \\
JavisGPT & 2025.12 & Qwen2.5-VL \textcolor{orange}{\faUnlock} & video & diffusion & UP+ST & no & decoupled & AR \\
UAE & 2025.10 & Qwen-2.5-VL-3B \textcolor{orange}{\faUnlock} & image & diffusion & UP+ST+AT & no (OOD) & decoupled & AR \\
TokenFlow-XL & 2024.12 & Qwen-2.5-14B \textcolor{orange}{\faUnlock} & image & AR & UP+ST & no (OOD) & end-to-end & AR \\
UniPic2 & 2025.09 & Qwen-2.5-VL \textcolor{cyan}{\faSnowflake} & image & diffusion & UP+ST+AT & yes & decoupled & AR \\
Janus-Pro & 2025.01 & DeepSeek-LLM \textcolor{orange}{\faUnlock} & image & AR & UP+ST & no (OOD) & end-to-end & AR \\
Show-o2 & 2025.06 & Qwen2.5-7B \textcolor{orange}{\faUnlock} & image, video & Flow & UP+ST & yes & end-to-end & AR \\
ILLUME+ & 2025.04 & Qwen-2.5 \textcolor{orange}{\faUnlock} & image & AR + diffusion & UP+ST & yes & end-to-end & AR \\
STAR-7B & 2025.12 & Qwen-2.5-VL \textcolor{cyan}{\faSnowflake} & image & AR + diffusion & UP+ST & yes & end-to-end & AR \\
MMaDA & 2025.05 & LLaDA-Instruct \textcolor{orange}{\faUnlock} & image & Discrete Diffusion & UP+ST+AT & no (OOD) & end-to-end & Discrete Diffusion \\
FUDOKI & 2025.05 & Janus-1.5B \textcolor{orange}{\faUnlock} & image & Discrete Flow & UP+ST & no (OOD) & end-to-end & Discrete Flow \\
AIA & 2025.12 & Janus-Pro \textcolor{orange}{\faUnlock} & image & AR & ST & no (OOD) & end-to-end & AR \\
Uni-Video & 2026.01 & Qwen2.5-VL-7B \textcolor{cyan}{\faSnowflake} & image, video & diffusion & UP+ST & yes & decoupled & AR \\
MammothModa2 & 2025.11 & Qwen3-VL-8B \textcolor{orange}{\faUnlock} & image & AR + diffusion & UP+ST+AT & yes & end-to-end & AR \\
\bottomrule
\end{tabular}%
}
\end{table*}

All systems are evaluated under the same dataset, prompt templates, and budget constraints described in Section \ref{sec:eval-protocol}.

\subsection{Research Questions}
\label{sec:research-questions}

We define four primary research questions (RQs) to systematically guide our empirical study of the Generation-to-Understanding (G2U) paradigm. We systematically address these RQs in the subsequent Results section.

\paragraph{\textcolor{blue}{RQ1: Unified vs.\ Base Models.}}
Do unified multimodal models exhibit consistent capability advantages over their strictly paired base VLMs on our diagnostic benchmark?
\par\noindent\textbf{Experiment design:} For each unified model, we identify its corresponding purely discriminative base VLM and evaluate both under identical datasets, prompt templates, and inference budgets. We report the absolute performance and the Direct G2U Gain ($\Delta_{\mathrm{Direct}}$) overall and per cognitive category to characterize where implicit parametric unification yields benefits.

\paragraph{\textcolor{blue}{RQ2: Generate-then-Answer (GtA) vs.\ Direct Inference.}}
For complex reasoning tasks, do unified models benefit more from the \emph{explicit} visual externalization of intermediate steps (GtA protocol), or does \emph{Direct} inference achieve comparable understanding performance?
\par\noindent\textbf{Experiment design:} We evaluate each unified model under both the \textbf{Direct} and \textbf{GtA} protocols (\cref{sec:eval-protocol}) using matched compute budgets. We then quantify the performance delta ($\Delta_{\mathrm{GtA}}$) between the two modes to strictly isolate the causal contribution of explicitly generating intermediate visual reasoning scaffolds.

\paragraph{\textcolor{blue}{RQ3: Behavioral Correlation within Architectural Classes.}}
Within specific architectural families of unified models, do we observe correlated performance patterns that suggest class-consistent strengths and vulnerabilities in G2U tasks?
\par\noindent\textbf{Experiment design:} Following the taxonomy in \cref{tab:unified-models}, we group models by their image-generation paradigms (e.g., Autoregressive vs.\ Diffusion vs.\ Flow) and unification strategies (End-to-End vs.\ Decoupled). We construct model-by-task performance vectors and compute pairwise rank correlations to determine if fundamental architectural choices dictate distinct G2U behaviors.

\paragraph{\textcolor{blue}{RQ4: Fidelity and Utility of Intermediate Visualizations.}}
When explicitly prompted to generate intermediate auxiliary images under the GtA protocol, do unified models reliably follow instructions, and do they produce helpful visuals that genuinely facilitate downstream reasoning?
\par\noindent\textbf{Experiment design:} We diagnose the generated intermediate outputs using the two rubric-based evaluation metrics defined in \cref{sec:metrics}: \textbf{Reasoning-to-Visual Alignment (RA)} and \textbf{Answer-to-Visual Alignment (AL)}. We then analyze the correlation between RA/AL scores, downstream task accuracy, and the observed GtA-vs-Direct gains to mechanically trace the source of G2U successes and failures.

\section{Results Analysis}
\begin{figure*}[ht]
    \centering
    \includegraphics[width=0.95\linewidth]{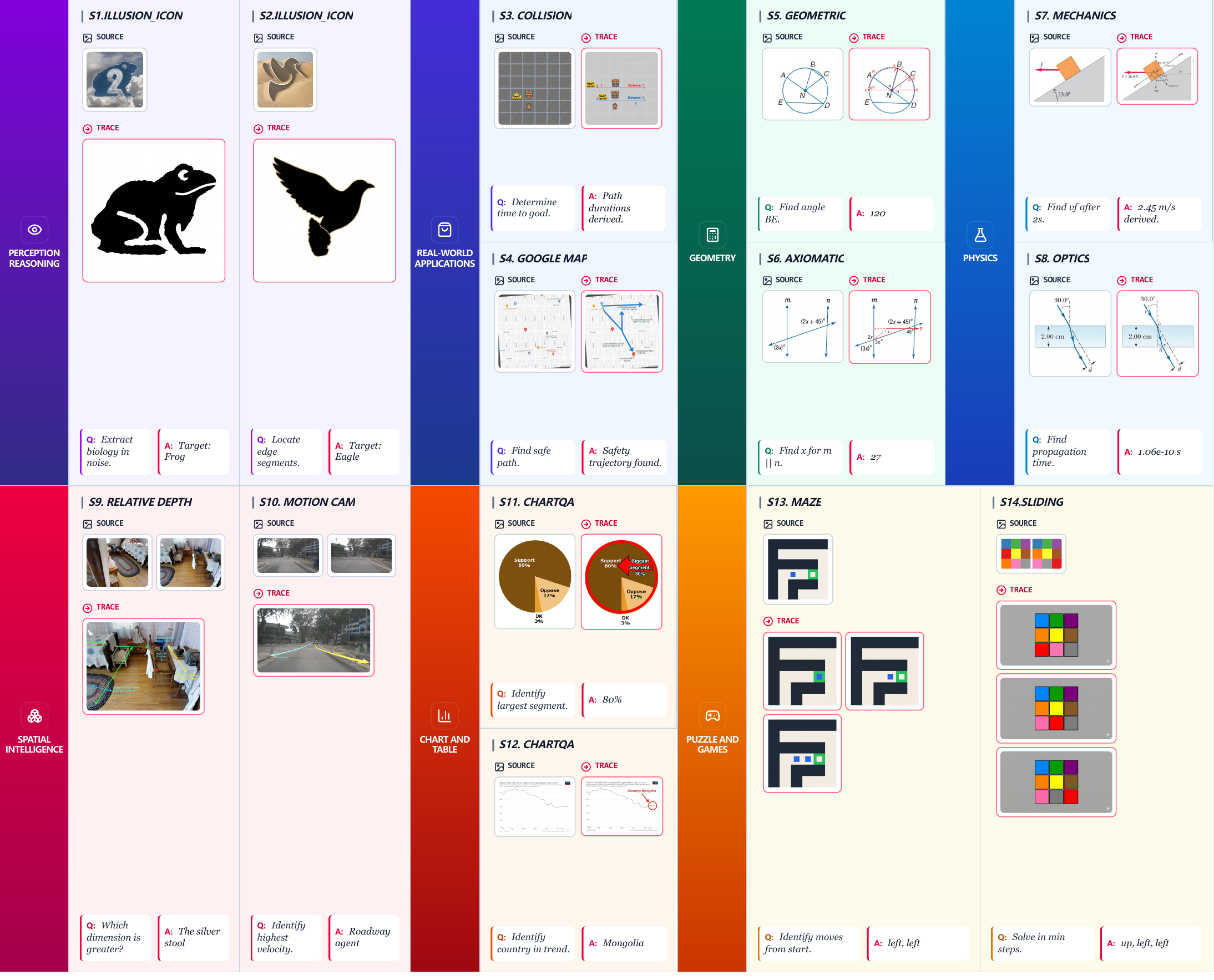}
    \caption{Representative case studies from UniG2U-Bench}
    \label{fig:placeholder}
\end{figure*}
\subsection{Main Results (RQ1)}
\label{sec:main-results}

\begin{table*}[!t]
\centering
\caption{\textbf{Main results on UniG2U across categories.} All values are reported as percentages ($\times 100$). The best result within each model group is \textbf{bolded}. Baseline models (marked with $*$) are highlighted with gray background. $\Delta$ denotes the performance gap to the respective baseline.}
\label{tab:main-results}
\resizebox{\textwidth}{!}{%
\setlength{\tabcolsep}{3.5pt}
\renewcommand{\arraystretch}{1.06}
\scriptsize
\begin{tabular}{@{}lcccccccc@{}}
\toprule
\textbf{Model} & \textbf{Real-world} & \textbf{Geometry} & \textbf{Physics} & \textbf{Puzzles} & \textbf{Chart} & \textbf{Spatial} & \textbf{Perception} & \textbf{Overall} \\
& \textbf{Apps} & \textbf{Reasoning} & & \textbf{\& Games} & & \textbf{Intel.} & \textbf{Reasoning} & \textbf{(G2U Gain$\Delta$)} \\
\midrule
\rowcolor{gray!12}
qwen2.5-vl-7b* & 40.25 & \textbf{23.00} & 34.00 & 25.73 & \textbf{86.00} & 23.40 & 39.42 & \textbf{34.45}~(+$0.00$) \\
OmniGen2(direct) & \textbf{41.25} & 14.00 & 41.00 & 20.91 & 86.00 & 23.80 & 35.62 & 31.99~\textcolor{blue}{(-$2.46$)} \\
OmniGen2(GtA) & 41.25 & 13.00 & \textbf{41.50} & 21.91 & 79.00 & \textbf{26.40} & 34.52 & 31.87~\textcolor{blue}{(-$2.58$)} \\
UniWorld-V1(GtA) & 24.00 & 19.00 & 34.50 & 24.72 & 80.00 & 24.80 & 39.35 & 32.96~\textcolor{blue}{(-$1.49$)} \\
OneCAT-3b(direct) & 39.50 & 17.00 & 32.00 & 22.05 & 76.68 & 25.20 & 34.55 & 31.15~\textcolor{blue}{(-$3.30$)} \\
OneCAT-3b(GtA) & 33.50 & 8.50 & 29.00 & 20.69 & 74.00 & 23.80 & 33.10 & 28.80~\textcolor{blue}{(-$5.65$)} \\
JavisGPT(direct) & 32.00 & 21.00 & 26.50 & 23.72 & 78.60 & 21.60 & 35.52 & 30.72~\textcolor{blue}{(-$3.73$)} \\
uni-video(GtA) & 34.00 & 21.50 & 29.00 & \textbf{29.33} & 61.00 & 20.00 & \textbf{42.76} & 34.25~\textcolor{blue}{(-$0.20$)} \\
UniPic2(GtA) & 17.50 & 17.00 & 31.00 & 23.95 & 47.00 & 24.60 & 38.80 & 30.65~\textcolor{blue}{(-$3.80$)} \\
STAR-7B(GtA) & 34.00 & 14.50 & 32.50 & 23.56 & 79.00 & 25.20 & 38.56 & 32.68~\textcolor{blue}{(-$1.77$)} \\
\midrule
\rowcolor{gray!12}
Qwen2.5-VL-3B* & 41.00 & 15.50 & \textbf{40.50} & 20.42 & 87.00 & \textbf{26.40} & 35.55 & \textbf{32.39}~(+$0.00$) \\
UAE(direct) & 36.75 & 10.50 & 30.50 & 19.06 & 85.00 & 25.20 & \textbf{36.37} & 30.94~\textcolor{blue}{(-$1.45$)} \\
Ovis-U1(direct) & \textbf{45.50} & \textbf{23.50} & 30.50 & \textbf{27.28} & \textbf{90.00} & 17.60 & 34.92 & 32.15~\textcolor{blue}{(-$0.24$)} \\
Ovis-U1(GtA) & 21.00 & 16.50 & 17.00 & 18.00 & 32.00 & 20.40 & 30.56 & 24.19~\textcolor{blue}{(-$8.20$)} \\
\midrule
\rowcolor{gray!12}
yi-6B-vl* & 24.00 & 0.00 & 33.50 & 13.22 & 19.00 & 22.60 & 31.20 & 23.73~(+$0.00$) \\
mio(direct) & 18.00 & 0.00 & 31.00 & 12.85 & 15.00 & \textbf{23.40} & 25.49 & 20.70~\textcolor{blue}{(-$3.03$)} \\
mio(GtA) & 15.00 & 0.00 & 28.00 & 8.94 & 13.00 & 21.60 & 24.94 & 19.00~\textcolor{blue}{(-$4.73$)} \\
\midrule
\rowcolor{gray!12}
deepseek-v2* & 11.00 & 4.50 & 22.00 & 18.02 & \textbf{74.00} & 17.40 & 35.07 & 25.86~(+$0.00$) \\
Janus-Pro(direct) & \textbf{27.50} & 7.50 & 19.00 & 22.49 & 29.00 & 24.20 & 35.08 & 27.39~\textcolor{red}{(+$1.53$)} \\
AIA(direct) & 13.50 & \textbf{8.50} & \textbf{31.00} & \textbf{23.95} & 33.00 & \textbf{26.60} & \textbf{36.34} & \textbf{28.65}~\textcolor{red}{(+$2.79$)} \\
\midrule
\rowcolor{gray!12}
LLaDA-Instruct* & 27.25 & 1.50 & 21.00 & \textbf{20.83} & 9.00 & 23.40 & 13.78 & 17.05~(+$0.00$) \\
MMaDA(direct) & \textbf{28.75} & \textbf{5.00} & \textbf{26.50} & 16.01 & \textbf{14.00} & \textbf{25.00} & \textbf{22.74} & \textbf{21.09}~\textcolor{red}{(+$4.04$)} \\
\midrule
\rowcolor{gray!12}
Janus-1.3B* & 13.50 & \textbf{2.50} & \textbf{29.00} & \textbf{12.10} & \textbf{50.00} & 21.80 & \textbf{23.52} & \textbf{20.36}~(+$0.00$) \\
FUDOK(direct) & \textbf{17.50} & 1.00 & 21.00 & 9.31 & 22.00 & \textbf{29.00} & 15.52 & 16.40~\textcolor{blue}{(-$3.96$)} \\
\midrule
\rowcolor{gray!12}
Qwen3-VL-8B* & \textbf{40.00} & \textbf{39.00} & \textbf{37.00} & \textbf{24.24} & 89.00 & \textbf{26.00} & \textbf{43.66} & \textbf{37.75}~(+$0.00$) \\
MammothModa2(direct) & 33.50 & 0.00 & 36.50 & 12.10 & \textbf{91.00} & 21.80 & 39.11 & 29.97~\textcolor{blue}{(-$7.78$)} \\
\midrule
\rowcolor{gray!12}
llava-hf* & \textbf{23.00} & 0.00 & \textbf{29.50} & \textbf{18.96} & 28.00 & \textbf{24.60} & \textbf{33.57} & \textbf{26.06}~(+$0.00$) \\
Emu3(direct) & 16.50 & \textbf{1.00} & 20.50 & 13.37 & \textbf{67.00} & 23.20 & 24.78 & 21.46~\textcolor{blue}{(-$4.60$)} \\
\midrule
\rowcolor{gray!12}
llava-onevision* & \textbf{43.00} & 10.50 & \textbf{44.00} & 24.48 & 78.00 & 25.40 & 37.13 & 33.35~(+$0.00$) \\
X-Omni(direct) & 33.00 & 9.50 & 29.50 & 26.24 & 83.00 & 27.00 & 35.31 & 31.63~\textcolor{blue}{(-$1.72$)} \\
bagel(direct) & 38.00 & 29.00 & 37.50 & 23.77 & \textbf{92.00} & 28.40 & \textbf{39.96} & 35.84~\textcolor{red}{(+$2.49$)} \\
bagel(GtA) & 32.75 & \textbf{30.00} & 40.50 & \textbf{31.58} & 91.00 & \textbf{29.00} & 37.29 & \textbf{36.10}~\textcolor{red}{(+$2.75$)} \\
TokenFlow-XL(direct) & 32.00 & 21.00 & 26.50 & 23.72 & 78.60 & 21.60 & 35.52 & 30.72~\textcolor{blue}{(-$2.63$)} \\
show-o2(direct) & 35.50 & 11.00 & 40.00 & 26.01 & 45.00 & 25.80 & 36.50 & 31.59~\textcolor{blue}{(-$1.76$)} \\
show-o2(GtA) & 33.00 & 9.00 & 40.00 & 26.39 & 35.00 & 20.40 & 28.11 & 26.59~\textcolor{blue}{(-$6.76$)} \\
ILLUME+(direct) & 24.50 & 7.50 & 35.00 & 21.06 & 80.00 & 23.20 & 35.08 & 29.54~\textcolor{blue}{(-$3.81$)} \\
ILLUME+(GtA) & 20.50 & 2.00 & 35.50 & 19.53 & 58.00 & 21.00 & 34.05 & 27.13~\textcolor{blue}{(-$6.22$)} \\
\midrule
\rowcolor{orange!8}
gpt4o & \textbf{45.25} & 27.00 & 34.50 & \textbf{30.73} & 58.00 & \textbf{36.00} & \textbf{43.73} & \textbf{38.96} \\
ovis2.5 & 44.00 & \textbf{32.50} & \textbf{43.00} & 23.13 & \textbf{89.00} & 27.80 & 42.28 & 37.51 \\
\midrule
GPT-4o + GPT-image & 32.75 & 27.00 & 30.50 & 32.72 & 60.00 & 27.00 & 40.54 & 35.44 \\
gemini pro +nano banana pro & \textbf{71.50} & \textbf{85.00} & \textbf{91.00} & \textbf{55.31} & 68.00 & \textbf{38.40} & \textbf{61.12} & \textbf{60.80} \\
Qwen2.5-7b+Qwen-edit & 33.75 & 17.00 & 31.50 & 27.79 & \textbf{80.00} & 21.60 & 38.56 & 32.96 \\
\bottomrule
\end{tabular}%
}

\vspace{2pt}

{\scriptsize \textbf{Note:} * denotes the base model. $\Delta$ is computed as overall minus the base model's overall within each block.}
{\scriptsize \textbf{Bold:} Within each block (same model family), the best score in each category is highlighted in bold; ties are broken by bolding the first occurrence.}
\end{table*}

Table~\ref{tab:main-results} summarizes the performance of all evaluated models across the seven cognitive categories of UniG2U. To rigorously answer RQ1, we group unified models with their strictly paired base VLMs (marked with $^*$). The metric $\Delta$ isolates the absolute G2U Gain (or loss) introduced by the unified architecture.\footnote{\textbf{Evaluation constraints:} To ensure strictly fair comparisons, we exclude GtA results for models inherently lacking image-editing training (OOD for visual modification). Similarly, we omit Direct results for models with frozen backbones, as their performance deviations stem from projection layers rather than parameter-level multimodal coupling. Detailed treatments and analyses are deferred to Appendix~\ref{sec:appendix_ood_ablation} and \ref{Frozen_Backbones}.}

\paragraph{Overall Trends: The ``Alignment Tax'' of Unification.} 
~\\
\textbf{\textcolor{blue}{Takeaway 1: On the majority of tasks, integrating generation capabilities leads to a degradation in pure understanding performance.}}

One might naturally hypothesize that this degradation in pure understanding is merely a symptom of ``catastrophic forgetting'' during the multimodal joint training phase. However, empirical evidence solidly refutes this. The native unified models evaluated in our benchmark undergo extensive co-training on massive understanding datasets alongside generation data, a recipe explicitly designed to preserve pure comprehension capabilities. Therefore, rather than a loss of previously acquired knowledge, the observed degradation must inherently stem from the cross-modal ``alignment tax'' and the objective interference introduced by routing through the generative pathway.

A striking observation across Table~\ref{tab:main-results} is that unified models frequently yield negative $\Delta$ values compared to their corresponding base VLMs. Crucially, this degradation persists even under the \emph{Direct} protocol (where no explicit image is generated). This indicates that the decline is not merely a byproduct of low-quality intermediate visual artifacts, but rather an intrinsic consequence of the \emph{parameter-level coupling} between generation and understanding. 

This phenomenon suggests a representational trade-off, akin to an ``alignment tax.'' Joint multimodal training objectives intended to fuse generation and perception often introduce \textbf{objective interference}, slightly compromising the fine-grained discriminative reasoning abilities originally present in the base understanding models.

\paragraph{Category-Level Behaviors: Isolated Gains Amidst Overall Decline.} 
~\\
\textbf{\textcolor{blue}{Takeaway 2: Despite the general degradation in overall understanding, unified models exhibit isolated resilience and notable gains in Spatial Intelligence and Visual Illusions.}}

A finer-grained category analysis reveals that the negative impact of parameter-level coupling is not absolute. While many general reasoning categories suffer from objective interference, unified models consistently demonstrate clearer resilience or even modest gains in \textbf{Spatial Intelligence} and \textbf{Visual Illusions} (a specific subset of Perception Reasoning). 

To illustrate this trend, Table~\ref{tab:small_excerpt} presents an excerpt of the results focusing on the \texttt{llava\_onevision} model family. As shown, despite the overall ``alignment tax,'' generation-integrated models like \texttt{bagel} and \texttt{show-o2} achieve notable positive gains ($\Delta$) in tasks requiring transformation-aware reasoning and geometric consistency. For the complete experimental results across all evaluated base VLMs, as well as a more detailed analysis of how generation-induced representations regularize internal spatial structures, please refer to Appendix~\ref{sec:appendix_subtask_analysis}.

\begin{table}[htbp]
\centering
\caption{\textbf{Excerpt of Absolute Results on Selected Illusion and Spatial Subtasks.} Baseline model is highlighted in gray. See Appendix~\ref{sec:appendix_subtask_analysis} for the comprehensive table encompassing all model blocks.}
\label{tab:small_excerpt}
\resizebox{\columnwidth}{!}{%
\begin{tabular}{lcccccc}
\toprule
\multirow{2}{*}{\textbf{Model}} 
& \multicolumn{2}{c}{\textbf{Illusion}} 
& \multicolumn{3}{c}{\textbf{Spatial Intelligence}} 
& \multirow{2}{*}{\textbf{Overall} \textbf{($\Delta$)}} \\
\cmidrule(lr){2-3}\cmidrule(lr){4-6}
& \textbf{icon\_shape} 
& \textbf{in\_shape}
& \textbf{MSR} 
& \textbf{Attr. (Meas.)} 
& \textbf{Motion (Cam.)}
& \\
\midrule
\rowcolor{gray!12}
llava\_onevision* & 6 & 18 & 17 & 26 & 20 & 17.40~(+$0.00$) \\
X-Omni(direct) & 5 & 11 & 20 & 31 & 24 & 18.20~\textcolor{red}{(+$0.80$)} \\
bagel(direct) & 2 & 18 & 26 & 31 & 26 & \textbf{20.60~\textcolor{red}{(+$3.20$)}} \\
bagel(GtA) & 2 & 17 & 25 & 36 & 24 & 20.80~\textcolor{red}{(+$3.40$)} \\
TokenFlow-XL(direct) & 12 & 13 & 16 & 24 & 19 & 16.80~\textcolor{blue}{(-$0.60$)} \\
show-o2(direct) & 11 & 22 & 27 & 24 & 28 & 22.40~\textcolor{red}{(+$5.00$)} \\
show-o2(GtA) & 6 & 18 & 30 & 15 & 23 & 18.40~\textcolor{blue}{(+$1.00$)} \\
ILLUME+(direct) & 4 & 20 & 28 & 29 & 14 & 19.00~\textcolor{red}{(+$1.60$)} \\
ILLUME+(GtA) & 14 & 16 & 29 & 28 & 16 & 20.60~\textcolor{red}{(+$3.20$)} \\
\bottomrule
\end{tabular}%
}
\end{table}

These specialized tasks heavily involve geometric transformations, state tracking, and resolving complex visual ambiguities. In these specific domains, the integration of generative capabilities does not act as an interfering distraction; rather, it appears to serve as a powerful structural regularizer. By learning to synthesize and manipulate images, the model significantly enhances its foundational \emph{perceptual acuity} and its ability to explicitly \emph{model visual structures}. This suggests that while current unification paradigms may dilute general reasoning, they hold distinct and unique promise for tasks strictly grounded in structural and spatial modeling.

\paragraph{Implications for RQ1.}
Our findings demonstrate that G2U is fundamentally a challenge of \emph{representation coupling}. Simply injecting a generative pathway into a VLM does not automatically yield a stronger reasoner. Unification only benefits reasoning when the generative and understanding objectives are coherently aligned (as seen in spatial tasks); otherwise, the added generative capacity acts as an interfering auxiliary objective, diluting the model's core cognitive reasoning power.

\subsection{Comparison Between GtA and Direct Inference (RQ2)}
\label{sec:rq2}

\begin{table}[ht]
\centering
\caption{\textbf{Direct vs.\ GtA accuracy on generation-friendly subtasks.} GtA shows consistent improvements in multi-step spatial reasoning regimes. Results are formatted as (Direct / GtA).}
\label{tab:rq2_spatial_subtasks}
\small
\begin{tabular}{@{}lccc@{}}
\toprule
\textbf{Model} & \textbf{MSR} & \textbf{Maze} & \textbf{Sliding} \\
& (Dir / GtA) & (Dir / GtA) & (Dir / GtA) \\
\midrule
OmniGen2 & 0.220 / \textbf{0.290} & 0.043 / \textbf{0.046} & 0.000 / \textbf{0.001} \\
OneCAT-3B & 0.320 / \textbf{0.390} & \textbf{0.084} / 0.074 & \textbf{0.112} / 0.031 \\
Ovis-U1 & 0.120 / \textbf{0.270} & \textbf{0.134} / 0.003 & \textbf{0.334} / 0.007 \\
MIO & 0.210 / \textbf{0.320} & 0.000 / 0.000 & 0.000 / 0.000 \\
Bagel & \textbf{0.260} / 0.250 & 0.021 / \textbf{0.281} & 0.009 / \textbf{0.195} \\
Show-o2 & 0.270 / \textbf{0.300} & 0.142 / \textbf{0.178} & 0.137 / \textbf{0.183} \\
ILLUME+ & 0.280 / \textbf{0.290} & 0.069 / \textbf{0.129} & \textbf{0.040} / 0.000 \\
\bottomrule
\end{tabular}
\end{table}

\begin{figure}[ht]
    \centering
    \includegraphics[width=0.45\textwidth]{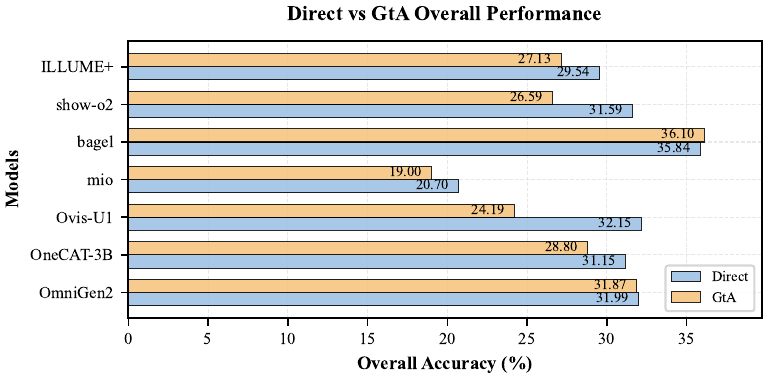}
    \caption{\textbf{GtA vs.\ Direct (Category-level).} Each point represents a (model, category) pair comparing Direct and GtA accuracy.}
    \label{fig:rq2_scatter}
\end{figure}

\begin{figure}[ht]
    \centering
    \includegraphics[width=0.5\textwidth]{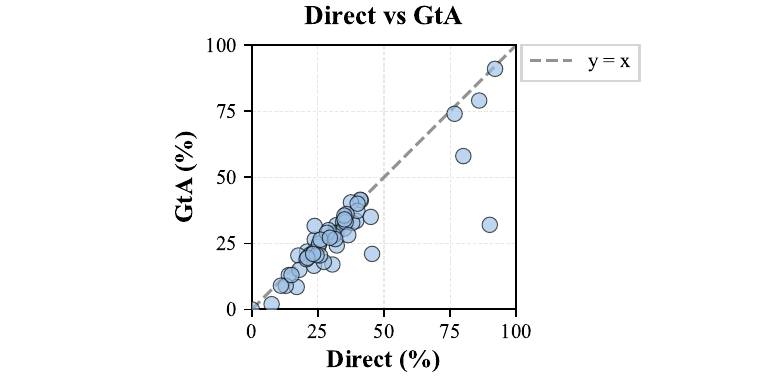}
    \caption{\textbf{GtA vs.\ Direct (Overall).} Overall accuracy comparison for models supporting both inference protocols.}
    \label{fig:rq2_overall_bar}
\end{figure}

To investigate whether explicit intermediate visual generation provides additional benefits over direct unified inference, we evaluate each unified model under both the \textbf{Direct} and \textbf{GtA} protocols with matched inference budgets.

\paragraph{Overall Trend: The Risk of Visual Error Propagation.}
~\\
\textbf{\textcolor{blue}{Takeaway 3: The explicitly generation-augmented paradigm (GtA) degrades performance across most logic-intensive tasks, as forced visual externalization often propagates errors.}}

\begin{table*}[!t]
\centering
\small
\caption{
\textbf{Intermediate-image alignment across task categories.} Each cell reports the \textbf{RA / AL} scores (out of 5.0). Higher RA indicates stronger adherence to the intended visual reasoning scaffold; higher AL indicates tighter logical consistency between the final answer and the generated artifact.
}
\label{tab:rq4_alignment}
\resizebox{\textwidth}{!}{%
\renewcommand{\arraystretch}{1.15}
\begin{tabular}{@{}lcccccccc@{}}
\toprule
\textbf{Model} & \textbf{Real-world} & \textbf{Geometry \&} & \textbf{Physics} & \textbf{Puzzles} & \textbf{Chart \&} & \textbf{Spatial} & \textbf{Perception} & \textbf{Overall} \\
& \textbf{Apps} & \textbf{Geometry} & & \textbf{\& Games} & \textbf{Table} & \textbf{Intel.} & \textbf{Reasoning} & \textbf{(RA / AL)} \\
\midrule
Bagel & 2.25/3.17 & 3.05/4.52 & 2.04/2.97 & 2.66/2.86 & 1.96/4.66 & 2.86/3.54 & 3.22/3.90 & 2.86/3.61 \\
UAE & 1.78/2.79 & 1.56/2.17 & 1.98/3.10 & 2.10/2.83 & 1.07/3.74 & 2.14/2.83 & 2.94/3.45 & 2.36/3.09 \\
OmniGen2 & 1.51/2.90 & 1.26/1.56 & 1.49/2.31 & 1.88/2.40 & 1.34/2.97 & 1.79/2.70 & 2.13/3.22 & 1.86/2.79 \\
GPT-4o + GPT-image & 2.43/3.90 & 4.69/4.72 & 2.04/2.97 & 2.70/2.75 & 3.38/4.71 & 4.35/4.34 & 4.26/4.63 & 3.73/4.09 \\
Ovis-U1 & 1.67/2.52 & 1.80/3.23 & 1.85/2.85 & 1.45/2.18 & 1.24/3.37 & 2.26/3.65 & 2.33/3.46 & 2.01/3.14 \\
ILLUME+ & 1.39/2.95 & 1.41/2.71 & 1.34/2.08 & 2.01/2.50 & 1.01/4.41 & 1.42/2.05 & 2.59/3.30 & 2.00/2.84 \\
OneCAT-3B & 1.87/2.59 & 1.75/2.92 & 1.61/2.12 & 1.94/2.58 & 1.35/4.07 & 1.90/2.66 & 2.92/3.41 & 2.29/2.99 \\
UniWorld-V1 & 1.62/2.88 & 1.48/2.15 & 1.74/2.76 & 1.60/2.41 & 1.38/3.58 & 2.01/3.12 & 2.36/3.44 & 1.98/3.04 \\
MIO & 1.92/2.70 & 2.42/2.88 & 1.98/2.91 & 2.24/2.36 & 1.95/3.82 & 2.65/3.22 & 3.18/3.76 & 2.67/3.24 \\
Qwen2.5-7B + Qwen-Edit & 1.68/2.95 & 1.32/1.88 & 1.55/2.48 & 1.72/2.46 & 1.47/3.44 & 1.92/3.08 & 2.41/3.55 & 1.99/3.05 \\
Uni-Video & 2.34/4.33 & 1.55/3.47 & 2.92/3.98 & 2.84/3.75 & 2.15/4.64 & 1.88/3.74 & 3.39/4.69 & 2.77/4.21 \\
Show-o2 & 1.08/2.04 & 1.00/1.26 & 1.17/1.61 & 1.06/1.26 & 1.50/3.47 & 1.03/1.63 & 1.68/2.74 & 1.34/2.09 \\
UniPic2 & 1.75/2.60 & 1.69/3.21 & 2.50/3.49 & 1.82/2.30 & 1.33/3.51 & 2.07/2.89 & 2.85/4.16 & 2.31/3.38 \\
STAR-7B & 2.11/3.60 & 1.54/2.44 & 2.37/3.50 & 1.85/2.43 & 2.77/3.72 & 2.22/3.53 & 2.85/3.75 & 2.39/3.36 \\
X-Omni & 1.78/2.71 & 1.11/1.29 & 2.21/2.95 & 1.97/2.37 & 1.75/2.45 & 1.60/2.18 & 1.85/2.42 & 1.80/2.35 \\
\bottomrule
\end{tabular}
}
\end{table*}

At the category level (Fig.~\ref{fig:rq2_scatter}), a majority of points lie below the $y=x$ line, indicating that GtA inference does not consistently outperform Direct inference across task regimes. This trend is further reflected in the aggregate comparison (Fig.~\ref{fig:rq2_overall_bar}), where GtA accuracy is frequently lower than Direct for models supporting both protocols. For several unified models (e.g., OneCAT-3B, Ovis-U1, and MIO), the degradation is substantial. 

This overall decline highlights a critical vulnerability: explicitly externalizing intermediate visuals does not automatically improve reasoning. Instead, GtA prompting introduces severe failure modes when the generated artifact is inaccurate, semantically misaligned, or poorly integrated. From a coupling perspective, Direct mode allows the model to implicitly leverage generation-trained representations within its hidden space, completely avoiding the risk of committing to a flawed concrete image. In contrast, GtA forces a definitive visual hypothesis; if generative precision is lacking, this flawed visualization becomes a source of noise that misleads the subsequent answering module (see qualitative failure cases in Appendix~\ref{sec:appendix_failure_cases}).

\paragraph{Transformation-Intensive Regimes: Visual CoT.}
~\\
\textbf{\textcolor{blue}{Takeaway 4: Explicit intermediate generation (GtA) yields noticeable performance gains in transformation-intensive scenarios where visual states serve as an external workspace.}}

Despite the overall drop, GtA is not uniformly detrimental. A fine-grained subtask analysis reveals that explicit generation yields consistent improvements in specific transformation-intensive regimes, such as \textbf{Maze navigation}, \textbf{Sliding puzzles}, and \textbf{Multi-Step Spatial Reasoning (MSR)} (Table~\ref{tab:rq2_spatial_subtasks}). 

These tasks share a defining structural property: they demand sequential, multi-step spatial manipulation and iterative state tracking. In such scenarios, explicit intermediate generation effectively functions as a \emph{Visual Chain-of-Thought (CoT)}. By explicitly drawing out intermediate spatial configurations, the model decomposes complex reasoning trajectories into manageable sub-steps. The generated artifact acts as an external cognitive workspace, stabilizing state tracking and significantly reducing the implicit memory burden required to track moving objects or paths across a long horizon.

Taken together, these findings indicate that the decisive factor for G2U success is not merely the act of explicit generation, but whether the generation functions as a reliable structural scaffold. Direct inference captures broad representational benefits, whereas GtA provides reasoning gains \emph{only} when intermediate visuals are both structurally accurate and semantically aligned with sequential spatial logic.

\subsection{Correlation Analysis of G2U Gains Across Tasks and Models (RQ3)}
\label{sec:rq3}

\begin{figure*}[ht]
    \centering
    \begin{subfigure}[t]{0.32\textwidth}
        \centering
        \includegraphics[width=\textwidth]{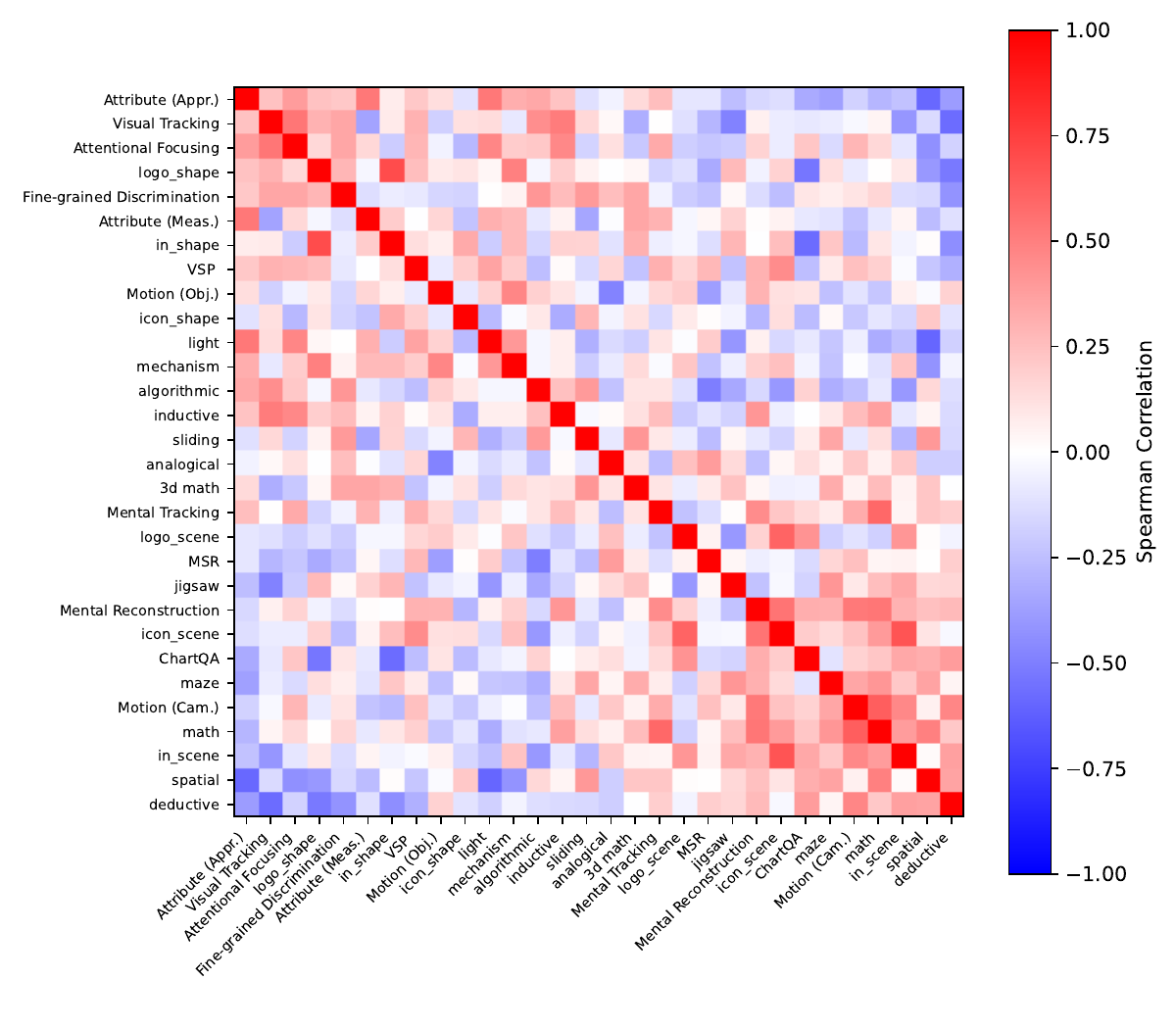}
        \caption{Task-level correlation}
        \label{fig:rq3_task_corr}
    \end{subfigure}
    \hfill
    \begin{subfigure}[t]{0.32\textwidth}
        \centering
        \includegraphics[width=\textwidth]{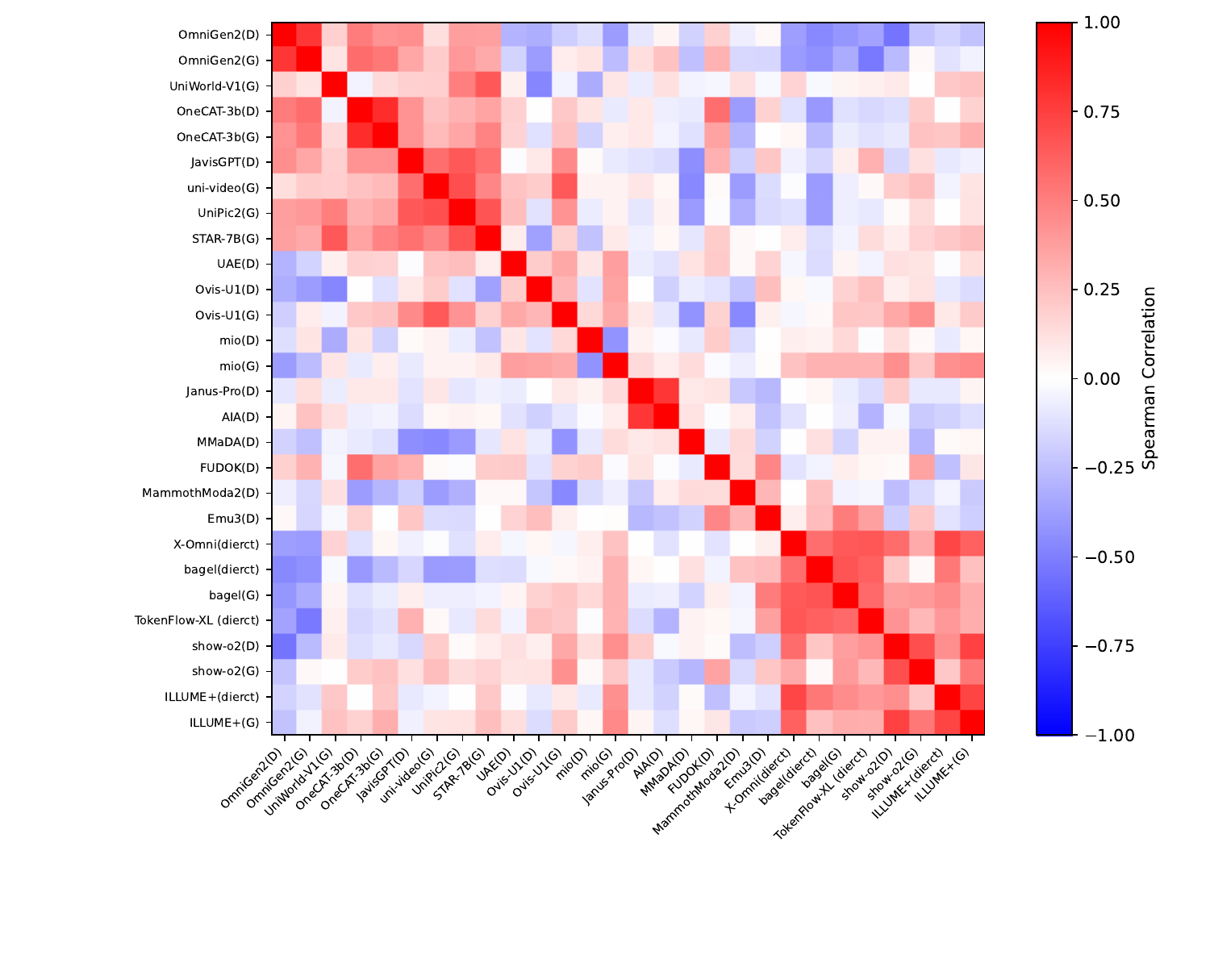}
        \caption{Model-level correlation (by Base VLM)}
        \label{fig:rq3_model_corr_s3}
    \end{subfigure}
    \hfill
    \begin{subfigure}[t]{0.32\textwidth}
        \centering
        \includegraphics[width=\textwidth]{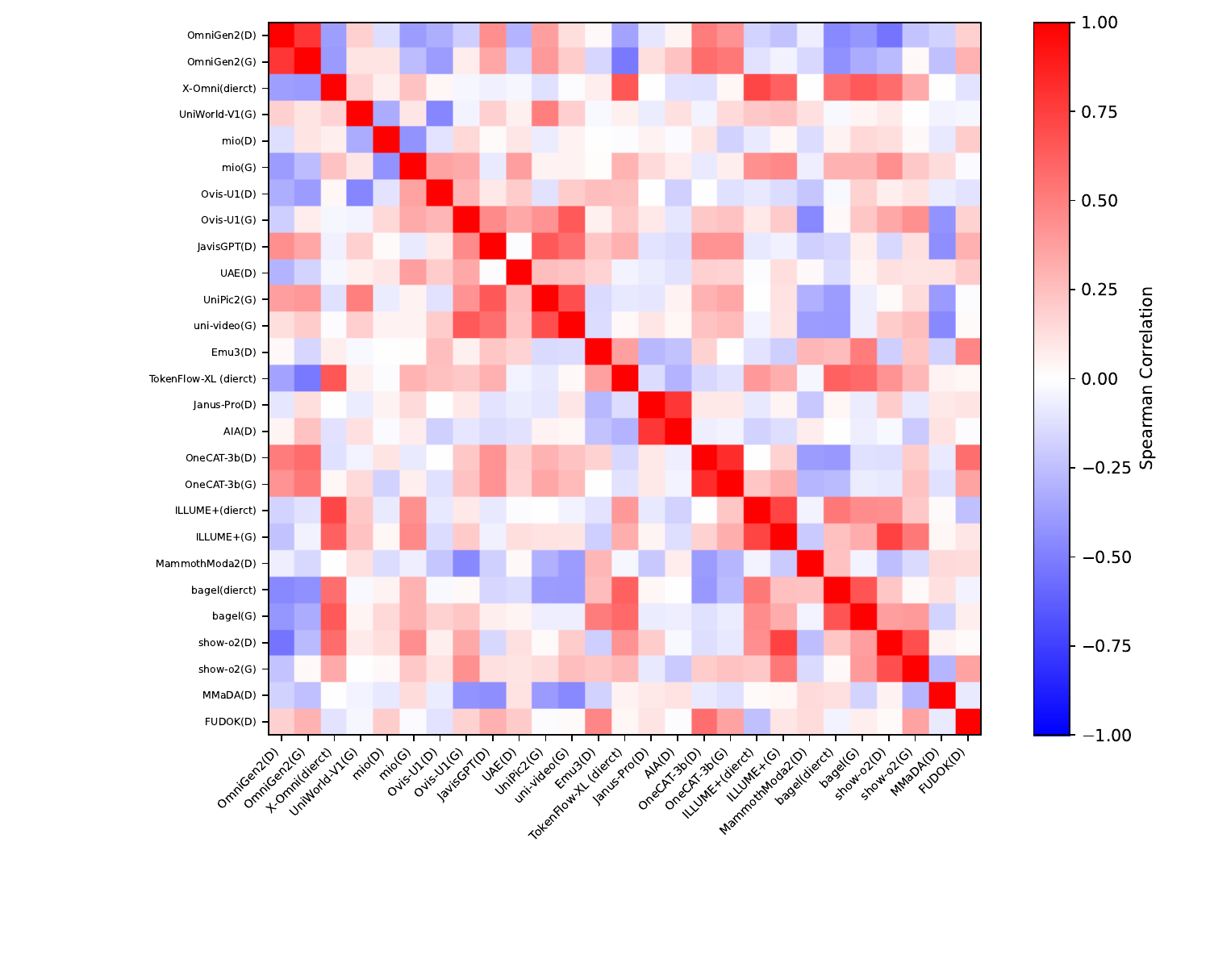}
        \caption{Model-level correlation (by Architecture)}
        \label{fig:rq3_model_corr_s5}
    \end{subfigure}
\caption{
    \textbf{Spearman correlation heatmaps of G2U gains ($\Delta$).} 
    Rather than raw accuracy, these matrices visualize the correlation of the relative offset ($\Delta$) to isolate generation-induced effects. 
    (a) \textbf{Task-level:} Evaluates how different subtasks align in terms of G2U gains. The subtasks along the axes are roughly ordered from perception-oriented (left/top) to reasoning-oriented (right/bottom). Perception tasks and reasoning tasks each form distinctly positive internal clusters, yet they exhibit strong negative correlations with one another, highlighting a fundamental cognitive trade-off. 
    (b) \textbf{Model-level (Base VLM):} Models sharing the same foundational base weights exhibit highly similar G2U behavior, forming distinct high-correlation blocks. 
    (c) \textbf{Model-level (Architecture):} Grouping models purely by their generative paradigm (e.g., Diffusion vs.\ Autoregressive) yields weaker correlation patterns, indicating that inherited foundational representations govern G2U effects more strongly than the chosen generative architecture.
    }
    \label{fig:rq3_all_corr}
\end{figure*}

While RQ1 and RQ2 examine absolute performance and specific inference protocols, RQ3 investigates a deeper systemic question: \emph{do the observed G2U gains ($\Delta$) exhibit structured, predictable correlation patterns?}

\textbf{\textcolor{blue}{Takeaway 5: At the task level, perception-oriented tasks and reasoning-oriented tasks each form internally correlated clusters, but typically exhibit negative correlations with one another. At the model level, unified models built upon the same base model demonstrate strong behavioral correlations, whereas those merely sharing architectural similarities exhibit much weaker correlations.}}

To rigorously answer this, we analyze the relative offset $\Delta$ (unified model minus its strictly paired base model). This isolation explicitly removes the confounding variable of underlying backbone strength, leaving only the pure generation-induced capability shift.

\paragraph{Task-Level Correlation}
We compute pairwise Spearman correlations between subtasks by treating each subtask as a vector of $\Delta$ values across all evaluated models. The resulting heatmap (Fig.~\ref{fig:rq3_task_corr}) reveals that G2U gains form distinct cognitive clusters. Subtasks leaning towards rigorous structural reasoning (e.g., jigsaw, maze) exhibit strong positive mutual correlations. Similarly, perception-oriented tasks (e.g., fine-grained recognition) also show a certain degree of internal similarity. Conversely, reasoning-oriented subtasks frequently exhibit \emph{negative} correlations with perception-oriented tasks. This divergence provides compelling statistical evidence for the ``alignment tax'' discussed in RQ1: optimizing a model to benefit structural generation tasks often induces a representational trade-off that subtly degrades raw perceptual discrimination.

\paragraph{Model-Level Correlation}
We further compute correlations between models by treating each model as a vector of $\Delta$ across all subtasks. The model-level heatmaps (Fig.~\ref{fig:rq3_model_corr_s3} and \ref{fig:rq3_model_corr_s5}) unearth a striking grouping effect. Unified models constructed upon the exact same foundational Base VLM (Fig.~\ref{fig:rq3_model_corr_s3}) exhibit exceptionally strong correlations in their behavioral patterns, forming distinct high-correlation blocks. In stark contrast, models that merely share similar generative architectures (e.g., grouped by Autoregressive vs.\ Diffusion paradigms in Fig.~\ref{fig:rq3_model_corr_s5}) demonstrate much weaker correlation signatures. This yields a critical insight for future unified model design: the mechanism by which generative capacity transfers to understanding is predominantly dictated by the \emph{inherited foundational representations} (the VLM base), while merely adopting a similar generative architecture provides minimal behavioral consistency.
\subsection{Are Intermediate Visualizations Aligned and Helpful? (RQ4)}
\label{sec:rq4}

We diagnose the quality of explicitly generated intermediate images using the \textbf{RA} (Reasoning-to-Visual Alignment) and \textbf{AL} (Answer-to-Visual Alignment) metrics defined in Section~\ref{sec:metrics}. Detailed evaluation prompts and grading rubrics are provided in Appendix~\ref{app:ra-al}. Here, we analyze the alignment patterns across task categories and models to mechanically explain the GtA effectiveness observed in RQ2.

\textbf{\textcolor{blue}{Takeaway 6: High alignment fidelity is a necessary but not sufficient condition for G2U gains. While models draw faithfully in Perception tasks, explicit externalization is redundant; conversely, in structurally constrained domains, poor alignment leads to severe error propagation.}}

Two distinct patterns emerge from the diagnostic scores in Table~\ref{tab:rq4_alignment}, which perfectly corroborate our findings in RQ1 and RQ2. 

First, \textbf{recognition-oriented tasks} (e.g., Perception Reasoning) universally achieve the highest RA/AL scores across almost all models. This indicates that current unified models can reliably generate visually faithful intermediate artifacts when the required transformation is a relatively direct visual translation. However, as established in RQ2, this high fidelity does not translate into performance gains. Because perception tasks are heavily reliant on low-level visual discrimination rather than multi-step structural logic, drawing the intermediate state is computationally redundant. The models ``draw well'' but do not actually need the drawings to answer.

Second, \textbf{logic-intensive and structurally constrained tasks} (such as Math \& Geometry or specific physical simulations) exhibit significantly lower RA scores. This reflects the profound difficulty of faithfully externalizing complex, strict mathematical logic into a visual format. In these low-alignment settings, the generated auxiliary diagrams often omit critical constraints or distort geometric relationships. As a result, the subsequent answering module consumes a flawed visual premise (low AL), directly leading to the severe performance degradation (error propagation) observed in the GtA protocol.

Ultimately, the alignment diagnostic reveals that explicit generation is only beneficial when it operates in the ``sweet spot'': tasks where visual externalization is structurally necessary to reduce cognitive load (like Spatial Intelligence, where RA scores are moderate but useful) \emph{and} the model possesses sufficient alignment fidelity to produce trustworthy visual scaffolds.

\section{Conclusion and Future Work}

In this work, we introduce \textbf{UniG2U}, a large-scale diagnostic benchmark for systematically studying when and how generation helps understanding in unified multimodal models. Across 3,000 curated instances spanning seven cognitive regimes, we show that generation-to-understanding (G2U) gains are highly task-dependent and architecture-dependent. Unified models do not consistently outperform their base VLMs; instead, improvements emerge primarily in spatial transformation and state-tracking tasks, where generation-trained representations regularize structural reasoning. Moreover, explicit Generate-then-Answer generation does not universally surpass direct unified inference. While intermediate visual externalization can assist reasoning, it often introduces new failure modes when generated artifacts are inaccurate or misaligned. Our correlation analyses further reveal that G2U gains follow structured task- and model-level patterns shaped by representational priors inherited from base models.

Looking forward, advancing G2U requires stronger integration mechanisms between generation and reasoning. Future directions include representation-level alignment objectives, reliability-aware or self-verifying intermediate generation, and closed-loop agentic refinement that mitigates error propagation. Extending beyond image-only intermediates to structured or multimodal reasoning artifacts may further enhance robustness. Additionally, investigating scaling laws of G2U effects and designing causal evaluation protocols to isolate representation changes remain important open problems. We hope UniG2U provides a principled foundation for future research toward unified multimodal systems in which generation and understanding mutually reinforce rather than interfere with one another.



\bibliography{example_paper}
\bibliographystyle{plainnat}

\newpage
\appendix
\onecolumn
\appendix
\small 

\section{Performance Analysis of Unified Models with Frozen Backbones (Non-CoT Version)}
\label{Frozen_Backbones}
\begin{table}[ht]
\centering
\caption{\textbf{Direct (Non-CoT) results of frozen-backbone models on UniG2U.} All values are reported as percentages ($\times 100$). The Overall column represents the average accuracy across all task categories.}
\label{tab:frozen-direct-appendix}
\resizebox{\linewidth}{!}{%
\setlength{\tabcolsep}{4.5pt}
\renewcommand{\arraystretch}{1.1}
\scriptsize
\begin{tabular}{lcccccccc}
\toprule
\textbf{Model} & \textbf{Real-world} & \textbf{Geometry} & \textbf{Physics} & \textbf{Puzzles} & \textbf{Chart} & \textbf{Spatial} & \textbf{Perception} & \textbf{Overall} \\
& \textbf{Apps} & \textbf{Reasoning} & & \textbf{\& Games} & & \textbf{Intel.} & \textbf{Reasoning} & \\
\midrule
UniPic2(direct) & 36.50 & 23.50 & 36.00 & 24.48 & 67.00 & 24.40 & 39.43 & 33.68 \\
UniWorld-V1(direct) & 35.50 & 22.00 & 33.00 & 25.62 & 65.00 & 27.40 & 36.66 & 32.79 \\
STAR-7B(direct) & 32.50 & 19.00 & 30.50 & 26.25 & 86.00 & 25.40 & 38.56 & 33.50 \\
Uni-Video(direct) & 30.00 & 20.50 & 29.50 & 27.16 & 81.00 & 26.00 & 39.67 & 33.93 \\
\bottomrule
\end{tabular}%
}
\end{table}

Table~\ref{tab:frozen-direct-appendix} presents the Direct (Non-CoT) results of unified models that employ a frozen backbone. Theoretically, since the generative pathway is bypassed and the base vision-language model (VLM) backbone remains frozen during training, their performance should exactly match that of the base model (\textit{qwen2.5-vl-7b}). However, empirical results show measurable deviations. We attribute these discrepancies to several implementation-level factors. First, subtle weight differences can emerge if a unified model utilizes Exponential Moving Average (EMA) checkpoints rather than the exact original base model weights, leading to slight performance drifts. Second, variations in inference hyperparameters—such as maximum generation length, input image resolution scaling, and decoding strategies—can significantly alter the final predictions. Third, inconsistencies in the software environment, such as different versions of PyTorch, CUDA, or the transformers library, often introduce deterministic execution variances. Furthermore, differences in visual pre-processing pipelines, particularly image encoding formats (e.g., reading raw images via PIL versus applying JPEG compression), affect pixel-level inputs. Finally, operating in different numerical precisions can cause compounding errors. For instance, computing Rotary Position Embeddings (RoPE) in bfloat16 instead of float32 can accumulate numerical errors over long sequences, eventually shifting the output logits and altering the results.

\section{Ablation Study for Models without Edit Training}
\label{sec:appendix_ood_ablation}

\begin{table}[ht]
\centering
\small
\setlength{\tabcolsep}{4pt}
\renewcommand{\arraystretch}{1.15}

\begin{tabularx}{\linewidth}{>{\raggedright\arraybackslash}X *{8}{c}}
\toprule
Model &
Real &
Geometry &
Physics &
Puzzle &
Chart &
Spatial &
Perception &
Overall \\
\midrule

UAE (Direct) 
& 36.75 & 10.50 & 30.50 & 19.06 & 85.00 & 25.20 & 36.37 & 30.94 \\

UAE (GtA) 
& 23.00 & 7.50 & 32.50 & 20.33 & 71.00 & 22.80 & 34.68 
& 28.61 {\textcolor{blue}{(-2.33)}} \\

\midrule

X-Omni (Direct) 
& 21.50 & 9.50 & 29.50 & 26.24 & 83.00 & 27.00 & 35.31 & 30.86 \\

X-Omni (GtA) 
& 16.50 & 4.00 & 29.00 & 23.31 & 22.00 & 20.20 & 19.72 
& 19.87 {\textcolor{blue}{(-10.99)}} \\

\midrule

TokenFlow-XL (Direct) 
& 19.00 & 10.50 & 46.00 & 14.97 & 43.00 & 21.80 & 38.88 & 29.15 \\

TokenFlow-XL (GtA) 
& 14.50 & 9.50 & 26.50 & 10.06 & 38.00 & 21.80 & 34.28 
& 24.50 {\textcolor{blue}{(-4.65)}} \\

\bottomrule
\end{tabularx}

\caption{Sheet7 results (values $\times 100$). Blue numbers indicate the overall drop relative to the corresponding Direct setting.}
\label{tab:sheet7_scaled}
\end{table}

\begin{figure*}[ht]
\centering

\begin{subfigure}[t]{0.24\textwidth}
\centering
\includegraphics[width=\linewidth]{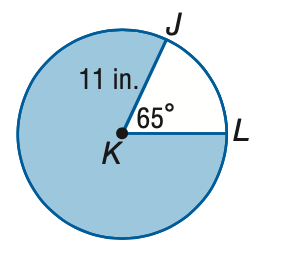}
\caption{Question Image}
\end{subfigure}\hfill
\begin{subfigure}[t]{0.24\textwidth}
\centering
\includegraphics[width=\linewidth]{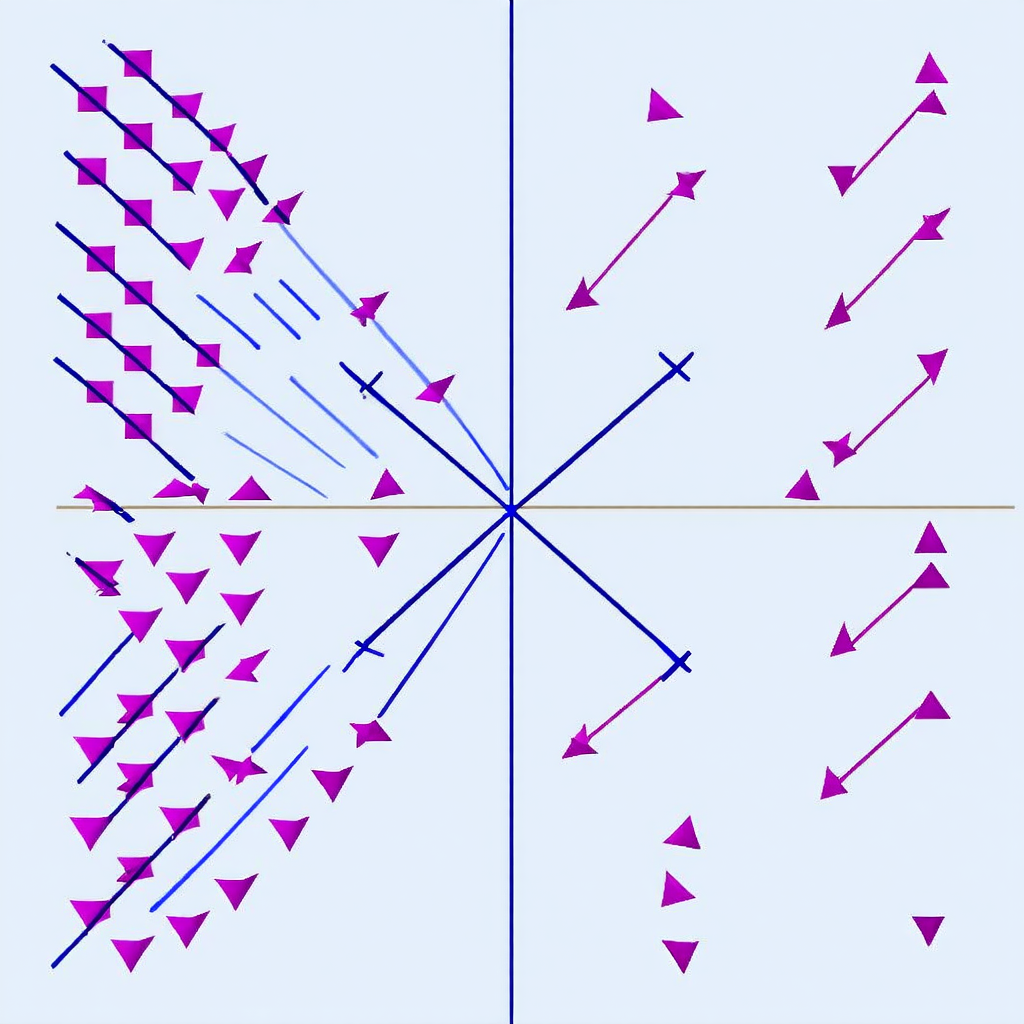}
\caption{UAE Generated Image}
\end{subfigure}\hfill
%
\begin{subfigure}[t]{0.24\textwidth}
\centering
\includegraphics[width=\linewidth]{appendix/x-omini-oringal.png}
\caption{Question Image}
\end{subfigure}\hfill
\begin{subfigure}[t]{0.24\textwidth}
\centering
\includegraphics[width=\linewidth]{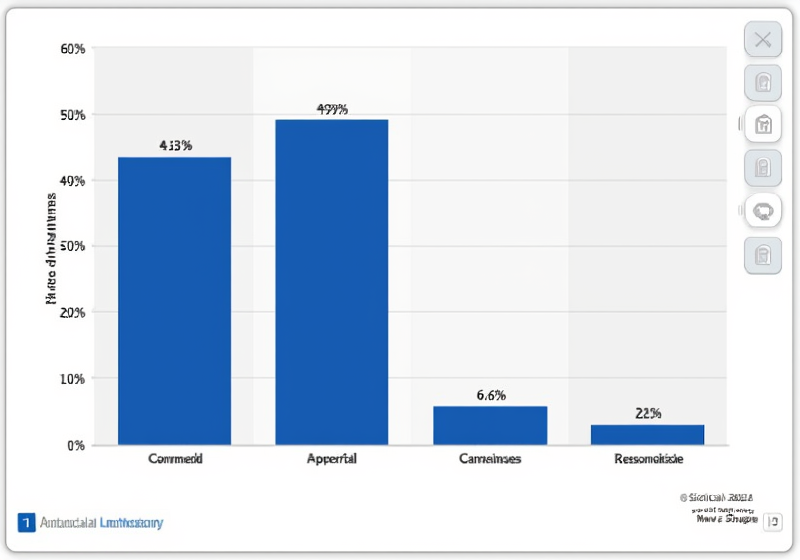}
\caption{X-Omni Generated Image}
\end{subfigure}

\caption{
Examples of erroneous generation from edit-untrained models.
}
\label{fig:irr_chart_know_quad11}
\end{figure*}

As discussed in Section~5, not all unified models evaluated in this work are trained with explicit image-to-image (edit) capabilities. For such models, Generate-then-Answer generation that requires intermediate image editing inherently falls outside their training distribution, effectively constituting an out-of-distribution (OOD) setting. Including their Generate-then-Answer results directly in the main comparison table could therefore introduce an unfair bias.

To better understand the effect of this OOD intermediate generation, we conduct additional ablation studies on a representative subset of edit-untrained models, with the results summarized in Table~\ref{tab:sheet7_scaled}. Specifically, we categorize their Generate-then-Answer evaluation into two alternative variants:

\paragraph{(1) OOD-Image Generation (UAE and X-Omni).}
Despite lacking edit training, the first two models are forced to generate an intermediate image under the standard Generate-then-Answer protocol (i.e., conditioned on the original input image). The generated image is then re-consumed for answering. This setting measures how severely forcing OOD intermediate generation degrades downstream reasoning performance.

\paragraph{(2) Prompt-Only Intermediate Generation (TokenFlow-XL).}
Instead of feeding the original input image into the intermediate generation stage, we restrict the final model to producing an auxiliary image conditioned strictly on the textual generation prompt (i.e., no image-to-image conditioning). The generated image is subsequently used as additional context for answering. This setting isolates the effect of prompt-level generation without invoking explicit edit-style image transformations.

\paragraph{Findings.}
As demonstrated in Table~\ref{tab:sheet7_scaled}, both Generate-then-Answer variants consistently lead to overall performance degradation compared to their Direct inference counterparts. Specifically, forcing OOD-Image generation causes overall performance drops of 2.33\% and 10.99\% for UAE and X-Omni, respectively, while relying on Prompt-Only generation leads to a 4.65\% drop for TokenFlow-XL. 

Furthermore, the qualitative examples presented in Figure~\ref{fig:irr_chart_know_quad11} corroborate these quantitative declines. Visual inspection reveals that the intermediate images produced by these edit-untrained models are often completely chaotic, structurally irrelevant, or fail to encode the necessary information. Instead of aiding the reasoning process, these misaligned generations actively mislead the subsequent answering stage.

Overall, these ablation results validate our decision to exclude the Generate-then-Answer scores of edit-untrained models from the main comparison table. They strongly emphasize that Generate-to-Understand (G2U) gains rely critically on explicit and training-aligned intermediate generation capabilities.

\section{Additional Result Analysis: Structured Subtask Gains}
\label{sec:appendix_subtask_analysis}

In this section, we analyze the complete experimental results (Table~\ref{tab:main-results-part1}) to identify subtasks where generation-to-understanding (G2U) gains are particularly evident.

While overall improvements are not universal, we observe that unified models demonstrate clearer and more repeatable gains on specific structured subtasks. In particular, \textbf{Spatial Intelligence} and \textbf{illusion-related perceptual reasoning} subtasks show more consistent positive offsets compared to base VLMs. These regimes require transformation-aware reasoning, geometric consistency, or resolving visually ambiguous patterns where generation-trained representations appear to regularize internal spatial structure.

For example, in the \textit{spatial} subtask under Perception Reasoning, multiple unified variants (e.g., UniWorld-V1, JavisGPT, uni-video, STAR-7B) achieve higher scores than the corresponding base model. Similarly, illusion-sensitive discrimination subtasks (e.g., logo-related and shape-sensitive categories) exhibit non-trivial improvements in certain unified configurations. These gains suggest that generation-induced representation coupling may enhance the model’s robustness to visually ambiguous or structurally complex inputs.

In contrast, purely recognition-heavy subtasks (e.g., icon\_scene, in\_scene) do not show systematic improvement and sometimes experience mild degradation. This further supports our central claim: G2U benefits are not uniform but emerge in transformation-centric or structurally constrained regimes.

Table~\ref{tab:main-results-part1} provides the detailed subtask-level results supporting this analysis.
\begin{table*}[!htbp]
\centering
\caption{\textbf{Absolute results on selected Illusion and Spatial subtasks.} 
All values are reported as percentages ($\times 100$). 
Baseline models (marked with $*$) are highlighted with gray background. 
\textbf{Overall} is computed as the average over the 5 shown subtasks, 
and the value in parentheses denotes the gap to the corresponding baseline within each model block. 
The best \textbf{Overall} within each block is \textbf{bolded}.}
\label{tab:shape_spatial_abs_overall_reduced}
\resizebox{\textwidth}{!}{%
\setlength{\tabcolsep}{4pt}
\renewcommand{\arraystretch}{1.08}
\scriptsize
\begin{tabular}{@{}lcccccc@{}}
\toprule
\multirow{2}{*}{\textbf{Model}} 
& \multicolumn{2}{c}{\textbf{Illusion}} 
& \multicolumn{3}{c}{\textbf{Spatial Intelligence}} 
& \multirow{2}{*}{\textbf{Overall} \textbf{($\Delta$)}} \\
\cmidrule(lr){2-3}\cmidrule(lr){4-6}
& \textbf{icon\_shape} 
& \textbf{in\_shape}
& \textbf{MSR} 
& \textbf{Attr. (Meas.)} 
& \textbf{Motion (Cam.)}
& \\
\midrule

\rowcolor{gray!12}
qwen2.5-vl-7b* & 5 & 10 & 25 & 20 & 29 & 17.80~(+$0.00$) \\
OmniGen2(direct) & 6 & 8 & 22 & 33 & 12 & 16.20~\textcolor{blue}{(-$1.60$)} \\
OmniGen2(GtA) & 6 & 11 & 29 & 33 & 18 & 19.40~\textcolor{red}{(+$1.60$)} \\
UniWorld-V1(GtA) & 4 & 10 & 27 & 22 & 21 & 16.80~\textcolor{blue}{(-$1.00$)} \\
OneCAT-3b(direct) & 2 & 8 & 32 & 30 & 24 & 19.20~\textcolor{red}{(+$1.40$)} \\
OneCAT-3b(GtA) & 1 & 7 & 39 & 28 & 3 & 15.60~\textcolor{blue}{(-$2.20$)} \\
JavisGPT(direct) & 12 & 13 & 16 & 24 & 19 & 16.80~\textcolor{blue}{(-$1.00$)} \\
uni-video(GtA) & 18 & 36 & 23 & 18 & 10 & \textbf{21.00~\textcolor{red}{(+$3.20$)}} \\
UniPic2(GtA) & 8 & 24 & 21 & 23 & 23 & 19.80~\textcolor{red}{(+$2.00$)} \\
STAR-7B(GtA) & 7 & 13 & 27 & 22 & 21 & 18.00~\textcolor{red}{(+$0.20$)} \\

\midrule

\rowcolor{gray!12}
Qwen2.5-VL-3B* & 6 & 9 & 22 & 32 & 18 & \textbf{17.40~(+$0.00$)} \\
UAE(direct) & 7 & 8 & 26 & 22 & 20 & 16.60~\textcolor{blue}{(-$0.80$)} \\
Ovis-U1(direct) & 7 & 10 & 12 & 28 & 18 & 15.00~\textcolor{blue}{(-$2.40$)} \\
Ovis-U1(GtA) & 5 & 15 & 27 & 27 & 11 & 17.00~\textcolor{blue}{(-$0.40$)} \\

\midrule

\rowcolor{gray!12}
yi-6B-vl* & 17 & 16 & 24 & 26 & 11 & \textbf{18.80~(+$0.00$)} \\
mio(direct) & 5 & 13 & 21 & 24 & 23 & 17.20~\textcolor{blue}{(-$1.60$)} \\
mio(GtA) & 15 & 16 & 32 & 24 & 22 & 21.80~\textcolor{red}{(+$3.00$)} \\

\midrule

\rowcolor{gray!12}
deepseek-v2* & 8 & 13 & 23 & 19 & 14 & 15.40~(+$0.00$) \\
Janus-Pro(direct) & 4 & 21 & 28 & 37 & 16 & 21.20~\textcolor{red}{(+$5.80$)} \\
AIA(direct) & 5 & 18 & 23 & 43 & 19 & \textbf{21.60~\textcolor{red}{(+$6.20$)}} \\

\midrule

\rowcolor{gray!12}
LLaDA-Instruct* & 0 & 8 & 21 & 41 & 11 & 16.20~(+$0.00$) \\
MMaDA(direct) & 1 & 3 & 32 & 27 & 20 & \textbf{16.60~\textcolor{red}{(+$0.40$)}} \\

\midrule

\rowcolor{gray!12}
Janus-1.3B* & 6 & 10 & 15 & 28 & 18 & 15.40~(+$0.00$) \\
FUDOK(direct) & 10 & 5 & 29 & 28 & 29 & \textbf{20.20~\textcolor{red}{(+$4.80$)}} \\

\midrule

\rowcolor{gray!12}
Qwen3-VL-8B* & 13 & 22 & 23 & 35 & 20 & \textbf{22.60~(+$0.00$)} \\
MammothModa2(direct) & 5 & 10 & 14 & 19 & 18 & 13.20~\textcolor{blue}{(-$9.40$)} \\

\midrule

\rowcolor{gray!12}
llava-hf* & 10 & 20 & 28 & 24 & 14 & \textbf{19.20~(+$0.00$)} \\
Emu3(direct) & 0 & 2 & 31 & 23 & 14 & 14.00~\textcolor{blue}{(-$5.20$)} \\

\midrule

\rowcolor{gray!12}
llava\_onevision* & 6 & 18 & 17 & 26 & 20 & 17.40~(+$0.00$) \\
X-Omni(direct) & 5 & 11 & 20 & 31 & 24 & 18.20~\textcolor{red}{(+$0.80$)} \\
bagel(direct) & 2 & 18 & 26 & 31 & 26 & \textbf{20.60~\textcolor{red}{(+$3.20$)}} \\
bagel(GtA) & 2 & 17 & 25 & 36 & 24 & 20.80~\textcolor{red}{(+$3.40$)} \\
TokenFlow-XL(direct) & 12 & 13 & 16 & 24 & 19 & 16.80~\textcolor{blue}{(-$0.60$)} \\
show-o2(direct) & 11 & 22 & 27 & 24 & 28 & 22.40~\textcolor{red}{(+$5.00$)} \\
show-o2(GtA) & 6 & 18 & 30 & 15 & 23 & 18.40~\textcolor{blue}{(+$1.00$)} \\
ILLUME+(direct) & 4 & 20 & 28 & 29 & 14 & 19.00~\textcolor{red}{(+$1.60$)} \\
ILLUME+(GtA) & 14 & 16 & 29 & 28 & 16 & 20.60~\textcolor{red}{(+$3.20$)} \\

\bottomrule
\end{tabular}%
}
\end{table*}

\section{Failure Case Taxonomy for Generation-Augmented Understanding}
\label{sec:appendix_failure_cases}

In this appendix, we present representative failure cases of generation-augmented reasoning.
We organize all examples under a unified three-way taxonomy based on the relationship
between generated diagrams and downstream reasoning performance.

\paragraph{Why failure cases matter.}
Generation-augmented understanding is often motivated by the intuition that ``drawing helps thinking'':
a faithful diagram can externalize latent structure, reduce working-memory burden, and provide
geometric/physical constraints that guide computation.
However, \emph{visual generation is itself a structured prediction problem}.
When the generated artifact is even slightly inconsistent with the underlying task, the downstream solver
may inherit or amplify those errors (e.g., by trusting a distorted geometry, an incorrect scale, or a spurious path).
Thus, it is important to characterize \emph{how} generation fails, not merely \emph{whether} an answer is wrong.

\paragraph{Selection protocol.}
We curate cases from diverse domains (geometry, physics, charts, and factual QA) to ensure the taxonomy
covers (i) tasks where diagrams are essential (geometry/physics),
(ii) tasks where diagrams can be helpful but must encode precise structure (charts),
and (iii) tasks where diagrams are unnecessary (pure textual knowledge).
For each case, we pair the original prompt with a model-generated visual output and
highlight the key mismatch that explains downstream failure.

\paragraph{Three criteria: validity, relevance, and utility.}
Across domains, successful generation must satisfy three requirements:
(i) \textbf{structural validity} (the diagram is internally consistent and respects physical/geometric rules),
(ii) \textbf{semantic relevance} (the content corresponds to the task entities and relations),
and (iii) \textbf{reasoning utility} (the diagram adds constraints or reduces ambiguity in a way that supports solving).
Our taxonomy maps naturally to violations of these criteria: Category~I breaks validity,
Category~II satisfies relevance but lacks utility, and Category~III fails relevance altogether.

\newcommand{\subcap}[1]{\caption*{\textbf{#1}}}
\newcommand{\smallnote}[1]{\vspace{2pt}\par{\footnotesize #1}}
\newcommand{\tightcaption}[1]{\caption{\vspace{-2pt}#1\vspace{-6pt}}}

\paragraph{Category I: Capability Failure.}
The model fails to generate structurally correct or quantitatively valid diagrams.
Rendered visuals are geometrically distorted, structurally inconsistent, or physically invalid,
making downstream reasoning unreliable or impossible.
This category is the most direct failure mode: even if the image appears ``diagram-like'',
it violates invariants required by the task.

\smallnote{\textbf{Common symptoms.}}
\vspace{-2pt}
\begin{itemize}\setlength{\itemsep}{1pt}\setlength{\topsep}{2pt}
\item \emph{Geometric inconsistency:} angles/parallel lines/circle tangency implied by the prompt are not preserved.
\item \emph{Quantitative distortion:} lengths, slopes, or axis scales are visually inconsistent with stated values.
\item \emph{Physical invalidity:} forces/velocities are drawn with wrong directions or incompatible magnitudes.
\item \emph{Missing constraints:} key elements (labels, right-angle marks, reference points) are omitted or misplaced.
\end{itemize}

\smallnote{\textbf{Why it breaks reasoning.}}
Even a strong textual solver may implicitly treat the diagram as ground truth.
When the generated visual is invalid, the solver can be pushed toward incorrect intermediate computations
(e.g., wrong triangle similarity, wrong area under a curve, wrong cross-product direction), or may be forced to ignore
the diagram entirely, losing the intended benefit of externalization.

\begin{figure*}[t]
\centering
\begin{minipage}[t]{0.24\textwidth}
\centering
\includegraphics[width=\linewidth]{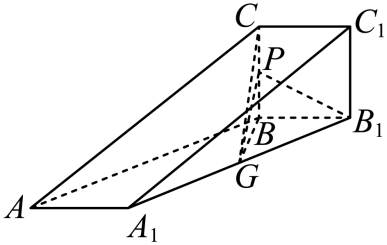}
\subcap{Original}
\smallnote{Find the length of $BP$}
\end{minipage}\hfill
\begin{minipage}[t]{0.24\textwidth}
\centering
\includegraphics[width=\linewidth]{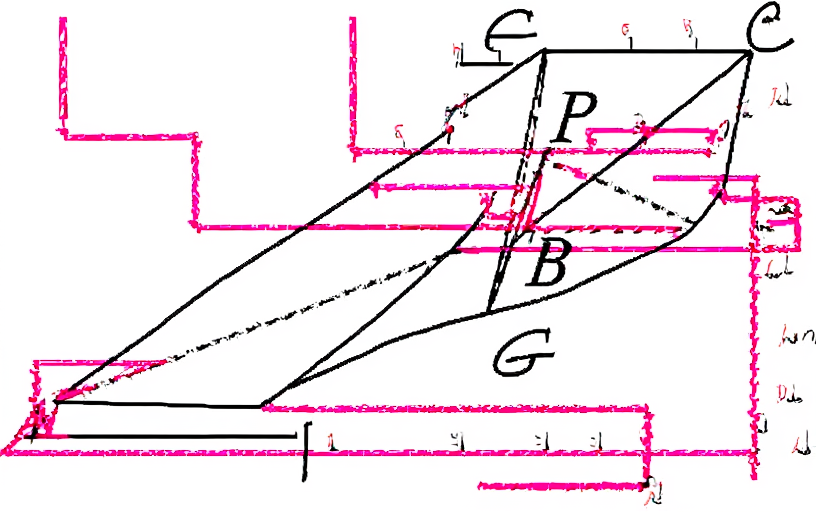}
\subcap{Generated}
\smallnote{Bagel}
\end{minipage}\hfill
\begin{minipage}[t]{0.24\textwidth}
\centering
\includegraphics[width=\linewidth]{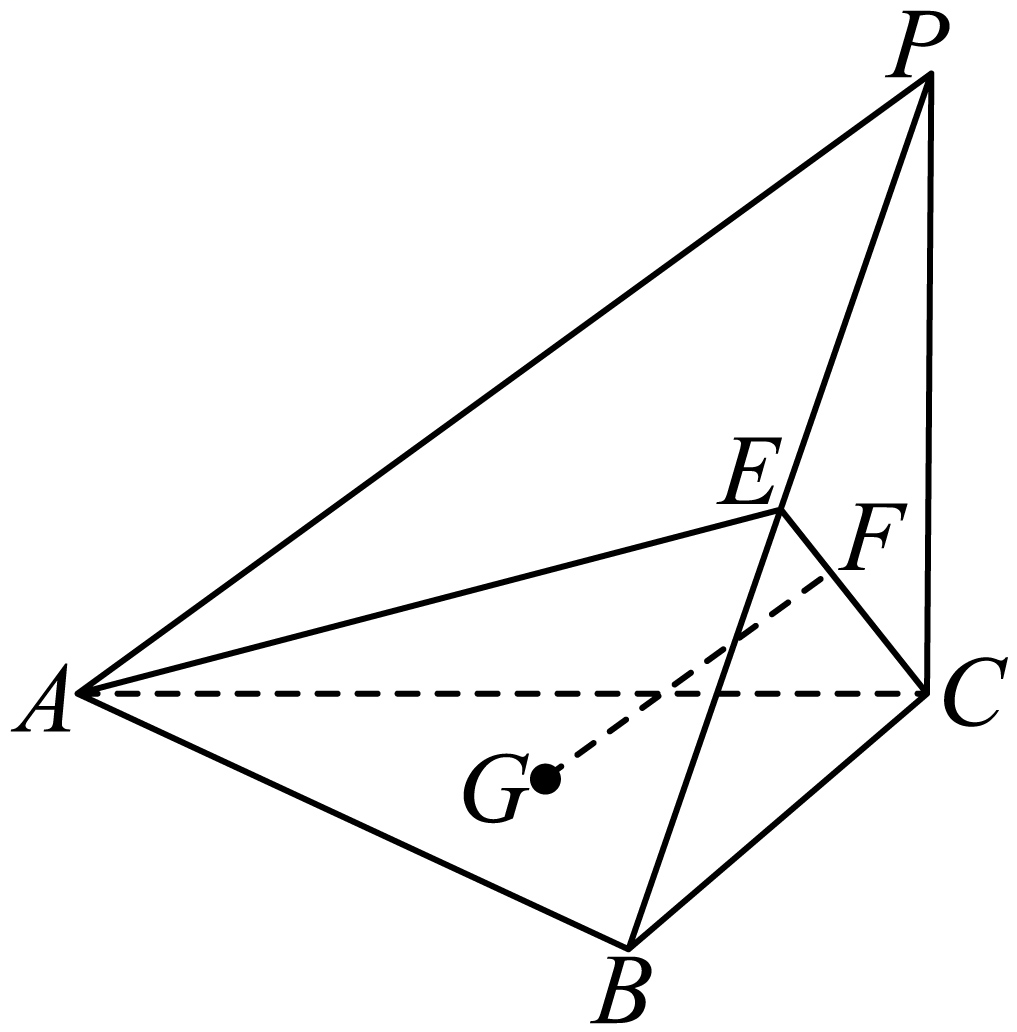}
\subcap{Original}
\smallnote{When the length of $PC$ is such that the dihedral angle $E - AC - B$ is 60°, determine its value}
\end{minipage}\hfill
\begin{minipage}[t]{0.24\textwidth}
\centering
\includegraphics[width=\linewidth]{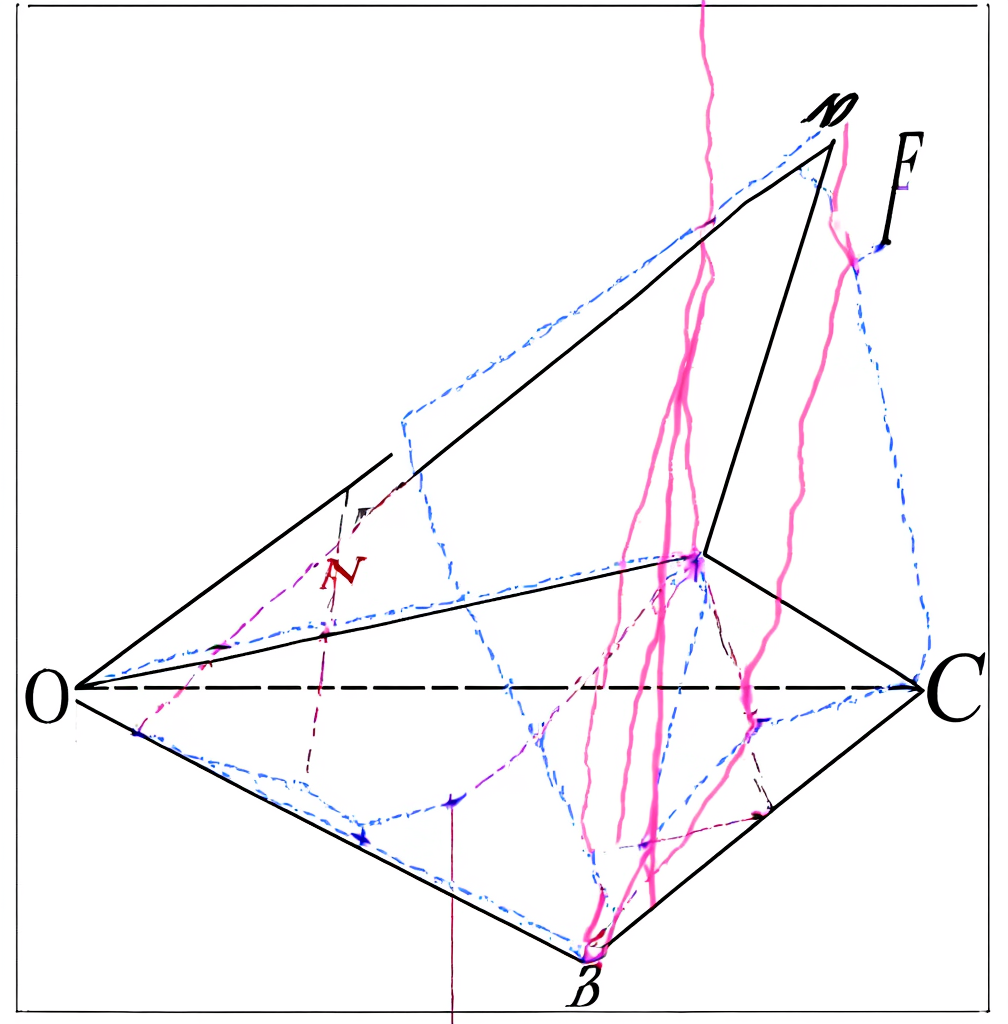}
\subcap{Generated}
\smallnote{Bagel}
\end{minipage}
\tightcaption{Capability failure. The model produces geometrically inconsistent or physically invalid diagrams, preventing correct downstream reasoning.}
\label{fig:cap_geo_solid_quad}
\end{figure*}

Figure~\ref{fig:cap_geo_solid_quad} shows two geometry instances where the generated diagrams violate the required geometric/physical constraints.

\begin{figure*}[t]
\centering
\begin{minipage}[t]{0.24\textwidth}
\centering
\includegraphics[width=\linewidth]{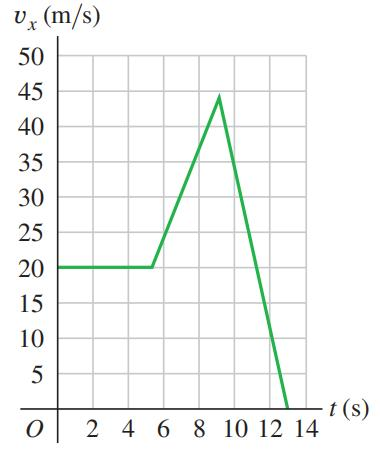}
\subcap{Original (Case 1)}
\smallnote{How far in first $13\,\mathrm{s}$?
\emph{Options:} A 320m, B 230m, C 280m, D 300m.}
\end{minipage}\hfill
\begin{minipage}[t]{0.24\textwidth}
\centering
\includegraphics[width=\linewidth]{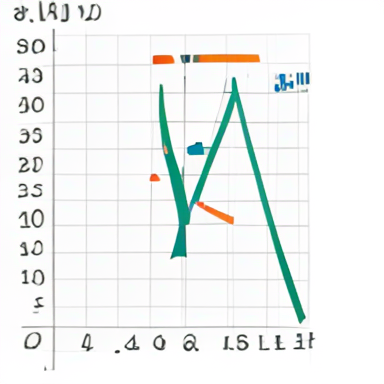}
\subcap{Generated (Case 1)}
\smallnote{STAR-7B.}
\end{minipage}\hfill
\begin{minipage}[t]{0.24\textwidth}
\centering
\includegraphics[width=\linewidth]{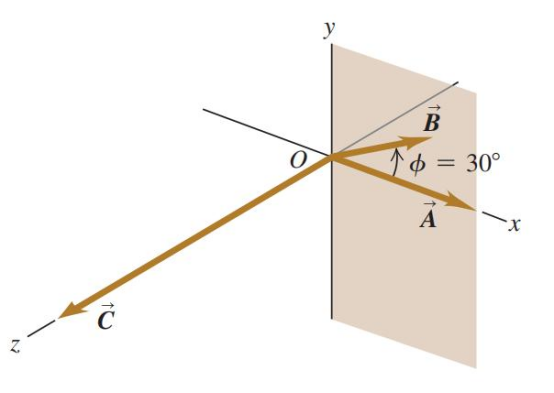}
\subcap{Original (Case 2)}
\smallnote{$\vec{A}$: mag 6 along $+x$; $\vec{B}$: mag 4 in $xy$.
\emph{Options:} A $15\hat{k}$, B $12\hat{k}$, C $8\hat{k}$, D $18\hat{k}$.}
\end{minipage}\hfill
\begin{minipage}[t]{0.24\textwidth}
\centering
\includegraphics[width=\linewidth]{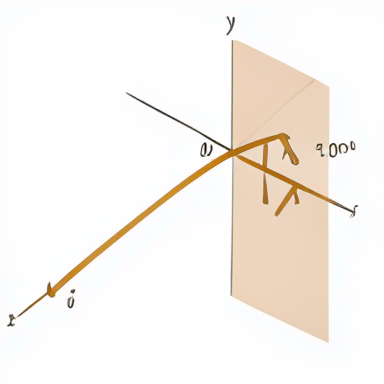}
\subcap{Generated (Case 2)}
\smallnote{STAR-7B.}
\end{minipage}
\tightcaption{Capability failure in physics reasoning. Generated diagrams fail to preserve quantitative/structural relations required for correct integration- or geometry-based computation.}
\label{fig:cap_phy_quad}
\end{figure*}

Figure~\ref{fig:cap_phy_quad} presents two physics cases where the generated visuals fail to preserve quantitative/structural relations needed for integration- or geometry-based computation.

\smallnote{\textbf{Implication.}}
Category~I indicates a bottleneck in the generator's \emph{diagram grammar}:
models may learn surface styles (lines, arrows, labels) without learning domain invariants.
A practical mitigation is to introduce lightweight validation or self-checking
(e.g., verify right angles, axis monotonicity, connectivity constraints) before passing the diagram to the solver,
or to route such instances back to a pure-text solution when validity is uncertain.

\paragraph{Category II: Surface-Relevant but Non-Utility Generation.}
The model generates diagrams that are semantically related to the prompt,
yet they add no constraints or computational benefit for solving.
In these cases, the generation is ``on-topic'' but \emph{not instrumental}:
the diagram either restates what is already explicit in text, or fails to encode the specific structure needed.

\smallnote{\textbf{Common symptoms.}}
\vspace{-2pt}
\begin{itemize}\setlength{\itemsep}{1pt}\setlength{\topsep}{2pt}
\item \emph{Correct but redundant:} the diagram matches the description but does not reduce uncertainty.
\item \emph{Missing operational structure:} key relations for computation (e.g., graph scales, maze connectivity) are absent.
\item \emph{Aesthetic completion:} the model fills the page with plausible visuals rather than task-sufficient constraints.
\end{itemize}

\smallnote{\textbf{Why it fails to help.}}
Utility is not guaranteed by relevance: even a perfectly relevant illustration can be useless if it does not support the
specific operation needed (e.g., reading off a slope, tracing a path, or identifying a computable sub-structure).
This failure mode is particularly important because it can \emph{waste} generation budget without improving accuracy,
and may even distract downstream reasoning by introducing extra, non-actionable details.

\begin{figure*}[t]
\centering
\begin{minipage}[t]{0.24\textwidth}
\centering
\includegraphics[width=\linewidth]{appendix/category2_1_origin.png}
\subcap{Original (Geom)}
\end{minipage}\hfill
\begin{minipage}[t]{0.24\textwidth}
\centering
\includegraphics[width=\linewidth]{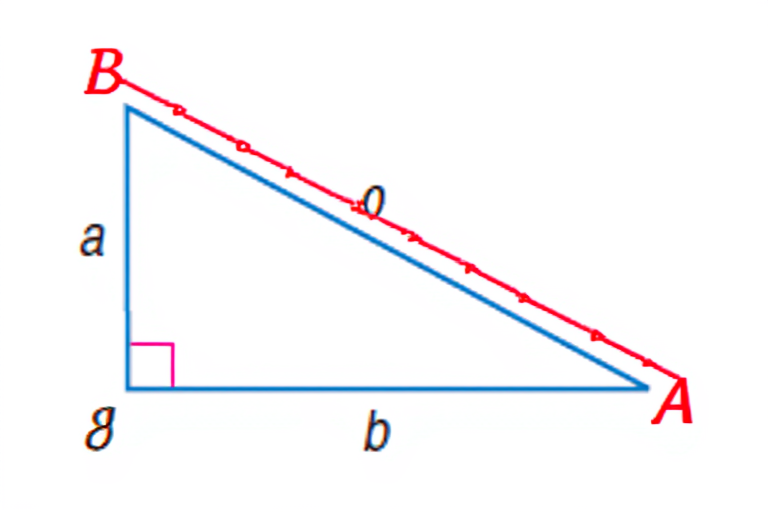}
\subcap{Generated (Geom)}
\smallnote{Surface-relevant but adds no utility(Bagel).}
\end{minipage}\hfill
\begin{minipage}[t]{0.24\textwidth}
\centering
\includegraphics[width=\linewidth]{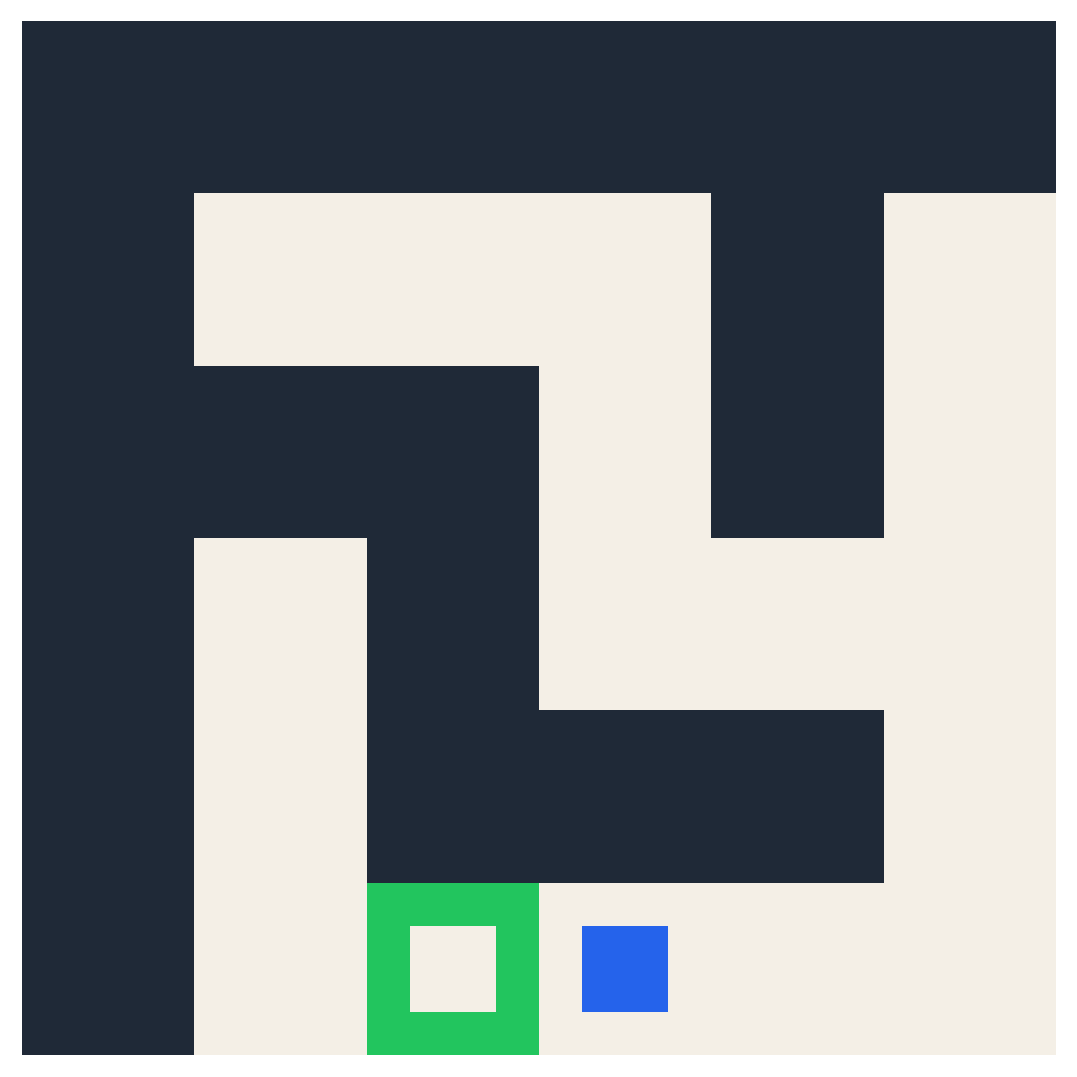}
\subcap{Original (Maze)}
\end{minipage}\hfill
\begin{minipage}[t]{0.24\textwidth}
\centering
\includegraphics[width=\linewidth]{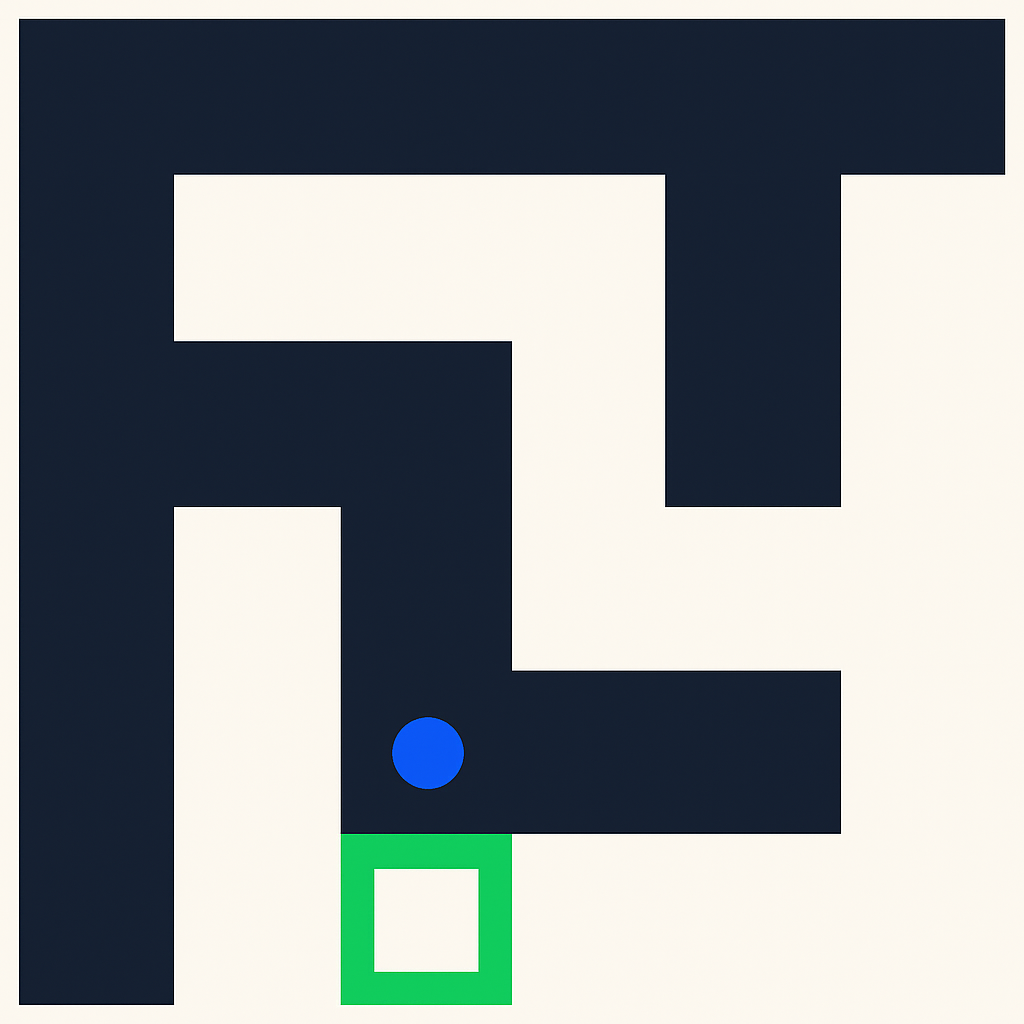}
\subcap{Generated (Maze)}
\smallnote{No valid connectivity/path(GPT-4o + GPT-image).}
\end{minipage}
\tightcaption{Surface-relevant but non-utility generation. Outputs are related to the prompt but provide no additional constraints or computational benefit.}
\label{fig:surf_geom_maze_quad}
\end{figure*}

Figure~\ref{fig:surf_geom_maze_quad} shows cases where the output looks on-topic but does not help: one is correct yet uninformative, and the other lacks valid connectivity/path structure.

\begin{figure*}[t]
\centering
\begin{minipage}[t]{0.24\textwidth}
\centering
\includegraphics[width=\linewidth]{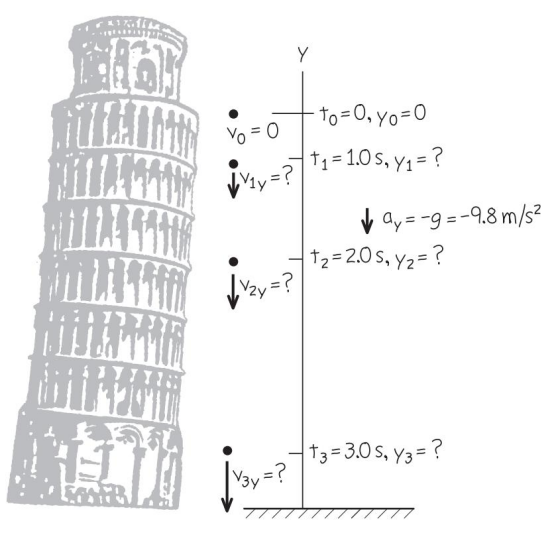}
\subcap{Original (Case 1)}
\smallnote{Coin dropped from rest. Position after $3.0\,\mathrm{s}$?
\emph{Options:} A -35m, B -44m, C -65m, D -51m.}
\end{minipage}\hfill
\begin{minipage}[t]{0.24\textwidth}
\centering
\includegraphics[width=\linewidth]{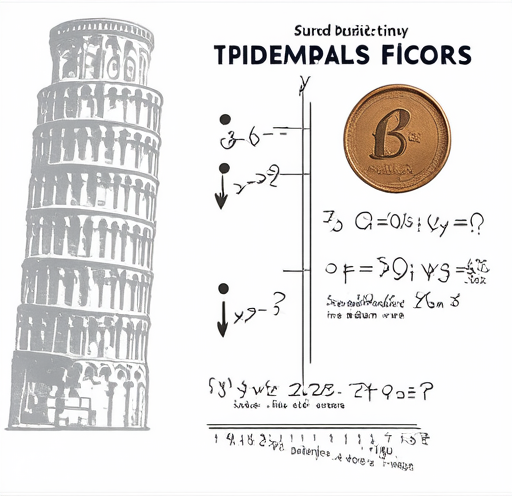}
\subcap{Generated (Case 1)}
\smallnote{UniPic2-Metaquery-9B.}
\end{minipage}\hfill
\begin{minipage}[t]{0.24\textwidth}
\centering
\includegraphics[width=\linewidth]{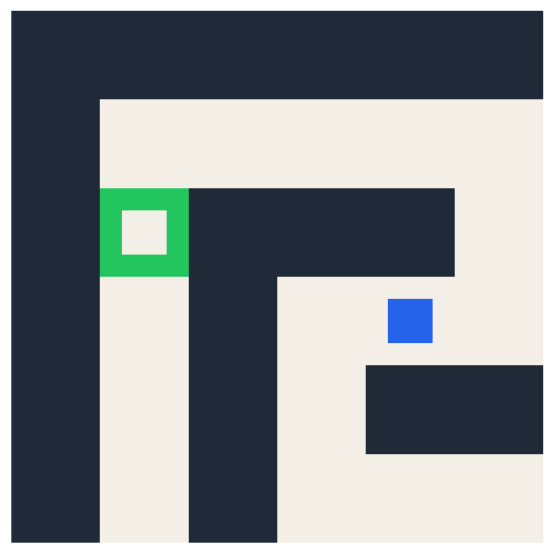}
\subcap{Original (Case 2)}
\end{minipage}\hfill
\begin{minipage}[t]{0.24\textwidth}
\centering
\includegraphics[width=\linewidth]{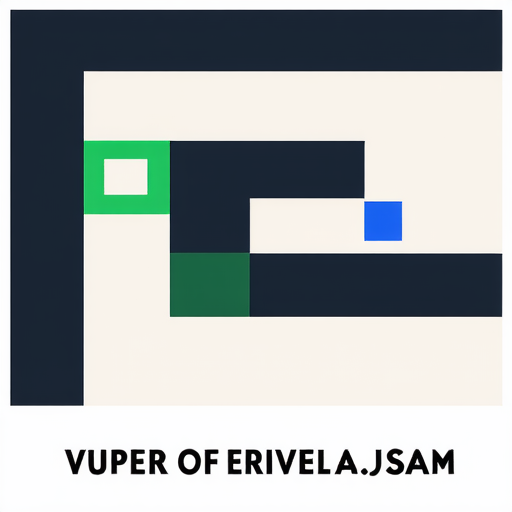}
\subcap{Generated (Case 2)}
\smallnote{UniPic2-Metaquery-9B.}
\end{minipage}
\tightcaption{Surface-triggered generation. The model draws a ``coin'' due to lexical cues, while the task requires kinematic computation; the generated visuals provide little to no reasoning utility.}
\label{fig:surf_freefall_quad}
\end{figure*}

Figure~\ref{fig:surf_freefall_quad} illustrates surface-triggered generation in kinematics: lexical cues (e.g., ``coin'') induce drawing, but the task requires computation, so the visuals contribute little to reasoning.

\smallnote{\textbf{Implication.}}
Category~II suggests that ``draw something relevant'' is an insufficient objective.
For G2U, the generator must be \emph{goal-conditioned}: it should produce the minimal structure that makes the
answering computation easier (e.g., annotated axes, explicit path graph, marked equal angles),
rather than a generic illustration.
This also motivates \emph{budget-aware routing}: when predicted utility is low, skipping generation can improve
value-per-token without sacrificing accuracy.

\paragraph{Category III: Irrelevant Generation.}
The generated image is not meaningfully related to the task, indicating complete
generation--task misalignment.
Here, generation is triggered when it should not be (e.g., factual QA),
or the model produces a visual that does not encode the required structured content (e.g., ChartQA).

\smallnote{\textbf{Common symptoms.}}
\vspace{-2pt}
\begin{itemize}\setlength{\itemsep}{1pt}\setlength{\topsep}{2pt}
\item \emph{Mode confusion:} the model treats a purely textual query as an invitation to draw.
\item \emph{Structure loss:} the output resembles an image but lacks the schema needed (bars/axes/labels) to be a chart.
\item \emph{Hallucinated content:} visuals introduce unrelated entities, styles, or scenes not grounded in the prompt.
\end{itemize}

\smallnote{\textbf{Why it hurts.}}
Irrelevant generation is harmful beyond wasted compute: it can shift the solver's attention to spurious cues,
reduce trust in the system, and complicate debugging (the output looks ``confident'' but is unrelated).
In knowledge QA, the correct behavior is often to answer directly; in chart reasoning, the correct behavior is to
preserve structured data rather than produce decorative images.

\begin{figure*}[t]
\centering
\begin{minipage}[t]{0.24\textwidth}
\centering
\includegraphics[width=\linewidth]{appendix/category3_1_origin.png}
\subcap{Original (ChartQA)}
\end{minipage}\hfill
\begin{minipage}[t]{0.24\textwidth}
\centering
\includegraphics[width=\linewidth]{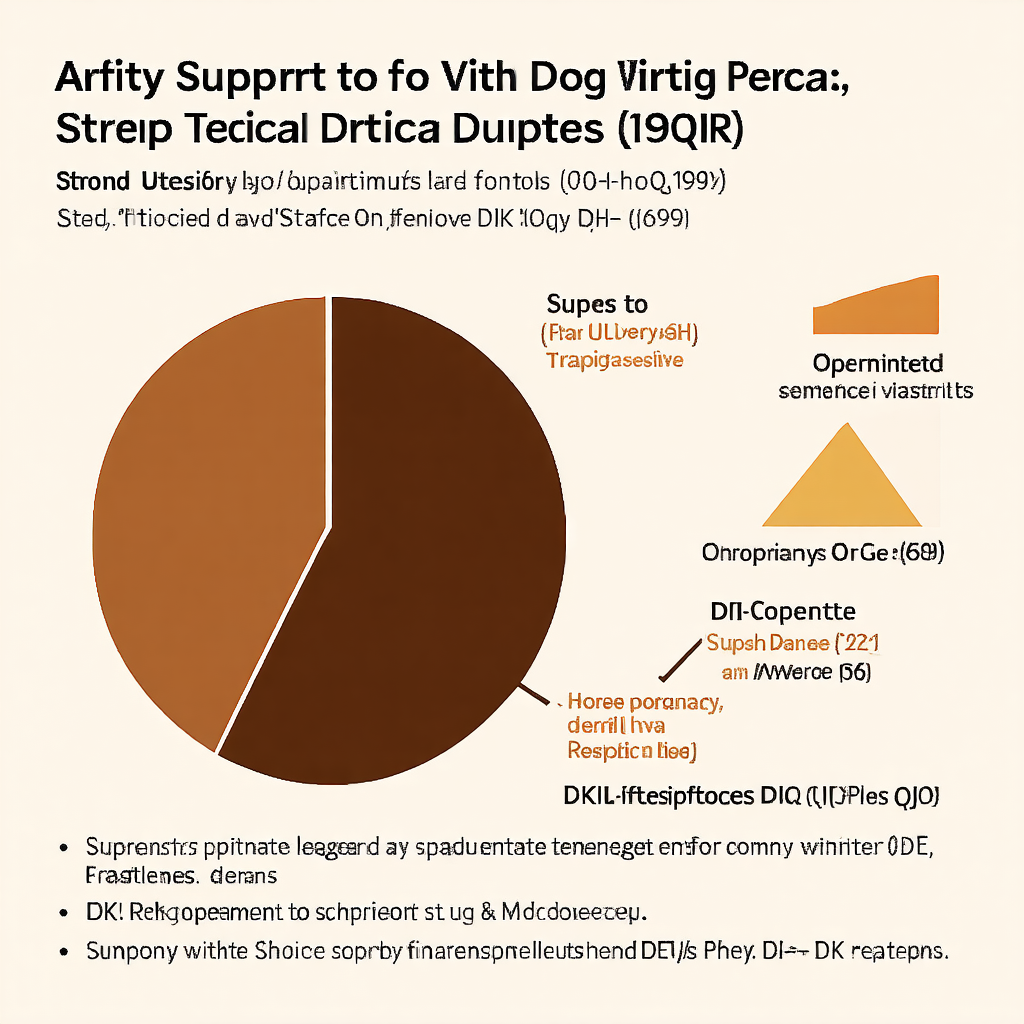}
\subcap{Generated (ChartQA)}
\smallnote{Not a structured chart(GPT-4o + GPT-image).}
\end{minipage}\hfill
\begin{minipage}[t]{0.24\textwidth}
\centering
\includegraphics[width=\linewidth]{appendix/knowledge_case1_original.png}
\subcap{Original (Know, C1)}
\smallnote{What is the value of Czechia? One word.}
\end{minipage}\hfill
\begin{minipage}[t]{0.24\textwidth}
\centering
\includegraphics[width=\linewidth]{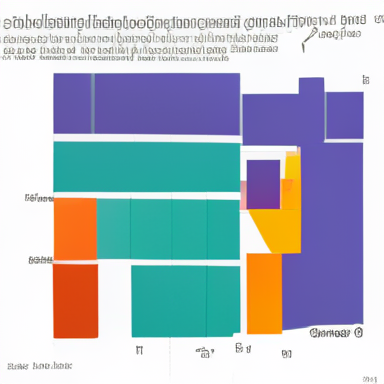}
\subcap{Generated (Know, C1)}
\smallnote{STAR-7B.}
\end{minipage}
\tightcaption{Irrelevant generation. Outputs are misaligned with the task: chart generations fail to encode structured data, and factual QA triggers unnecessary visual generation.}
\label{fig:irr_chart_know_quad}
\end{figure*}

Figure~\ref{fig:irr_chart_know_quad} shows irrelevant outputs in chart and factual QA settings: chart generations fail to encode structured data, while factual questions unnecessarily trigger visual generation.

\begin{figure*}[t]
\centering
\begin{minipage}[t]{0.24\textwidth}
\centering
\includegraphics[width=\linewidth]{appendix/knowledge_case2_original.png}
\subcap{Original (Know, C2)}
\smallnote{What is the value of Czechia? One word.}
\end{minipage}\hfill
\begin{minipage}[t]{0.24\textwidth}
\centering
\includegraphics[width=\linewidth]{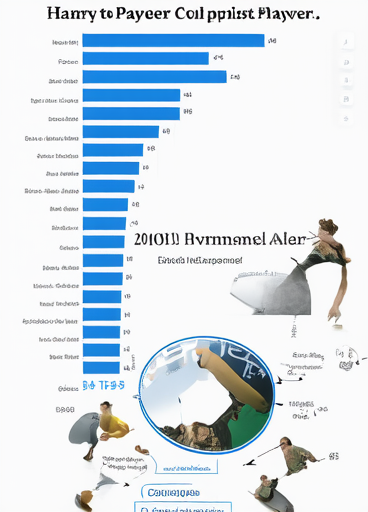}
\subcap{Generated (Know, C2)}
\smallnote{UniPic2-Metaquery-9B.}
\end{minipage}\hfill
\begin{minipage}[t]{0.24\textwidth}
\centering
\includegraphics[width=\linewidth]{appendix/category3_2_origin.png}
\subcap{Original (Case 3)}
\end{minipage}\hfill
\begin{minipage}[t]{0.24\textwidth}
\centering
\includegraphics[width=\linewidth]{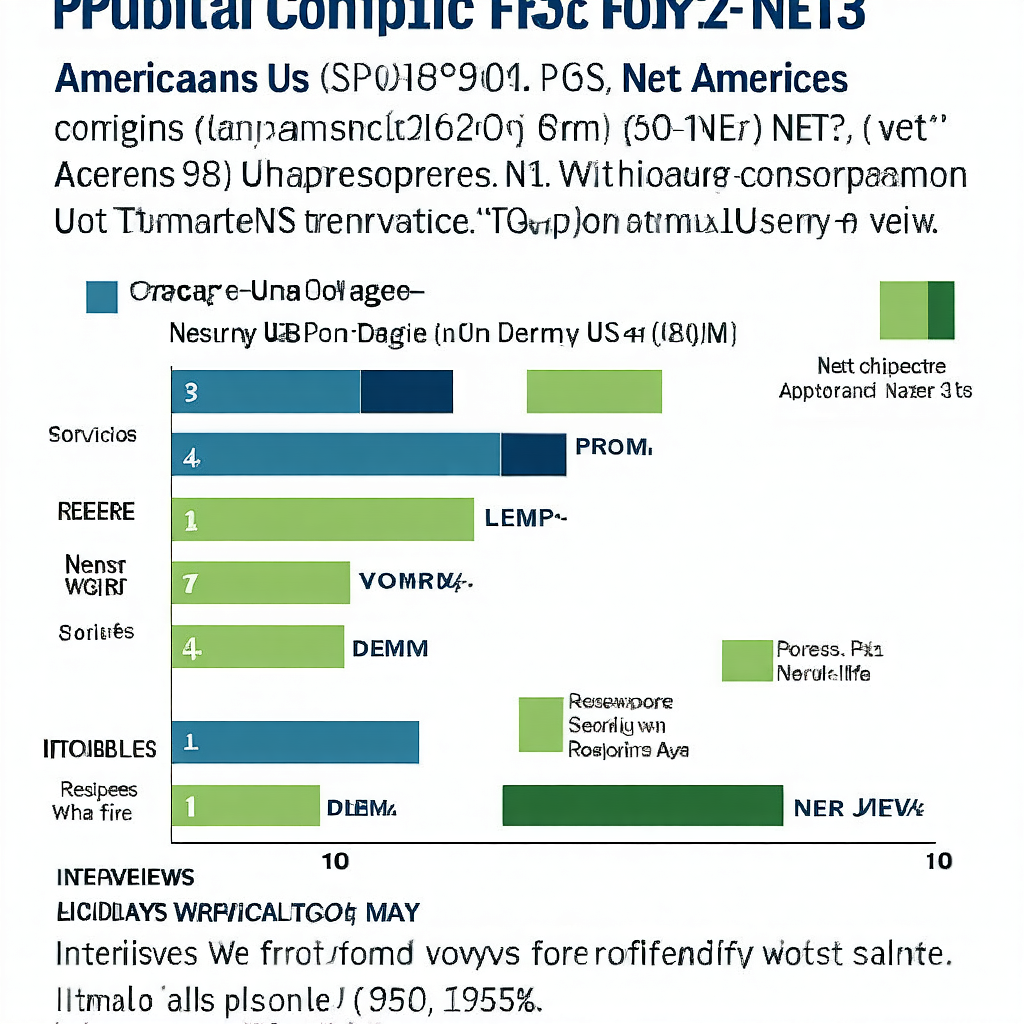}
\subcap{Generated (Case 3)}
\smallnote{Bagel}
\end{minipage}
\tightcaption{Irrelevant generation in textual knowledge tasks. The model produces visual content despite the task being purely factual, indicating generation--task misalignment.}
\label{fig:irr_qa_quad}
\end{figure*}

Figure~\ref{fig:irr_qa_quad} provides additional knowledge-QA examples where the model produces visual content despite the task being purely factual.

\smallnote{\textbf{Practical lesson for G2U system design.}}
\vspace{-2pt}
\begin{enumerate}\setlength{\itemsep}{1pt}\setlength{\topsep}{2pt}
\item \textbf{Validate} when diagrams are intended to be constraints (Category~I risk).
\item \textbf{Predict utility} and skip redundant illustrations (Category~II risk).
\item \textbf{Route by modality} and avoid image generation for pure text tasks (Category~III risk).
\end{enumerate}


\section{Uni-MMMU Task Design}
\label{sec:uni-mmmu}

Uni-MMMU consists of three task types: Jigsaw, Maze, and Sliding Puzzle. Each task evaluates different aspects of visual reasoning through interleaved image-text generation. ( following case from Model - Nano Banana )

\subsection{Jigsaw Completion Task}

\subsubsection{Task Description}

\textbf{Input:}
\begin{itemize}
    \item Reference image: 2×2 grid with bottom-right cell hidden
    \item Candidate 0: First candidate patch
    \item Candidate 1: Second candidate patch
\end{itemize}

\textbf{Task:} Select which candidate correctly completes the reference image.

\textbf{Output:} \verb|<FINAL_ANSWER_JSON>{"choice": 0 or 1, "rationale": "..."}|

\textbf{Metrics:} \texttt{jigsaw\_text\_acc} (binary accuracy)

\textbf{Dataset:} 100 samples

\subsubsection{Visual CoT Generation Flow}

\textbf{Step 1 - Input Images:}

\begin{figure}[h]
\centering
\begin{subfigure}{0.3\textwidth}
    \includegraphics[width=\textwidth]{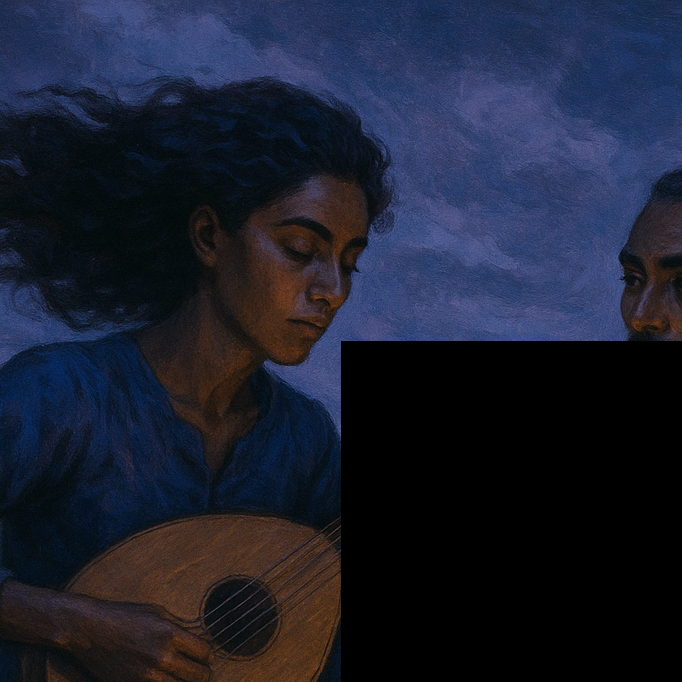}
    \caption{Reference (bottom-right hidden)}
\end{subfigure}
\hfill
\begin{subfigure}{0.3\textwidth}
    \includegraphics[width=\textwidth]{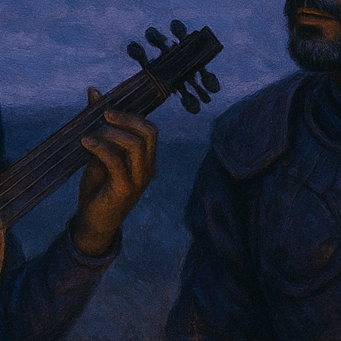}
    \caption{Candidate 0}
\end{subfigure}
\hfill
\begin{subfigure}{0.3\textwidth}
    \includegraphics[width=\textwidth]{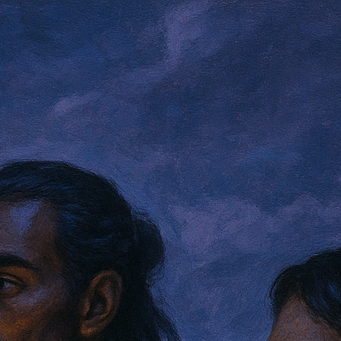}
    \caption{Candidate 1}
\end{subfigure}
\caption{Jigsaw task input images}
\end{figure}

\textbf{Step 2 - Generate Completion with Candidate 0:}

Model generates a 2×2 image with Candidate 0 placed in the bottom-right cell. The other three cells remain pixel-identical to the reference.

\begin{figure}[h]
\centering
\includegraphics[width=0.4\textwidth]{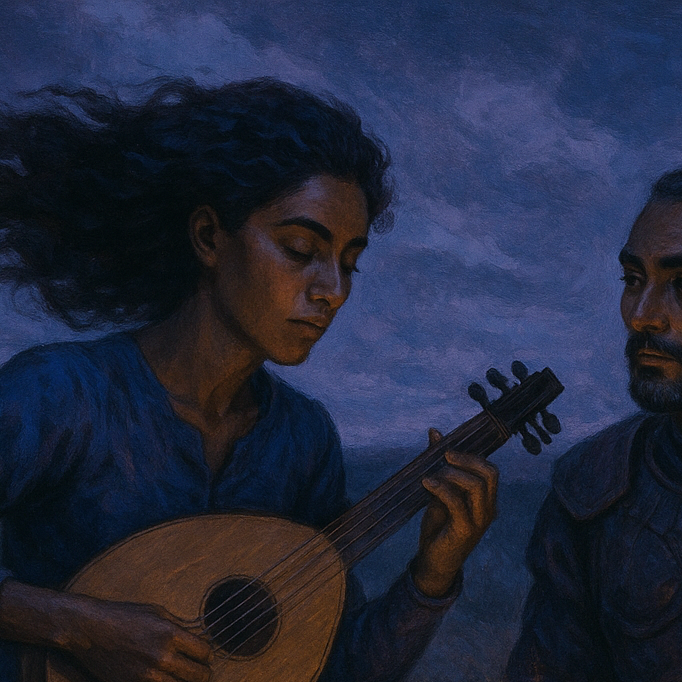}
\caption{Generated image: Completion with Candidate 0}
\end{figure}

\textbf{Step 3 - Generate Completion with Candidate 1:}

Model generates a 2×2 image with Candidate 1 placed in the bottom-right cell. The other three cells remain pixel-identical to the reference.

\begin{figure}[h]
\centering
\includegraphics[width=0.4\textwidth]{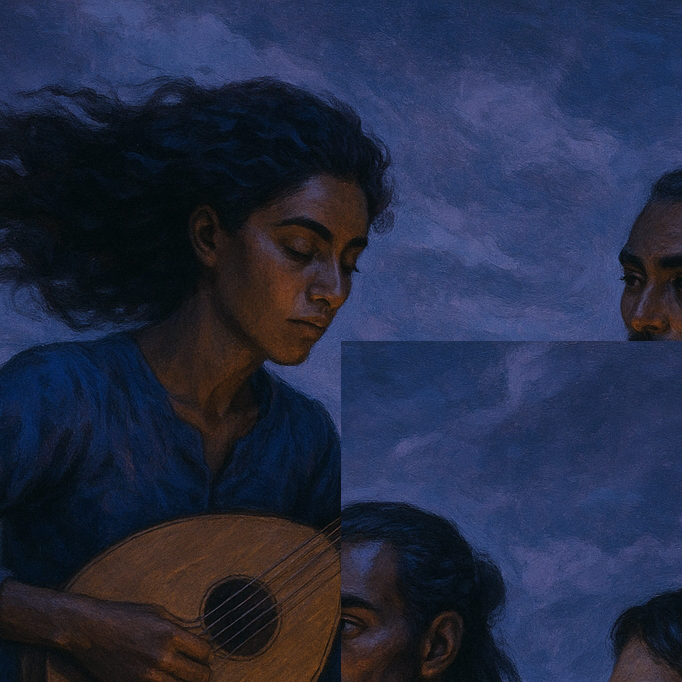}
\caption{Generated image: Completion with Candidate 1}
\end{figure}

\textbf{Step 4 - Generate Final Answer:}

Model analyzes both generated completions and outputs:
\begin{verbatim}
<FINAL_ANSWER_JSON>
{"choice": 0, "rationale": "Candidate 0 shows better seam continuity"}
</FINAL_ANSWER_JSON>
\end{verbatim}

\subsection{Maze Solving Task}

\subsubsection{Task Description}

\textbf{Input:} Maze image with:
\begin{itemize}
    \item Black squares: walls (impassable)
    \item White squares: paths (walkable)
    \item Blue dot: starting position
    \item Green frame: goal region
\end{itemize}

\textbf{Task:} Find a sequence of moves from start to goal.

\textbf{Constraints:} Legal moves: up, down, left, right; one cell per step; no diagonals; no wall crossing.

\textbf{Output:} \verb|<ANSWER_JSON>["right", "down", ...]|

\textbf{Metrics:} \texttt{maze\_text\_exact}, \texttt{maze\_text\_frame\_acc}

\textbf{Dataset:} 100 samples

\subsubsection{Visual CoT Generation Flow}

\textbf{Step 1 - Input Image:}

\begin{figure}[h]
\centering
\includegraphics[width=0.5\textwidth]{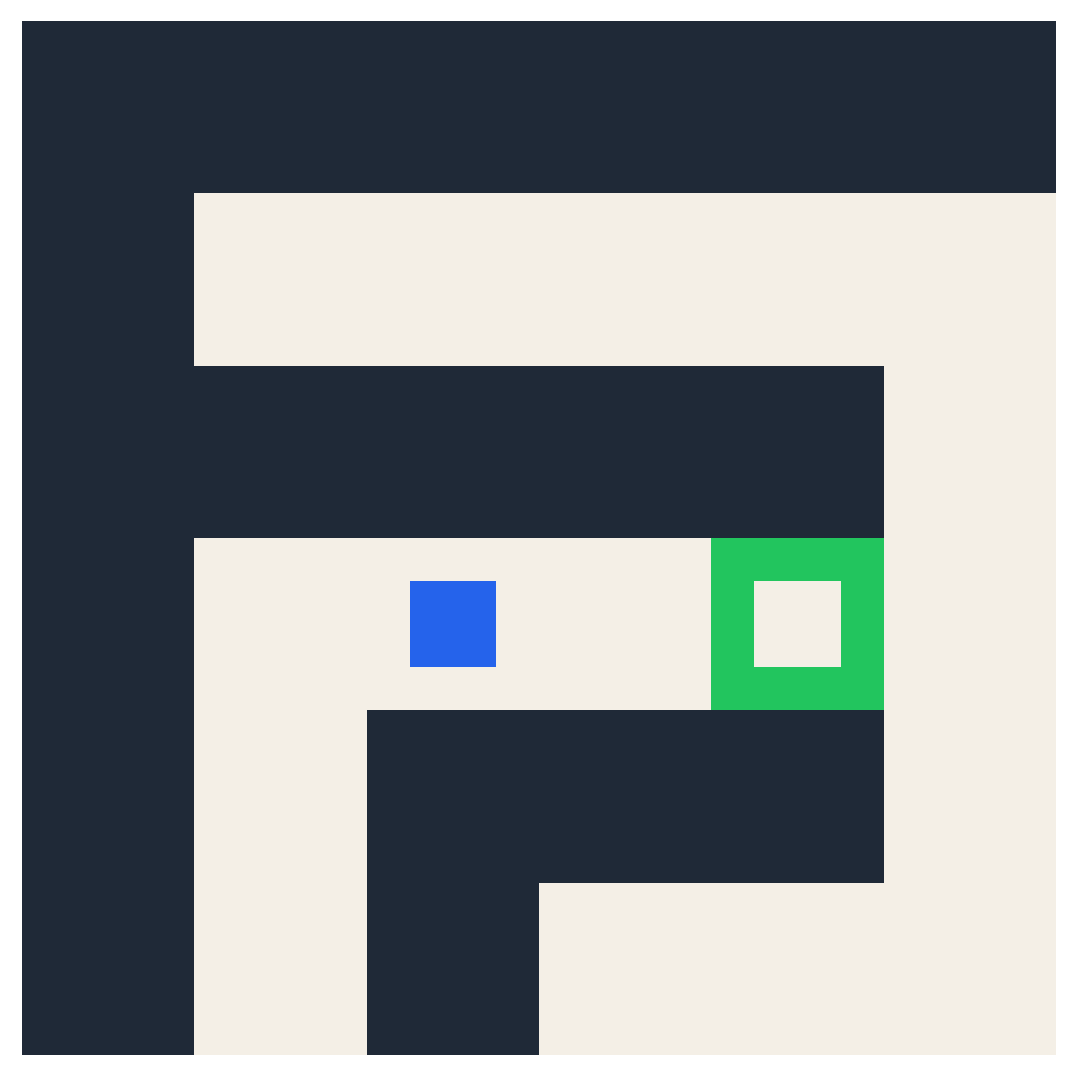}
\caption{Initial maze state (blue dot at start, green frame marks goal)}
\end{figure}

\textbf{Step 2 - Generate Step-by-Step Images:}

For each move in the solution, the model generates an image showing the maze state after that move. Only the blue dot position changes; all other elements remain identical.

\begin{figure}[h]
\centering
\begin{subfigure}{0.45\textwidth}
    \includegraphics[width=\textwidth]{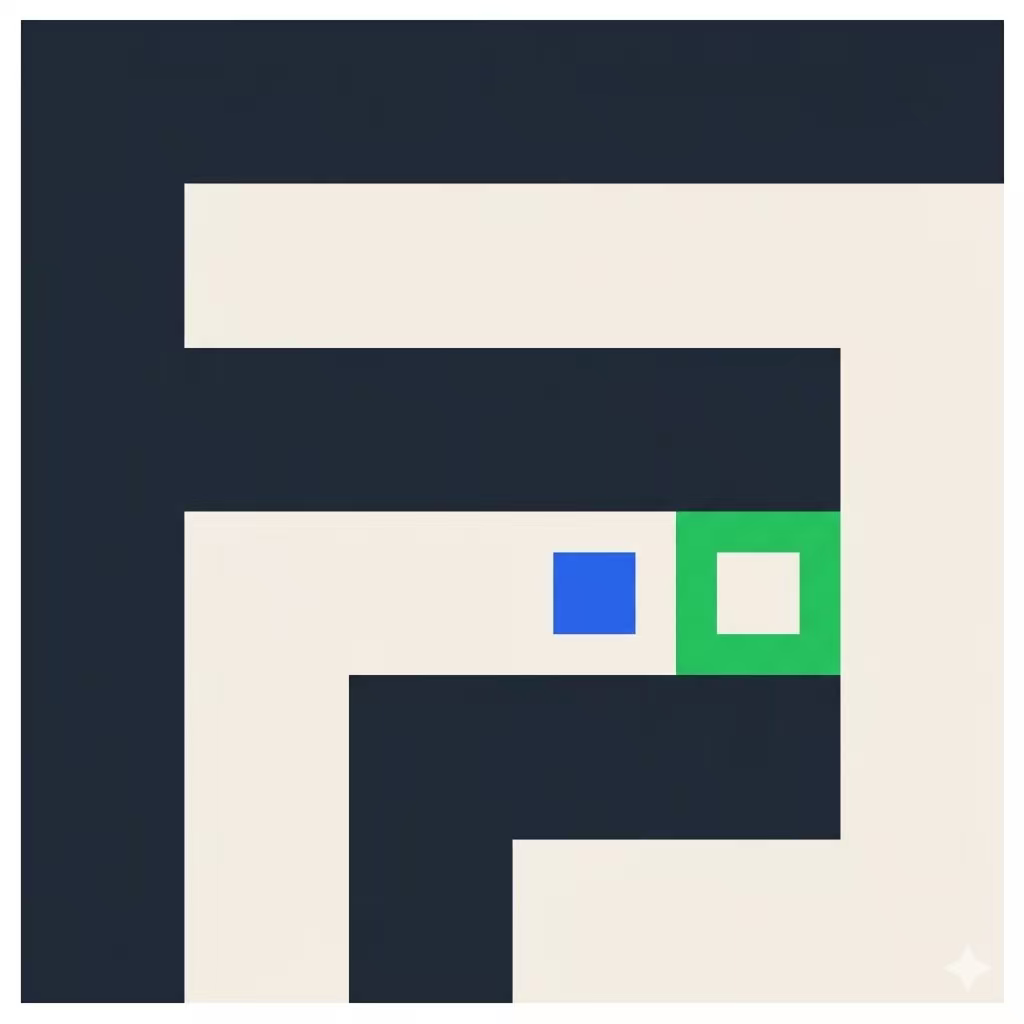}
    \caption{After move 1}
\end{subfigure}
\hfill
\begin{subfigure}{0.45\textwidth}
    \includegraphics[width=\textwidth]{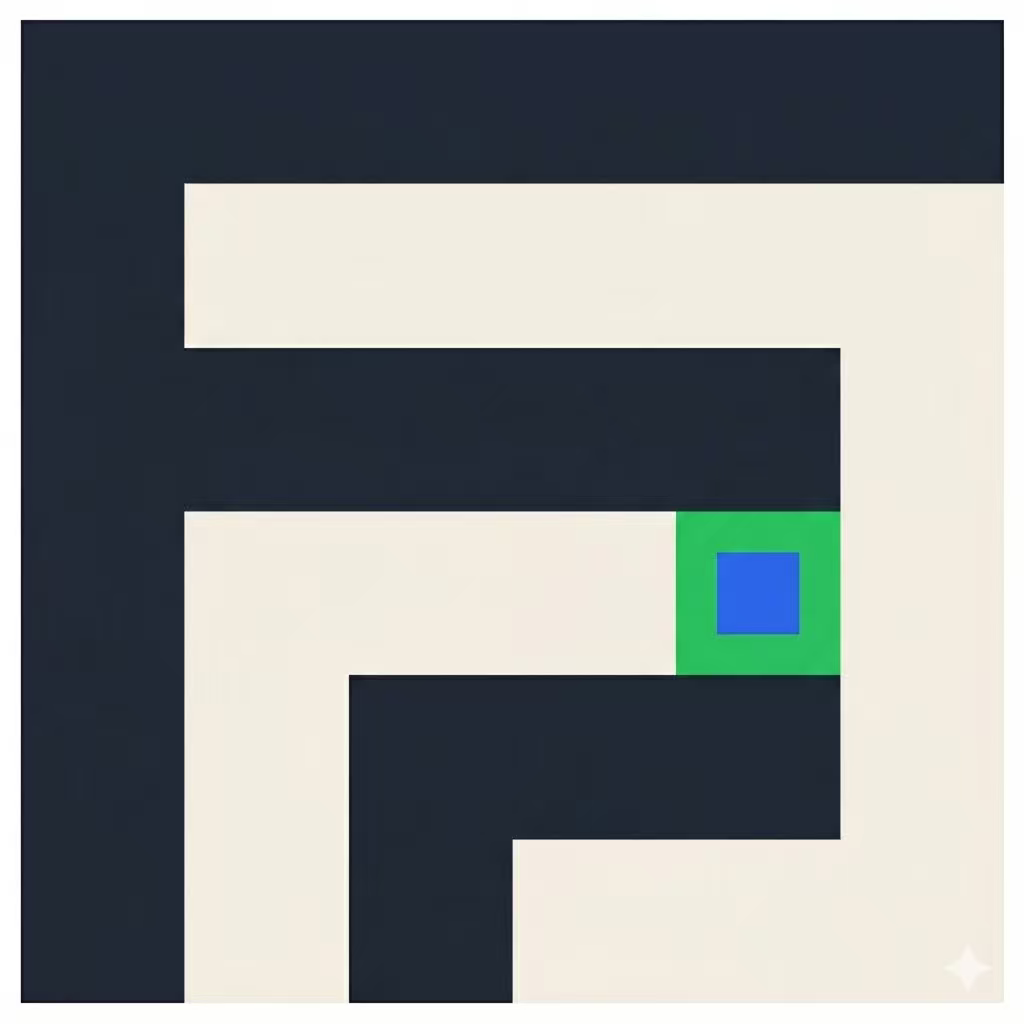}
    \caption{After move 2}
\end{subfigure}
\caption{Example step images (blue dot moves through the maze)}
\end{figure}

\textbf{Step 3 - Generate Final Answer:}

After generating all step images, the model outputs:
\begin{verbatim}
<ANSWER_JSON>["right", "down", "right", ...]</ANSWER_JSON>
\end{verbatim}

\textbf{Key Constraints:}
\begin{itemize}
    \item Number of generated images must equal number of moves
    \item Each image shows state after exactly one move
    \item Do not include initial state
    \item No collages, arrows, or annotations
\end{itemize}

\subsection{Sliding Puzzle Task}

\subsubsection{Task Description}

\textbf{Input:}
\begin{itemize}
    \item Initial state: 3×3 grid with 8 colored tiles and 1 empty space (red)
    \item Goal state: Solved puzzle configuration
\end{itemize}

\textbf{Task:} Find a sequence of moves to solve the puzzle.

\textbf{Move Semantics:} Moves are named by the direction the colored tile moves (not the empty space). Example: tile above empty moving down = "down".

\textbf{Constraints:} Legal moves: up, down, left, right; one tile per step.

\textbf{Output:} \verb|<ANSWER_JSON>["down", "right", ...]|

\textbf{Metrics:} \texttt{sliding\_text\_exact}, \texttt{sliding\_text\_frame\_acc}

\textbf{Dataset:} 54 samples

\subsubsection{Visual CoT Generation Flow}

\textbf{Step 1 - Input Images:}

\begin{figure}[h]
\centering
\begin{subfigure}{0.4\textwidth}
    \includegraphics[width=\textwidth]{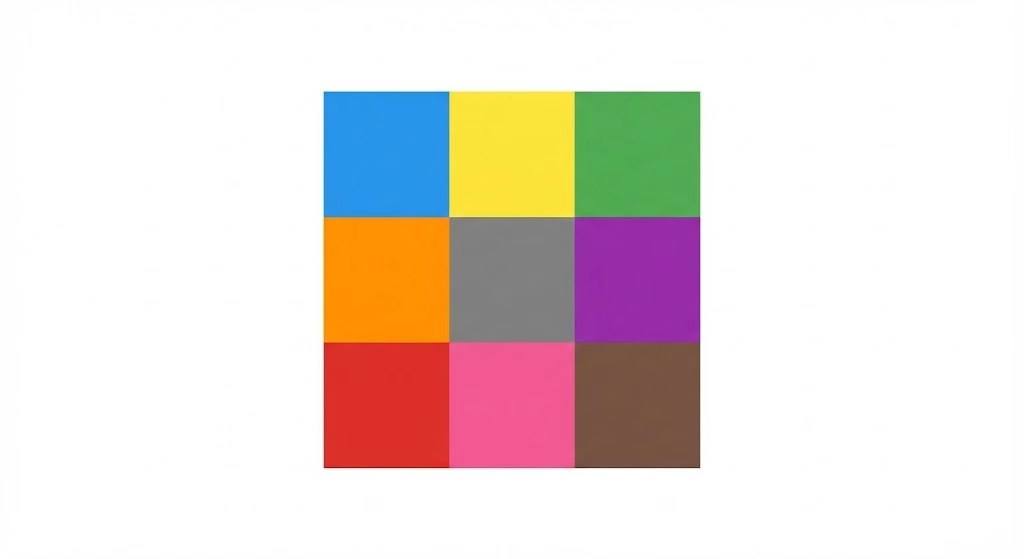}
    \caption{Initial state (scrambled)}
\end{subfigure}
\hfill
\begin{subfigure}{0.4\textwidth}
    \includegraphics[width=\textwidth]{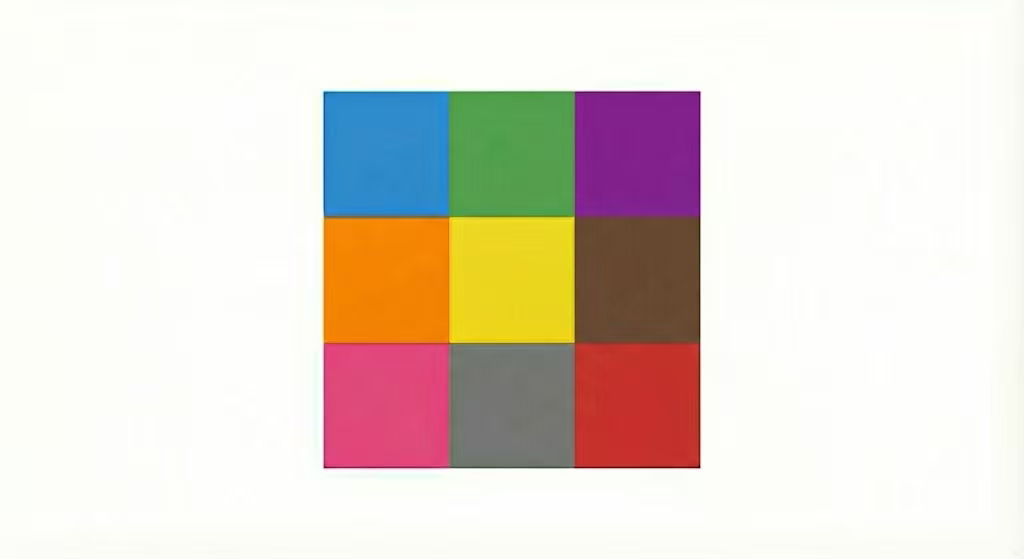}
    \caption{Goal state (solved)}
\end{subfigure}
\caption{Sliding puzzle input: initial and goal states}
\end{figure}

\textbf{Step 2 - Generate Step-by-Step Images:}

For each move in the solution, the model generates an image showing the puzzle state after that move. Only tile positions change; colors remain identical.

\begin{figure}[h]
\centering
\begin{subfigure}{0.3\textwidth}
    \includegraphics[width=\textwidth]{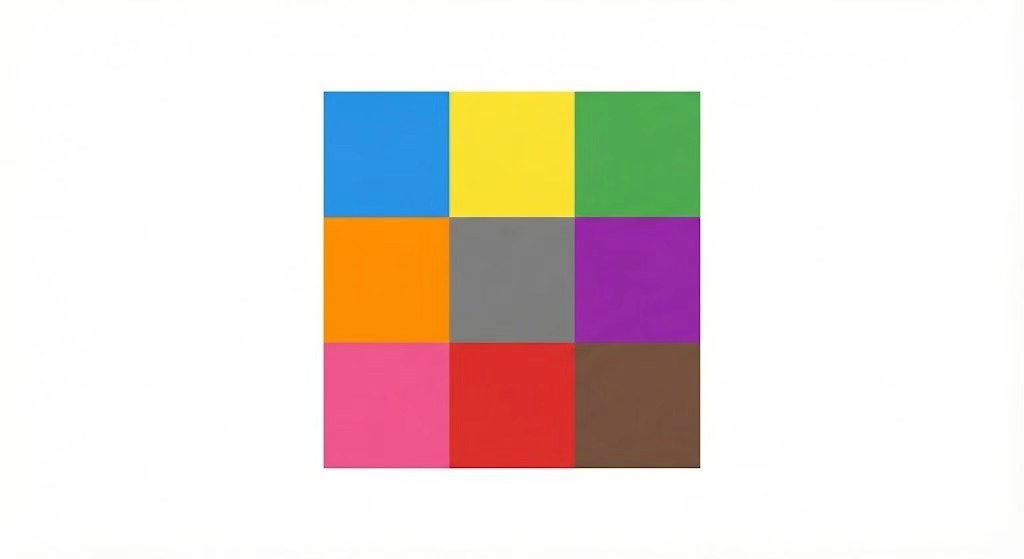}
    \caption{After move 1}
\end{subfigure}
\hfill
\begin{subfigure}{0.3\textwidth}
    \includegraphics[width=\textwidth]{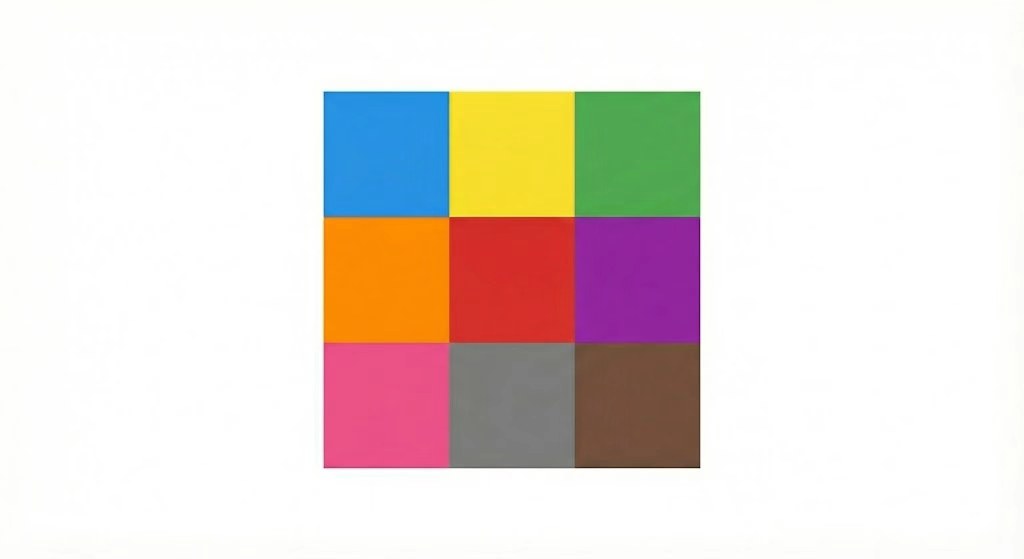}
    \caption{After move 2}
\end{subfigure}
\hfill
\begin{subfigure}{0.3\textwidth}
    \includegraphics[width=\textwidth]{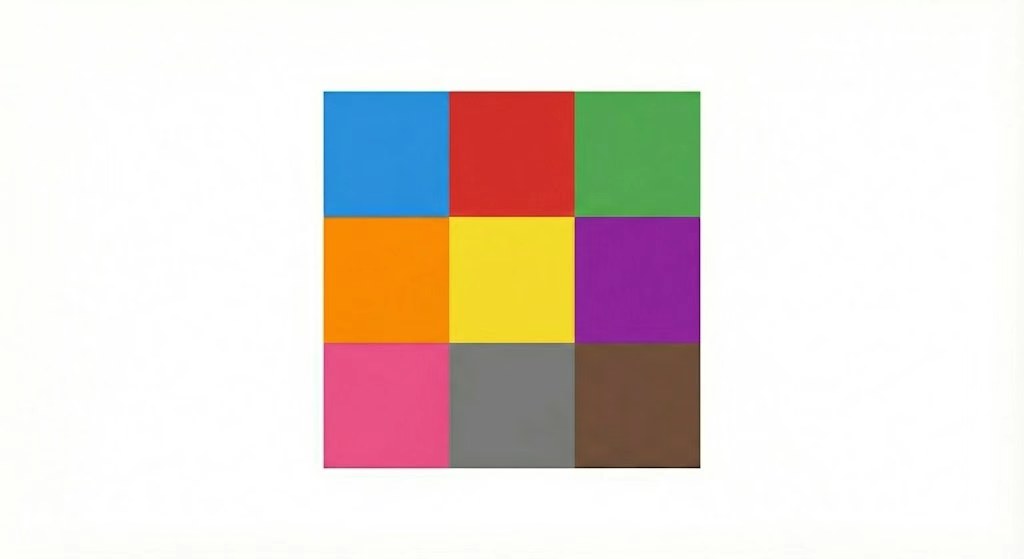}
    \caption{After move 3}
\end{subfigure}

\begin{subfigure}{0.3\textwidth}
    \includegraphics[width=\textwidth]{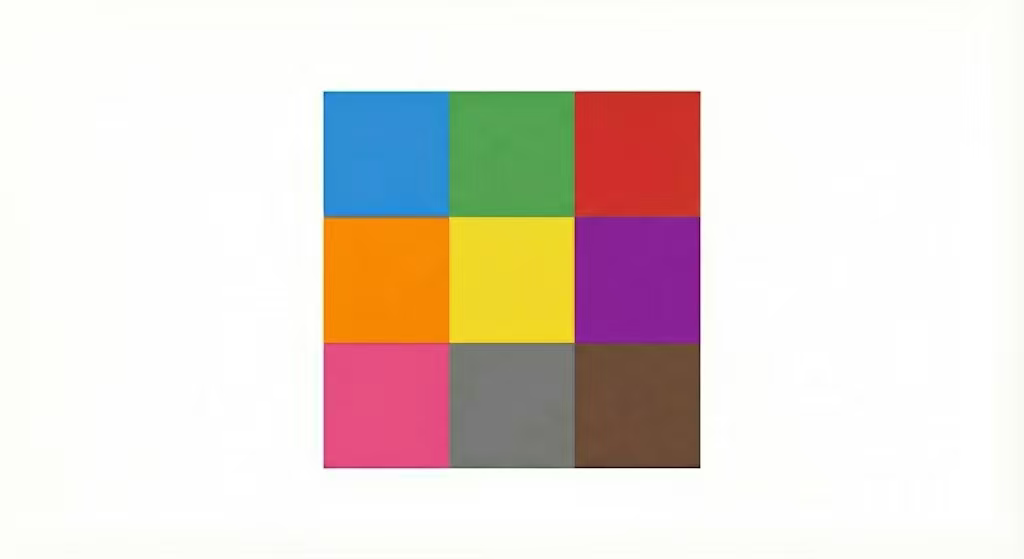}
    \caption{After move 4}
\end{subfigure}
\hfill
\begin{subfigure}{0.3\textwidth}
    \includegraphics[width=\textwidth]{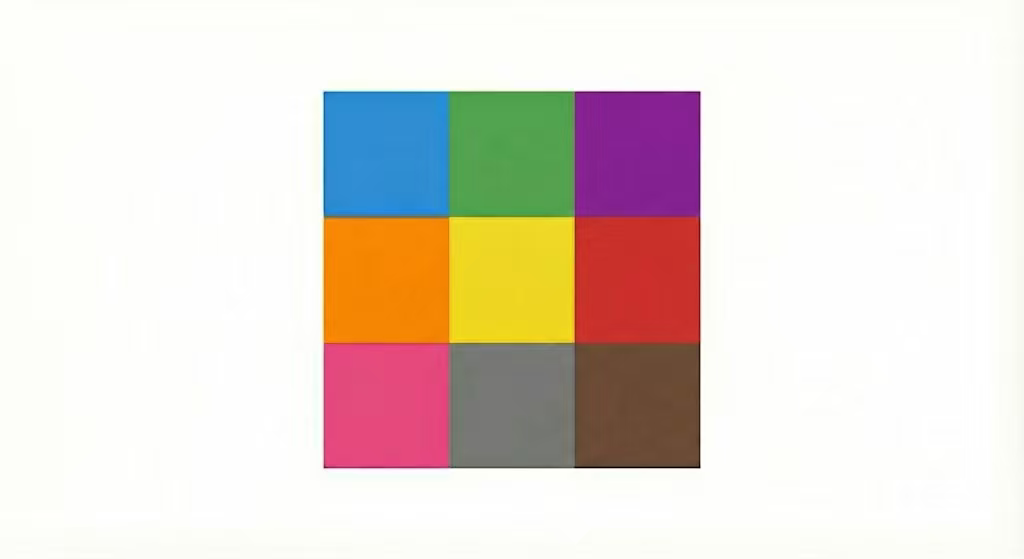}
    \caption{After move 5}
\end{subfigure}
\hfill
\begin{subfigure}{0.3\textwidth}
    \includegraphics[width=\textwidth]{sliding/26.png}
    \caption{After move 6 (solved)}
\end{subfigure}
\caption{Example step-by-step puzzle solving sequence}
\end{figure}

\textbf{Step 3 - Generate Final Answer:}

After generating all step images, the model outputs:
\begin{verbatim}
<ANSWER_JSON>["down", "right", "up", "left", "down", "right"]</ANSWER_JSON>
\end{verbatim}

\textbf{Key Constraints:}
\begin{itemize}
    \item Number of generated images must equal number of moves
    \item Each image shows state after exactly one move
    \item Do not include initial state
    \item No collages or multi-panel images
    \item Tile colors must remain pixel-identical
\end{itemize}

\section{Experimental Results}

\begin{table*}[!htbp]
\centering
\caption{\textbf{Main results on UniG2U across categories (Part 1/2).} All values are reported as percentages ($\times 100$). Baseline models (marked with $*$) are highlighted with gray background.}
\label{tab:main-results-part1}
\resizebox{\textwidth}{!}{%
\setlength{\tabcolsep}{3.5pt}
\renewcommand{\arraystretch}{1.06}
\scriptsize
\begin{tabular}{@{}lccccccccccccccc@{}}
\toprule
\multirow{2}{*}{\textbf{Model}} & \multicolumn{12}{c}{\textbf{Perception Reasoning}} & \multicolumn{2}{c}{\textbf{Real-world Apps}} & \multicolumn{1}{c}{\textbf{Chart}} \\
\cmidrule(lr){2-13} \cmidrule(lr){14-15} \cmidrule(lr){16-16}
& \textbf{icon\_scene} & \textbf{icon\_shape} & \textbf{in\_scene} & \textbf{in\_shape} & \textbf{logo\_scene} & \textbf{logo\_shape} & \textbf{algorithmic} & \textbf{analogical} & \textbf{deductive} & \textbf{inductive} & \textbf{spatial} & \textbf{Fine Discrim.} & \textbf{Attn. Focus.} & \textbf{VSP Reason.} & \textbf{ChartQA} \\
\midrule
\rowcolor{gray!12}
qwen2.5-vl-7b* & 98 & 5 & 98 & 10 & 94 & 20 & 31 & 24 & 46 & 26 & 28 & 11 & 44 & 36.5 & 86 \\
OmniGen2(direct) & 89 & 6 & 88 & 8 & 89 & 18 & 33 & 20 & 39 & 18 & 24 & 11 & 50 & 32.5 & 86 \\
OmniGen2(GtA) & 80 & 6 & 81 & 11 & 78 & 23 & 28 & 27 & 39 & 25 & 22 & 9.82 & 49 & 33.5 & 79 \\
UniWorld-V1(GtA) & 99 & 4 & 99 & 10 & 95 & 19 & 36 & 22 & 47 & 21 & 34 & 6.75 & 14 & 34 & 80 \\
OneCAT-3b(direct) & 85 & 2 & 84 & 8 & 89 & 13 & 30 & 29 & 38 & 20 & 22 & 10.02 & 46 & 33 & 76.68 \\
OneCAT-3b(GtA) & 87 & 1 & 82 & 7 & 90 & 4 & 23 & 29 & 35 & 21 & 24 & 9.2 & 33 & 34 & 74 \\
JavisGPT(direct) & 86 & 12 & 75 & 13 & 78 & 15 & 36 & 23 & 43 & 22 & 32 & 8.32 & 36 & 28 & 78.6 \\
uni-video(GtA) & 97 & 18 & 96 & 36 & 79 & 50 & 32 & 21 & 37 & 26 & 32 & 9.82 & 39 & 29 & 61 \\
UniPic2(GtA) & 94 & 8 & 88 & 24 & 90 & 22 & 32 & 22 & 40 & 23 & 28 & 11.66 & 21 & 14 & 47 \\
STAR-7B(GtA) & 95 & 7 & 98 & 13 & 93 & 19 & 32 & 21 & 46 & 23 & 29 & 6.75 & 35 & 33 & 79 \\
\midrule
\rowcolor{gray!12}
Qwen2.5-VL-3B* & 89 & 6 & 87 & 9 & 90 & 18 & 30 & 22 & 44 & 22 & 21 & 6.75 & 51 & 31 & 87 \\
UAE(direct) & 89 & 7 & 88 & 8 & 90 & 18 & 31 & 32 & 34 & 26 & 20 & 10 & 45 & 28.5 & 85 \\

Ovis-U1(direct) & 91 & 7 & 83 & 10 & 87 & 18 & 28 & 27 & 30 & 21 & 27 & 7.36 & 59 & 32 & 90 \\
Ovis-U1(GtA) & 75 & 5 & 73 & 15 & 48 & 21 & 30 & 22 & 30 & 27 & 30 & 6.13 & 23 & 19 & 32 \\
\midrule
\rowcolor{gray!12}
yi-6B-vl* & 81 & 17 & 20 & 16 & 84 & 23 & 31 & 18 & 27 & 25 & 33 & 11.66 & 27 & 21 & 19 \\
mio(direct) & 46 & 5 & 36 & 13 & 30 & 27 & 32 & 22 & 35 & 28 & 24 & 14.72 & 35 & 1 & 15 \\
mio(GtA) & 53 & 15 & 36 & 16 & 28 & 19 & 21 & 23 & 32 & 28 & 27 & 10.43 & 30 & 0 & 13 \\
\midrule
\rowcolor{gray!12}
deepseek-v2* & 60 & 8 & 78 & 13 & 90 & 19 & 40 & 23 & 35 & 22 & 34 & 12.88 & 22 & 0 & 74 \\
Janus-Pro(direct) & 82 & 4 & 85 & 21 & 74 & 32 & 26 & 21 & 30 & 25 & 27 & 9.82 & 27 & 28 & 29 \\
AIA(direct) & 85 & 5 & 87 & 18 & 81 & 28 & 31 & 22 & 24 & 31 & 29 & 11.04 & 25 & 2 & 33 \\
\midrule
\rowcolor{gray!12}
LLaDA-Instruct* & 7 & 0 & 9 & 8 & 7 & 5 & 29 & 24 & 32 & 24 & 24 & 3.07 & 32 & 22.5 & 9 \\
MMaDA(direct) & 55 & 1 & 47 & 3 & 44 & 4 & 25 & 29 & 31 & 18 & 22 & 5 & 32 & 25.5 & 14 \\
\midrule
\rowcolor{gray!12}
Janus-1.3B* & 57 & 6 & 35 & 10 & 35 & 23 & 21 & 23 & 25 & 22 & 22 & 11.04 & 26 & 1 & 50 \\
FUDOK(direct) & 10 & 10 & 10 & 5 & 11 & 4 & 25 & 24 & 32 & 28 & 24 & 8 & 34 & 1 & 22 \\
\midrule
\rowcolor{gray!12}
Qwen3-VL-8B* & 100 & 13 & 100 & 22 & 97 & 28 & 45 & 29 & 40 & 21 & 32 & 15 & 48 & 32 & 89 \\
MammothModa2(direct) & 98 & 5 & 98 & 10 & 94 & 20 & 37 & 26 & 48 & 20 & 30 & 4.91 & 45 & 22 & 91 \\
\midrule
\rowcolor{gray!12}
llava-hf* & 84 & 10 & 81 & 20 & 59 & 31 & 23 & 30 & 29 & 20 & 22 & 9.2 & 27 & 19 & 28 \\
Emu3(direct) & 51 & 0 & 60 & 2 & 35 & 5 & 33 & 25 & 34 & 24 & 28 & 9.82 & 31 & 2 & 67 \\
\midrule
\rowcolor{gray!12}
llava\_onevision* & 86 & 6 & 74 & 18 & 91 & 34 & 35 & 25 & 28 & 32 & 22 & 11.04 & 54 & 32 & 78 \\
X-Omni(direct) & 91 & 5 & 97 & 11 & 85 & 13 & 27 & 24 & 35 & 17 & 27 & 8.59 & 42 & 24 & 83 \\
bagel(direct) & 91 & 2 & 88 & 18 & 91 & 20 & 31 & 26 & 46 & 42 & 35 & 9 & 50 & 26 & 92 \\
bagel(GtA) & 81 & 2 & 81 & 17 & 80 & 20 & 30 & 26 & 43 & 40 & 35 & 9.82 & 46 & 19.5 & 91 \\
TokenFlow-XL(direct) & 86 & 12 & 75 & 13 & 78 & 15 & 36 & 23 & 43 & 22 & 32 & 8.32 & 36 & 28 & 78.6 \\
show-o2(direct) & 89 & 11 & 80 & 22 & 81 & 37 & 22 & 24 & 36 & 21 & 26 & 7.36 & 38 & 33 & 45 \\
show-o2(GtA) & 56 & 6 & 51 & 18 & 53 & 31 & 21 & 23 & 32 & 23 & 25 & 9.82 & 35 & 31 & 35 \\
ILLUME+(direct) & 86 & 4 & 84 & 20 & 91 & 26 & 22 & 28 & 30 & 18 & 23 & 6.75 & 36 & 13 & 80 \\
ILLUME+(GtA) & 78 & 14 & 83 & 16 & 87 & 21 & 18 & 27 & 37 & 17 & 25 & 4.29 & 21 & 20 & 58 \\
\midrule
\rowcolor{orange!8}
gpt4o & 96 & 16 & 93 & 28 & 100 & 33 & 37 & 32 & 42 & 27 & 32 & 10 & 53 & 37.5 & 58 \\
ovis2.5 & 99 & 8 & 92 & 21 & 97 & 27 & 34 & 34 & 49 & 25 & 32 & 9.82 & 57 & 31 & 89 \\
\midrule
GPT-4o + GPT-image & 90 & 17 & 89 & 25 & 92 & 30 & 35 & 27 & 32 & 30 & 19 & 15.95 & 47 & 18.5 & 60 \\
gemini pro & 100 & 38 & 100 & 43 & 100 & 58 & 79 & 73 & 50 & 31 & 62 & 23.31 & 64 & 79 & 68 \\
qwenedit & 99 & 5 & 98 & 11 & 89 & 25 & 24 & 21 & 45 & 27 & 31 & 7.36 & 42 & 25.5 & 80 \\
\bottomrule
\end{tabular}%
}
\end{table*}

\begin{table*}[!htbp]
\centering
\caption{\textbf{Main results on UniG2U across categories (Part 2/2).} Continuing from Part 1 with the remaining columns.}
\label{tab:main-results-part2}
\resizebox{\textwidth}{!}{%
\setlength{\tabcolsep}{3.5pt}
\renewcommand{\arraystretch}{1.06}
\scriptsize
\begin{tabular}{@{}lccccccccccccccc@{}}
\toprule
\multirow{2}{*}{\textbf{Model}} & \multicolumn{2}{c}{\textbf{Geometry Reasoning}} & \multicolumn{5}{c}{\textbf{Spatial Intel.}} & \multicolumn{6}{c}{\textbf{Puzzles \& Games}} & \multicolumn{2}{c}{\textbf{Physics}} \\
\cmidrule(lr){2-3} \cmidrule(lr){4-8} \cmidrule(lr){9-14} \cmidrule(lr){15-16}
& \textbf{Geometry} & \textbf{3d Geometry} & \textbf{MSR} & \textbf{Attr. (Meas.)} & \textbf{Attr. (Appr.)} & \textbf{Motion (Cam.)} & \textbf{Motion (Obj.)} & \textbf{Mental Recon.} & \textbf{Mental Track.} & \textbf{Visual Track.} & \textbf{maze} & \textbf{jigsaw} & \textbf{sliding} & \textbf{light} & \textbf{mechanism} \\
\midrule
\rowcolor{gray!12}
qwen2.5-vl-7b* & 46 & 0 & 25 & 20 & 10 & 29 & 33 & 36 & 25 & 9.64 & 14.12 & 52 & 5.61 & 34 & 34 \\
OmniGen2(direct) & 28 & 0 & 22 & 33 & 19 & 12 & 33 & 31 & 23 & 7.23 & 4.31 & 48 & 0 & 42 & 40 \\
OmniGen2(GtA) & 26 & 0 & 29 & 33 & 17 & 18 & 35 & 31 & 23 & 6.02 & 4.6 & 54 & 0.13 & 41 & 42 \\
UniWorld-V1(GtA) & 38 & 0 & 27 & 22 & 16 & 21 & 38 & 39 & 24 & 6.02 & 14.43 & 47 & 6.14 & 33 & 36 \\
OneCAT-3b(direct) & 34 & 0 & 32 & 30 & 27 & 24 & 13 & 24 & 25 & 6.02 & 8.4 & 50 & 11.18 & 41 & 23 \\
OneCAT-3b(GtA) & 17 & 0 & 39 & 28 & 23 & 3 & 26 & 23 & 24 & 6.02 & 7.43 & 50 & 3.1 & 37 & 21 \\
JavisGPT(direct) & 42 & 0 & 16 & 24 & 18 & 19 & 31 & 30 & 33 & 8.74 & 0.48 & 51 & 10.44 & 25 & 28 \\
uni-video(GtA) & 43 & 0 & 23 & 18 & 17 & 10 & 32 & 28 & 26 & 9.64 & 21.9 & 57 & 30.79 & 31 & 27 \\
UniPic2(GtA) & 33 & 1 & 21 & 23 & 21 & 23 & 35 & 33 & 22 & 4.82 & 11.53 & 48 & 18.71 & 28 & 34 \\
STAR-7B(GtA) & 29 & 0 & 27 & 22 & 17 & 21 & 39 & 31 & 27 & 6.02 & 6.87 & 46 & 19.72 & 34 & 31 \\
\midrule
\rowcolor{gray!12}
Qwen2.5-VL-3B* & 31 & 0 & 22 & 32 & 24 & 18 & 36 & 31 & 24 & 3.61 & 1.64 & 50 & 0 & 42 & 39 \\
UAE(direct) & 21 & 0 & 26 & 22 & 26 & 20 & 32 & 33 & 19 & 4.82 & 5.17 & 38 & 5.91 & 35 & 26 \\
Ovis-U1(direct) & 47 & 0 & 12 & 28 & 7 & 18 & 23 & 31 & 28 & 8.43 & 13.44 & 49 & 33.39 & 35 & 26 \\
Ovis-U1(GtA) & 33 & 0 & 27 & 27 & 15 & 11 & 22 & 22 & 19 & 3.61 & 0.25 & 52 & 0.74 & 19 & 15 \\
\midrule
\rowcolor{gray!12}
yi-6B-vl* & 0 & 0 & 24 & 26 & 22 & 11 & 30 & 33 & 22 & 6.02 & 0 & 11 & 0 & 35 & 32 \\
mio(direct) & 0 & 0 & 21 & 24 & 24 & 23 & 25 & 25 & 20 & 3.61 & 0 & 21 & 0 & 30 & 32 \\
mio(GtA) & 0 & 0 & 32 & 24 & 14 & 22 & 16 & 27 & 17 & 3.61 & 0 & 1 & 0 & 27 & 29 \\
\midrule
\rowcolor{gray!12}
deepseek-v2* & 9 & 0 & 23 & 19 & 9 & 14 & 22 & 11 & 1 & 6.02 & 11.2 & 53 & 28.82 & 21 & 23 \\
Janus-Pro(direct) & 15 & 0 & 28 & 37 & 25 & 16 & 15 & 21 & 26 & 3.61 & 12.53 & 54 & 7.87 & 21 & 17 \\
AIA(direct) & 17 & 0 & 23 & 43 & 21 & 19 & 27 & 31 & 19 & 4.82 & 12.87 & 51 & 19.93 & 31 & 31 \\
\midrule
\rowcolor{gray!12}
LLaDA-Instruct* & 3 & 0 & 21 & 41 & 18 & 11 & 26 & 22 & 26 & 3.61 & 12.87 & 48 & 0 & 24 & 18 \\
MMaDA(direct) & 10 & 0 & 32 & 27 & 25 & 20 & 21 & 21 & 28 & 2.41 & 0 & 35 & 0 & 28 & 25 \\
\midrule
\rowcolor{gray!12}
Janus-1.3B* & 5 & 0 & 15 & 28 & 21 & 18 & 27 & 24 & 19 & 2.41 & 1.81 & 18 & 0.26 & 19 & 39 \\
FUDOK(direct) & 2 & 0 & 29 & 28 & 36 & 29 & 23 & 22 & 25 & 3.61 & 0 & 0 & 0 & 25 & 17 \\
\midrule
\rowcolor{gray!12}
Qwen3-VL-8B* & 67 & 11 & 23 & 35 & 22 & 20 & 30 & 32 & 26 & 8.43 & 15.53 & 45 & 8.59 & 37 & 37 \\
MammothModa2(direct) & 0 & 0 & 14 & 19 & 18 & 18 & 40 & 35 & 24 & 7.23 & 0 & 0 & 0 & 40 & 33 \\
\midrule
\rowcolor{gray!12}
llava-hf* & 0 & 0 & 28 & 24 & 30 & 14 & 27 & 26 & 20 & 4.82 & 0 & 50 & 3.35 & 21 & 38 \\
Emu3(direct) & 2 & 0 & 31 & 23 & 28 & 14 & 20 & 33 & 27 & 4.82 & 1.39 & 0 & 11.85 & 25 & 16 \\
\midrule
\rowcolor{gray!12}
llava\_onevision* & 21 & 0 & 17 & 26 & 29 & 20 & 35 & 31 & 23 & 16.87 & 7.46 & 51 & 9.25 & 42 & 46 \\
X-Omni(direct) & 19 & 0 & 20 & 31 & 26 & 24 & 34 & 30 & 27 & 6.02 & 12.87 & 60 & 11.23 & 35 & 24 \\
bagel(direct) & 58 & 0 & 26 & 31 & 20 & 26 & 39 & 37 & 30 & 7.23 & 2.15 & 52 & 0.9 & 35 & 40 \\
bagel(GtA) & 58 & 2 & 25 & 36 & 27 & 24 & 33 & 41 & 31 & 4.82 & 28.1 & 55 & 19.45 & 38 & 43 \\
TokenFlow-XL(direct) & 42 & 0 & 16 & 24 & 18 & 19 & 31 & 30 & 33 & 8.74 & 0.48 & 51 & 10.44 & 25 & 28 \\
show-o2(direct) & 22 & 0 & 27 & 24 & 21 & 28 & 29 & 32 & 22 & 9.64 & 14.24 & 56 & 13.74 & 37 & 43 \\
show-o2(GtA) & 18 & 0 & 30 & 15 & 0 & 23 & 34 & 30 & 23 & 10.84 & 17.84 & 52 & 18.32 & 43 & 37 \\
ILLUME+(direct) & 15 & 0 & 28 & 29 & 14 & 14 & 31 & 27 & 19 & 6.02 & 6.93 & 53 & 4.02 & 38 & 32 \\
ILLUME+(GtA) & 4 & 0 & 29 & 28 & 15 & 16 & 17 & 20 & 18 & 2.41 & 12.87 & 52 & 0 & 37 & 34 \\
\midrule
\rowcolor{orange!8}
gpt4o & 53 & 1 & 33 & 38 & 32 & 34 & 43 & 33 & 24 & 14.46 & 32.15 & 52 & 21.99 & 36 & 33 \\
ovis2.5 & 64 & 1 & 27 & 40 & 21 & 22 & 29 & 25 & 23 & 8.43 & 15.07 & 45 & 16.93 & 42 & 44 \\
\midrule
GPT-4o + GPT-image & 47 & 7 & 23 & 30 & 23 & 23 & 36 & 36 & 20 & 16.87 & 39.84 & 50 & 29.4 & 37 & 24 \\
gemini pro & 86 & 84 & 35 & 60 & 29 & 28 & 40 & 39 & 70 & 26.51 & 49.29 & 77 & 73.57 & 89 & 93 \\
qwenedit & 34 & 0 & 19 & 19 & 15 & 18 & 37 & 32 & 23 & 7.23 & 25.57 & 52 & 19.7 & 33 & 30 \\
\bottomrule
\end{tabular}%
}
\end{table*}

\begin{table*}[!htbp]
\centering
\caption{\textbf{Delta results on UniG2U across categories (Sheet 3, Part 1/2).} Baseline models (marked with $*$) are highlighted with gray background, and missing values are denoted by `-`.}
\label{tab:sheet3-results-part1}
\resizebox{\textwidth}{!}{%
\setlength{\tabcolsep}{3.5pt}
\renewcommand{\arraystretch}{1.06}
\scriptsize
\begin{tabular}{@{}lccccccccccccccc@{}}
\toprule
\multirow{2}{*}{\textbf{Model}} & \multicolumn{12}{c}{\textbf{Perception Reasoning}} & \multicolumn{2}{c}{\textbf{Real-world Apps}} & \multicolumn{1}{c}{\textbf{Chart}} \\
\cmidrule(lr){2-13} \cmidrule(lr){14-15} \cmidrule(lr){16-16}
& \textbf{icon\_scene} & \textbf{icon\_shape} & \textbf{in\_scene} & \textbf{in\_shape} & \textbf{logo\_scene} & \textbf{logo\_shape} & \textbf{algorithmic} & \textbf{analogical} & \textbf{deductive} & \textbf{inductive} & \textbf{spatial} & \textbf{Fine Discrim.} & \textbf{Attn. Focus.} & \textbf{VSP Reason.} & \textbf{ChartQA} \\
\midrule
\rowcolor{gray!12}
qwen2.5-vl-7b* & - & - & - & - & - & - & - & - & - & - & - & - & - & - & - \\
OmniGen2(direct) & -9 & 1 & -10 & -2 & -5 & -2 & 2 & -4 & -7 & -8 & -4 & 0 & 6 & -4 & 0 \\
OmniGen2(GtA) & -18 & 1 & -17 & 1 & -16 & 3 & -3 & 3 & -7 & -1 & -6 & -1.18 & 5 & -3 & -7 \\
UniWorld-V1(GtA) & 1 & -1 & 1 & 0 & 1 & -1 & 5 & -2 & 1 & -5 & 6 & -4.25 & -30 & -2.5 & -6 \\
OneCAT-3b(direct) & -13 & -3 & -14 & -2 & -5 & -7 & -1 & 5 & -8 & -6 & -6 & -0.98 & 2 & -3.5 & -9.32 \\
OneCAT-3b(GtA) & -11 & -4 & -16 & -3 & -4 & -16 & -8 & 5 & -11 & -5 & -4 & -1.8 & -11 & -2.5 & -12 \\
JavisGPT(direct) & -12 & 7 & -23 & 3 & -16 & -5 & 5 & -1 & -3 & -4 & 4 & -2.68 & -8 & -8.5 & -7.4 \\
uni-video(GtA) & -1 & 13 & -2 & 26 & -15 & 30 & 1 & -3 & -9 & 0 & 4 & -1.18 & -5 & -7.5 & -25 \\
UniPic2(GtA) & -4 & 3 & -10 & 14 & -4 & 2 & 1 & -2 & -6 & -3 & 0 & 0.66 & -23 & -22.5 & -39 \\
STAR-7B(GtA) & -3 & 2 & 0 & 3 & -1 & -1 & 1 & -3 & 0 & -3 & 1 & -4.25 & -9 & -3.5 & -7 \\
\midrule
\rowcolor{gray!12}
Qwen2.5-VL-3B* & - & - & - & - & - & - & - & - & - & - & - & - & - & - & - \\
UAE(direct) & 0 & 1 & 1 & -1 & 0 & 0 & 1 & 10 & -10 & 4 & -1 & 3.25 & -6 & -2.5 & -2 \\
Ovis-U1(direct) & 2 & 1 & -4 & 1 & -3 & 0 & -2 & 5 & -14 & -1 & 6 & 0.61 & 8 & 1 & 3 \\
Ovis-U1(GtA) & -14 & -1 & -14 & 6 & -42 & 3 & 0 & 0 & -14 & 5 & 9 & -0.62 & -28 & -12 & -55 \\
\midrule
\rowcolor{gray!12}
yi-6B-vl* & - & - & - & - & - & - & - & - & - & - & - & - & - & - & - \\
mio(direct) & -35 & -12 & 16 & -3 & -54 & 4 & 1 & 4 & 8 & 3 & -9 & 3.06 & 8 & -20 & -4 \\
mio(GtA) & 7 & 10 & 0 & 3 & -2 & -8 & -11 & 1 & -3 & 0 & 3 & -4.29 & -5 & -1 & -2 \\
\midrule
\rowcolor{gray!12}
deepseek-v2* & - & - & - & - & - & - & - & - & - & - & - & - & - & - & - \\
Janus-Pro(direct) & 22 & -4 & 7 & 8 & -16 & 13 & -14 & -2 & -5 & 3 & -7 & -3.06 & 5 & 28 & -45 \\
AIA(direct) & 25 & -3 & 9 & 5 & -9 & 9 & -9 & -1 & -11 & 9 & -5 & -1.84 & 3 & 2 & -41 \\
\midrule
\rowcolor{gray!12}
LLaDA-Instruct* & - & - & - & - & - & - & - & - & - & - & - & - & - & - & - \\
MMaDA(direct) & 48 & 1 & 38 & -5 & 37 & -1 & -4 & 5 & -1 & -6 & -2 & 1.93 & 0 & 3 & 5 \\
\midrule
\rowcolor{gray!12}
Janus-1.3B* & - & - & - & - & - & - & - & - & - & - & - & - & - & - & - \\
FUDOK(direct) & -47 & 4 & -25 & -5 & -24 & -19 & 4 & 1 & 7 & 6 & 2 & -3.04 & 8 & 0 & -28 \\
\midrule
\rowcolor{gray!12}
Qwen3-VL-8B* & - & - & - & - & - & - & - & - & - & - & - & - & - & - & - \\
MammothModa2(direct) & -2 & -8 & -2 & -12 & -3 & -8 & -8 & -3 & 8 & -1 & -2 & -10.09 & -3 & -10 & 2 \\
\midrule
\rowcolor{gray!12}
llava-hf* & - & - & - & - & - & - & - & - & - & - & - & - & - & - & - \\
Emu3(direct) & -33 & -10 & -21 & -18 & -24 & -26 & 10 & -5 & 5 & 4 & 6 & 0.62 & 4 & -17 & 39 \\
\midrule
\rowcolor{gray!12}
llava\_onevision* & - & - & - & - & - & - & - & - & - & - & - & - & - & - & - \\
X-Omni(direct) & 5 & -1 & 23 & -7 & -6 & -21 & -8 & -1 & 7 & -15 & 5 & -2.45 & -12 & -8 & 5 \\
bagel(direct) & 5 & -4 & 14 & 0 & 0 & -14 & -4 & 1 & 18 & 10 & 13 & -2.04 & -4 & -6 & 14 \\
bagel(GtA) & -5 & -4 & 7 & -1 & -11 & -14 & -5 & 1 & 15 & 8 & 13 & -1.22 & -8 & -12.5 & 13 \\
TokenFlow-XL(direct) & 0 & 6 & 1 & -5 & -13 & -19 & 1 & -2 & 15 & -10 & 10 & -2.72 & -18 & -4 & 0.6 \\
show-o2(direct) & 3 & 5 & 6 & 4 & -10 & 3 & -13 & -1 & 8 & -11 & 4 & -3.68 & -16 & 1 & -33 \\
show-o2(GtA) & -30 & 0 & -23 & 0 & -38 & -3 & -14 & -2 & 4 & -9 & 3 & -1.22 & -19 & -1 & -43 \\
ILLUME+(direct) & 0 & -2 & 10 & 2 & 0 & -8 & -13 & 3 & 2 & -14 & 1 & -4.29 & -18 & -19 & 2 \\
ILLUME+(GtA) & -8 & 8 & 9 & -2 & -4 & -13 & -17 & 2 & 9 & -15 & 3 & -6.75 & -33 & -12 & -20 \\
\midrule
\rowcolor{orange!8}
gpt4o & - & - & - & - & - & - & - & - & - & - & - & - & - & - & - \\
gpt4o & - & - & - & - & - & - & - & - & - & - & - & - & - & - & - \\
\midrule
GPT-4o + GPT-image & - & - & - & - & - & - & - & - & - & - & - & - & - & - & - \\
gemini pro & - & - & - & - & - & - & - & - & - & - & - & - & - & - & - \\
qwenedit & - & - & - & - & - & - & - & - & - & - & - & - & - & - & - \\
\bottomrule
\end{tabular}%
}
\end{table*}

\begin{table*}[!htbp]
\centering
\caption{\textbf{Delta results on UniG2U across categories (Sheet 3, Part 2/2).} Continuing from Part 1 with the remaining columns.}
\label{tab:sheet3-results-part2}
\resizebox{\textwidth}{!}{%
\setlength{\tabcolsep}{3.5pt}
\renewcommand{\arraystretch}{1.06}
\scriptsize
\begin{tabular}{@{}lccccccccccccccc@{}}
\toprule
\multirow{2}{*}{\textbf{Model}} & \multicolumn{2}{c}{\textbf{Geometry Reasoning}} & \multicolumn{5}{c}{\textbf{Spatial Intel.}} & \multicolumn{6}{c}{\textbf{Puzzles \& Games}} & \multicolumn{2}{c}{\textbf{Physics}} \\
\cmidrule(lr){2-3} \cmidrule(lr){4-8} \cmidrule(lr){9-14} \cmidrule(lr){15-16}
& \textbf{Geometry} & \textbf{3d Geometry} & \textbf{MSR} & \textbf{Attr. (Meas.)} & \textbf{Attr. (Appr.)} & \textbf{Motion (Cam.)} & \textbf{Motion (Obj.)} & \textbf{Mental Recon.} & \textbf{Mental Track.} & \textbf{Visual Track.} & \textbf{maze} & \textbf{jigsaw} & \textbf{sliding} & \textbf{light} & \textbf{mechanism} \\
\midrule
\rowcolor{gray!12}
qwen2.5-vl-7b* & - & - & - & - & - & - & - & - & - & - & - & - & - & - & - \\
OmniGen2(direct) & -18 & 0 & -3 & 13 & 9 & -17 & 0 & -5 & -2 & -2.41 & -9.81 & -4 & -5.61 & 8 & 6 \\
OmniGen2(GtA) & -20 & 0 & 4 & 13 & 7 & -11 & 2 & -5 & -2 & -3.62 & -9.52 & 2 & -5.48 & 7 & 8 \\
UniWorld-V1(GtA) & -8 & 0 & 2 & 2 & 6 & -8 & 5 & 3 & -1 & -3.62 & 0.31 & -5 & 0.53 & -1 & 2 \\
OneCAT-3b(direct) & -12 & 0 & 7 & 10 & 17 & -5 & -20 & -12 & 0 & -3.62 & -5.72 & -2 & 5.57 & 7 & -11 \\
OneCAT-3b(GtA) & -29 & 0 & 14 & 8 & 13 & -26 & -7 & -13 & -1 & -3.62 & -6.69 & -2 & -2.51 & 3 & -13 \\
JavisGPT(direct) & -4 & 0 & -9 & 4 & 8 & -10 & -2 & -6 & 8 & -0.9 & -13.64 & -1 & 4.83 & -9 & -6 \\
uni-video(GtA) & -3 & 0 & -2 & -2 & 7 & -19 & -1 & -8 & 1 & 0 & 7.78 & 5 & 25.18 & -3 & -7 \\
UniPic2(GtA) & -13 & 1 & -4 & 3 & 11 & -6 & 2 & -3 & -3 & -4.82 & -2.59 & -4 & 13.1 & -6 & 0 \\
STAR-7B(GtA) & -17 & 0 & 2 & 2 & 7 & -8 & 6 & -5 & 2 & -3.62 & -7.25 & -6 & 14.11 & 0 & -3 \\
\midrule
\rowcolor{gray!12}
Qwen2.5-VL-3B* & - & - & - & - & - & - & - & - & - & - & - & - & - & - & - \\
UAE(direct) & -10 & 0 & 4 & -10 & 2 & 2 & -4 & 2 & -5 & 1.21 & 3.53 & -12 & 5.91 & -7 & -13 \\
Ovis-U1(direct) & 16 & 0 & -10 & -4 & -17 & 0 & -13 & 0 & 4 & 4.82 & 11.8 & -1 & 33.39 & -7 & -13 \\
Ovis-U1(GtA) & 2 & 0 & 5 & -5 & -9 & -7 & -14 & -9 & -5 & 0 & -1.39 & 2 & 0.74 & -23 & -24 \\
\midrule
\rowcolor{gray!12}
yi-6B-vl* & - & - & - & - & - & - & - & - & - & - & - & - & - & - & - \\
mio(direct) & 0 & 0 & -3 & -2 & 2 & 12 & -5 & -8 & -2 & -2.41 & 0 & 10 & 0 & -5 & 0 \\
mio(GtA) & 0 & 0 & 11 & 0 & -10 & -1 & -9 & 2 & -3 & 0 & 0 & -20 & 0 & -3 & -3 \\
\midrule
\rowcolor{gray!12}
deepseek-v2* & - & - & - & - & - & - & - & - & - & - & - & - & - & - & - \\
Janus-Pro(direct) & 6 & 0 & 5 & 18 & 16 & 2 & -7 & 10 & 25 & -2.41 & 1.33 & 1 & -20.95 & 0 & -6 \\
AIA(direct) & 8 & 0 & 0 & 24 & 12 & 5 & 5 & 20 & 18 & -1.2 & 1.67 & -2 & -8.89 & 10 & 8 \\
\midrule
\rowcolor{gray!12}
LLaDA-Instruct* & - & - & - & - & - & - & - & - & - & - & - & - & - & - & - \\
MMaDA(direct) & 7 & 0 & 11 & -14 & 7 & 9 & -5 & -1 & 2 & -1.2 & -12.87 & -13 & 0 & 4 & 7 \\
\midrule
\rowcolor{gray!12}
Janus-1.3B* & - & - & - & - & - & - & - & - & - & - & - & - & - & - & - \\
FUDOK(direct) & -3 & 0 & 14 & 0 & 15 & 11 & -4 & -2 & 6 & 1.2 & -1.81 & -18 & -0.26 & 6 & -22 \\
\midrule
\rowcolor{gray!12}
Qwen3-VL-8B* & - & - & - & - & - & - & - & - & - & - & - & - & - & - & - \\
MammothModa2(direct) & -67 & -11 & -9 & -16 & -4 & -2 & 10 & 3 & -2 & -1.2 & -15.53 & -45 & -8.59 & 3 & -4 \\
\midrule
\rowcolor{gray!12}
llava-hf* & - & - & - & - & - & - & - & - & - & - & - & - & - & - & - \\
Emu3(direct) & 2 & 0 & 3 & -1 & -2 & 0 & -7 & 7 & 7 & 0 & 1.39 & -50 & 8.5 & 4 & -22 \\
\midrule
\rowcolor{gray!12}
llava\_onevision* & - & - & - & - & - & - & - & - & - & - & - & - & - & - & - \\
X-Omni(direct) & -2 & 0 & 3 & 5 & -3 & 4 & -1 & -1 & 4 & -10.85 & 5.41 & 9 & 1.98 & -7 & -22 \\
bagel(direct) & 37 & 0 & 9 & 5 & -9 & 6 & 4 & 6 & 7 & -9.64 & -5.31 & 1 & -8.35 & -7 & -6 \\
bagel(GtA) & 37 & 2 & 8 & 10 & -2 & 4 & -2 & 10 & 8 & -12.05 & 20.64 & 4 & 10.2 & -4 & -3 \\
TokenFlow-XL(direct) & 21 & 0 & -1 & -2 & -11 & -1 & -4 & -1 & 10 & -8.13 & -6.98 & 0 & 1.19 & -17 & -18 \\
show-o2(direct) & 1 & 0 & 10 & -2 & -8 & 8 & -6 & 1 & -1 & -7.23 & 6.78 & 5 & 4.49 & -5 & -3 \\
show-o2(GtA) & -3 & 0 & 13 & -11 & -29 & 3 & -1 & -1 & 0 & -6.03 & 10.38 & 1 & 9.07 & 1 & -9 \\
ILLUME+(direct) & -6 & 0 & 11 & 3 & -15 & -6 & -4 & -4 & -4 & -10.85 & -0.53 & 2 & -5.23 & -4 & -14 \\
ILLUME+(GtA) & -17 & 0 & 12 & 2 & -14 & -4 & -18 & -11 & -5 & -14.46 & 5.41 & 1 & -9.25 & -5 & -12 \\
\midrule
\rowcolor{orange!8}
gpt4o & - & - & - & - & - & - & - & - & - & - & - & - & - & - & - \\
ovis2.5 & - & - & - & - &- & - & - & - & - & - & - & - & - & - & - \\
\midrule
GPT-4o + GPT-image & - & - & - & - & - & - & - & - & - & - & - & - & - & - & - \\
gemini pro & - & - & - & - & - & - & - & - & - & - & - & - & - & - & - \\
qwenedit & - & - & - & - & - & - & - & - & - & - & - & - & - & - & - \\
\bottomrule
\end{tabular}%
}
\end{table*}

\section{Task Prompts (Non-CoT Version)}
\label{app:non-cot}

\subsection{MMSI-Bench Tasks}

\subsubsection{mmsi\_attribute\_appr (Attribute - Appearance) and mmsi\_attribute\_meas (Attribute - Measurable) and mmsi\_msr (Multi-Step Reasoning)}

\begin{verbatim}
pre_prompt: "Pay close attention to the visual appearance attributes
(color, shape, texture, orientation, count, visual features) in the
images. Observe details carefully to identify and count objects.\n\n"

post_prompt: "\n\nBased on your visual observation, answer with the
option's letter from the given choices directly. Enclose the option's
letter within ``."
\end{verbatim}

\subsubsection{mmsi\_motion\_cam (Motion - Camera) and mmsi\_motion\_obj (Motion - Object)}

\textbf{Note:} These two tasks currently share identical prompts in the non-CoT version.

\begin{verbatim}
pre_prompt: "Analyze the consecutive first-person perspective images.
Determine how the camera (viewpoint) is moving or rotating by observing
changes in the scene and object positions relative to the viewer.\n\n"

post_prompt: "\n\nBased on your analysis of camera motion, answer with
the option's letter from the given choices directly. Enclose the
option's letter within ``."
\end{verbatim}

\subsection{IllusionBench Tasks}

\subsubsection{ICON Shape Recognition}

\begin{verbatim}
shape_string = "Animal, Face_Emoji, Music, Sport, Stationery, Vehicle"
scene_string = "Ocean, Origami, Forest, Cloud, Sand_dune,
Medieval_Village, City, Underwater_ruins, Museum, Bazaar_market,
Time_square"

prompt = f"This image contains a icon integrated into a background,
where elements of the background contribute to forming the icon.
Identify the icon that is represented in the image by choosing
exclusively among the following options:{shape_string},{scene_string}
Provide your response by stating only the single, most accurate class
name that represents the icon. You have to respond with a single word."
\end{verbatim}

\subsubsection{ICON Scene Recognition, LOGO Scene Recognition, and IN Scene Recognition}

\textbf{Note:} These three tasks share identical prompts (only differ in shape\_string values).

\begin{verbatim}
prompt = f"This image contains an icon integrated into a background,
where elements of the background contribute to forming the icon.
Identify the background that is represented in the image by choosing
exclusively among the following options:{shape_string},{scene_string}.
Provide your response by stating only the single, most accurate class
name that represents the background. You have to respond with a single
word."
\end{verbatim}

\subsubsection{LOGO Shape Recognition}

\begin{verbatim}
shape_string = "Adidas, Amazon, Apple, Audi, BMW, Mercedes Benz,
Facebook, Google, Instagram, Mcdonalds, Nasa, Nike, Olympics,
Playstation, Puma, Reebok, Spotify, Starbucks, Tesla, Telegram, Ubuntu"
scene_string = "Ocean, Origami, Forest, Cloud, Sand_dune,
Medieval_Village, City, Underwater_ruins, Museum, Bazaar_market,
Time_square"

prompt = f"This image contains a icon integrated into a background,
where elements of the background contribute to forming the logo.
Identify the logo that is represented in the image by choosing
exclusively among the following options:{shape_string},{scene_string}
Provide your response by stating only the single, most accurate class
name that represents the logo. You have to respond with a single word."
\end{verbatim}

\subsubsection{IN Shape Recognition}

\begin{verbatim}
shape_string = "Airplane, Bicycle, Bird, Bottle, Car, Cat, Dog,
Dolphin, Fork, Guitar, Mug, Panda, Paper_clip, Sailboat, Scooter,
Teapot"
scene_string = "Ocean, Origami, Forest, Cloud, Sand_dune,
Medieval_Village, City, Underwater_ruins, Museum, Bazaar_market,
Time_square"

prompt = f"This image contains a icon integrated into a background,
where elements of the background contribute to forming the icon.
Identify the shape that is represented in the image by choosing
exclusively among the following options:{shape_string},{scene_string}
Provide your response by stating only the single, most accurate class
name that represents the icon. You have to respond with a single word."
\end{verbatim}

\subsection{VisualPuzzles Tasks}

\subsubsection{All Categories (Algorithmic, Analogical, Deductive, Inductive, Spatial)}

\begin{verbatim}
MULTI_CHOICE_DIRECT_PROMPT = "Answer the question with the option's
letter from the given choices directly."
\end{verbatim}

\textbf{Full prompt format:}

\begin{verbatim}
question = "Question: " + doc["question"].strip()
options = doc["options"]
if options is not None:
    question += "\nOptions:\n(A) " + options[0] + "\n(B) " + options[1]
    + "\n(C) " + options[2] + "\n(D) " + options[3]
else:
    question += "\nOptions: Choose from (A) (B) (C) (D) in the image."

question += "\n" + MULTI_CHOICE_DIRECT_PROMPT
\end{verbatim}

\subsection{Uni-MMMU Tasks}

·\subsubsection{uni\_mmmu\_jigsaw100\_visual\_cot}

\textbf{Full Prompt:}

\begin{verbatim}
You are a unified vision-language model. You will be given:
(1) a 2x2 reference image with the bottom-right cell hidden, and
(2) two candidate patch images ("Candidate 0" and "Candidate 1").

Your job:
- For each candidate, synthesize a completed 2x2 image by placing that
  candidate EXACTLY into the bottom-right cell. Keep the other three
  cells pixel-identical to the reference (no filtering, no
  re-rendering). If sizes differ, only scale the candidate to fit that
  quadrant; do NOT rotate, mirror, or alter colors.
- Compare the two completed results and decide which candidate yields
  the correct completion.

Output EXACTLY the following, in order:

1) A single image with Candidate 0 placed in the bottom-right cell

2) A single image with Candidate 1 placed in the bottom-right cell


3) analysis comparing seam continuity, color/texture gradient,
   structural alignment, and global semantics

4) One strict JSON object with your decision, wrapped as:
<FINAL_ANSWER_JSON>
{"choice": 0 or 1, "rationale": "<=30 words decisive cue"}
</FINAL_ANSWER_JSON>

Hard constraints:
- Deterministic, single outcome. No hedging, no multiple possibilities.
- No meta talk about prompts, models, or pipelines.
- Do not restate the task as the answer; reason from visual evidence.
- The only edits allowed are pasting the candidate into the bottom-
  right cell and necessary size matching for that cell. All other
  pixels must remain identical to the reference.

Inputs:
\end{verbatim}

\subsubsection{uni\_mmmu\_maze100\_visual\_cot}

\textbf{Full Prompt:}

\begin{verbatim}
You are a precise maze solver.

SEMANTICS (for all mazes)
- Black squares: walls (impassable)
- White squares: path (walkable)
- Blue dot: start (the agent)
- Green rectangular frame: goal (reaching any white cell inside the
  green frame counts as success)
- Legal moves: up, down, left, right only. One cell per step; no
  diagonals, no jumps; never cross walls.

OUTPUT FORMAT (STRICT)
1) MULTI-IMAGE MODE - generate a SEQUENCE OF SEPARATE IMAGES, one per
   move:
   - Each output image must depict the maze state AFTER applying
     exactly one legal move.
   - Do NOT include the initial (pre-move) state.
   - Keep palette/layout/scale identical to the input; only the blue
     dot moves.
   - The number of returned images MUST equal the number of moves in
     the final answer (see step 2).
   - Absolutely FORBIDDEN: any collage/montage/spritesheet/grid/
     multi-panel/side-by-side/stacked images; no arrows, captions, or
     overlays; no GIFs/animations/video.

2) After all step images, emit EXACTLY ONE LINE containing ONLY the
   final move list as a JSON array of lowercase strings, wrapped as:
   <ANSWER_JSON>["right","down","left"]</ANSWER_JSON>


NO EXTRAS
- No tools, no OCR, no explanations, and no text other than the single
  <ANSWER_JSON>.....</ANSWER_JSON> line.
- Do not restate the instructions or the condition.

REMINDERS
- Decide the full path first, then emit the image sequence (one image
  per move), then the single <ANSWER_JSON> line.
- One move per image; images must be separate files/parts, not stitched
  together in any way.
\end{verbatim}

\subsubsection{uni\_mmmu\_sliding54\_visual\_cot}

\textbf{Full Prompt:}

\begin{verbatim}
You are a precise sliding puzzle solver.

SEMANTICS
- The puzzle is a 3x3 grid with 8 colored tiles and one empty space
  (shown as red).
- A "move" slides an adjacent tile INTO the empty space.
- Moves are named by the direction the COLORED TILE moves (not the
  empty space).
- Legal moves: up, down, left, right only. One tile per step.

OUTPUT FORMAT (STRICT)
1) MULTI-IMAGE MODE - generate a SEQUENCE OF SEPARATE IMAGES, one per
   move:
   - Each output image must depict the puzzle state AFTER applying
     exactly one move.
   - Do NOT include the initial (pre-move) state.
   - Keep tile colors identical; only positions change.
   - The number of returned images MUST equal the number of moves in
     the final answer.
   - Absolutely FORBIDDEN: any collage/montage/spritesheet/grid/
     multi-panel/side-by-side/stacked images.

2) After all step images, emit EXACTLY ONE LINE containing ONLY the
   final move list as a JSON array of lowercase strings, wrapped as:
   <ANSWER_JSON>["down","right","up"]</ANSWER_JSON>

NO EXTRAS
- No explanations beyond the single <ANSWER_JSON>.....</ANSWER_JSON>
  line.
- Do not restate the instructions.
\end{verbatim}
\section{Task Prompts (Visual CoT Version)}
\label{app:visual-cot}

This appendix contains all the original English prompts for Visual CoT tasks.

All Visual CoT tasks use a two-stage format:
\begin{verbatim}
[GEN_PROMPT]...[/GEN_PROMPT][QUESTION]...[/QUESTION]
\end{verbatim}

\subsection{MMSI-Bench Tasks}

\subsubsection{mmsi\_attribute\_appr\_visual\_cot}

\textbf{Stage 1 - Generation Prompt:}

\begin{verbatim}
Create a visualization that highlights and labels the visual appearance
attributes (color, shape, texture, orientation, count) in the scene.
Use annotations, bounding boxes, and labels to make object features and
counts clearly visible.
\end{verbatim}

\textbf{Stage 2 - Question Prompt:}

\begin{verbatim}
pre_prompt: "You are given the original image(s) and a visualization
highlighting appearance attributes. Use both to analyze color, shape,
texture, orientation, and object counts.\n\n"

post_prompt: "\n\nBased on your visual observation, answer with the
option's letter from the given choices directly. Enclose the option's
letter within ``."
\end{verbatim}

\subsubsection{mmsi\_attribute\_meas\_visual\_cot}

\textbf{Stage 1 - Generation Prompt:}

\begin{verbatim}
Create a visualization that highlights and annotates the measurable
attributes (size, length, width, height, distance, area, volume) in
the scene. Draw measurement lines, labels, and comparison markers to
make quantitative relationships clear.
\end{verbatim}

\textbf{Stage 2 - Question Prompt:}

\begin{verbatim}
pre_prompt: "You are given the original image(s) and a visualization
highlighting measurable attributes. Use both to analyze size, length,
width, height, distance, area, and volume relationships.\n\n"

post_prompt: "\n\nBased on your measurement analysis, answer with the
option's letter from the given choices directly. Enclose the option's
letter within ``."
\end{verbatim}

\subsubsection{mmsi\_motion\_cam\_visual\_cot}

\textbf{Stage 1 - Generation Prompt:}

\begin{verbatim}
Create a visualization showing camera motion analysis. Draw arrows
indicating the direction of camera movement/rotation, highlight
reference points that shift between frames, and annotate the type of
camera motion (pan, tilt, zoom, dolly, etc.).
\end{verbatim}

\textbf{Stage 2 - Question Prompt:}

\begin{verbatim}
pre_prompt: "You are given consecutive first-person perspective images
and a visualization of camera motion. Use both to determine how the
camera (viewpoint) is moving or rotating.\n\n"

post_prompt: "\n\nBased on your analysis of camera motion, answer with
the option's letter from the given choices directly. Enclose the
option's letter within ``."
\end{verbatim}

\subsubsection{mmsi\_motion\_obj\_visual\_cot}

\textbf{Stage 1 - Generation Prompt:}

\begin{verbatim}
Create a visualization showing object motion tracking. Draw motion
trails, arrows indicating direction of movement, and highlight moving
objects with bounding boxes. Annotate relative speeds and trajectories
of different objects.
\end{verbatim}

\textbf{Stage 2 - Question Prompt:}

\begin{verbatim}
pre_prompt: "You are given consecutive images and a visualization of
object motion. Use both to track objects, identify which are moving,
their direction, and relative speeds.\n\n"

post_prompt: "\n\nBased on your analysis of object motion, answer with
the option's letter from the given choices directly. Enclose the
option's letter within ``."
\end{verbatim}

\subsubsection{mmsi\_msr\_visual\_cot}

\textbf{Stage 1 - Generation Prompt:}

\begin{verbatim}
Create a visualization that breaks down the multi-step reasoning
process. Annotate spatial relationships, object positions, and key
changes across images. Draw diagrams showing logical connections and
intermediate reasoning steps.
\end{verbatim}

\textbf{Stage 2 - Question Prompt:}

\begin{verbatim}
pre_prompt: "You are given the original images and a visualization
breaking down the reasoning steps. Use both to analyze spatial
relationships, object positions, and changes across images.\n\n"

post_prompt: "\n\nBased on your multi-step reasoning, answer with the
option's letter from the given choices directly. Enclose the option's
letter within ``."
\end{verbatim}

\subsection{IllusionBench Tasks}

\subsubsection{ICON Shape Recognition Visual CoT}

\textbf{Stage 1 - Generation Prompt:}

\begin{verbatim}
Create a visualization that highlights how the background elements
contribute to forming the icon shape. Draw outlines, arrows, and
annotations showing which parts of the scene create the icon's
silhouette or features.
\end{verbatim}

\textbf{Stage 2 - Question Prompt:}

\begin{verbatim}
shape_string = "Animal, Face_Emoji, Music, Sport, Stationery, Vehicle"
scene_string = "Ocean, Origami, Forest, Cloud, Sand_dune,
Medieval_Village, City, Underwater_ruins, Museum, Bazaar_market,
Time_square"

pre_prompt: "You are given the original image and a visualization
highlighting how background elements form the icon. Use both to
identify the icon shape.\n\n"

prompt = f"This image contains an icon integrated into a background,
where elements of the background contribute to forming the icon.
Identify the icon that is represented in the image by choosing
exclusively among the following options:{shape_string},{scene_string}
Provide your response by stating only the single, most accurate class
name that represents the icon. You have to respond with a single word."
\end{verbatim}

\subsubsection{ICON Scene Recognition Visual CoT, LOGO Scene Recognition Visual CoT, and IN Scene Recognition Visual CoT}

\textbf{Note:} These three tasks share identical prompts (only differ in shape\_string values).

\textbf{Stage 1 - Generation Prompt:}

\begin{verbatim}
Create a visualization that highlights the background scene elements.
Draw annotations identifying key features of the environment, landscape,
or setting that characterize the scene type.
\end{verbatim}

\textbf{Stage 2 - Question Prompt:}

\begin{verbatim}
pre_prompt: "You are given the original image and a visualization
highlighting the background scene elements. Use both to identify the
scene type.\n\n"

prompt = f"This image contains an icon integrated into a background,
where elements of the background contribute to forming the icon.
Identify the background that is represented in the image by choosing
exclusively among the following options:{shape_string},{scene_string}.
Provide your response by stating only the single, most accurate class
name that represents the background. You have to respond with a single
word."
\end{verbatim}

\subsubsection{LOGO Shape Recognition Visual CoT}

\textbf{Stage 1 - Generation Prompt:}

\begin{verbatim}
Create a visualization that highlights how the background elements
contribute to forming the logo shape. Draw outlines, arrows, and
annotations showing which parts of the scene create the logo's
silhouette or features.
\end{verbatim}

\textbf{Stage 2 - Question Prompt:}

\begin{verbatim}
shape_string = "Adidas, Amazon, Apple, Audi, BMW, Mercedes Benz,
Facebook, Google, Instagram, Mcdonalds, Nasa, Nike, Olympics,
Playstation, Puma, Reebok, Spotify, Starbucks, Tesla, Telegram, Ubuntu"
scene_string = "Ocean, Origami, Forest, Cloud, Sand_dune,
Medieval_Village, City, Underwater_ruins, Museum, Bazaar_market,
Time_square"

pre_prompt: "You are given the original image and a visualization
highlighting how background elements form the logo. Use both to
identify the logo.\n\n"

prompt = f"This image contains an icon integrated into a background,
where elements of the background contribute to forming the logo.
Identify the logo that is represented in the image by choosing
exclusively among the following options:{shape_string},{scene_string}
Provide your response by stating only the single, most accurate class
name that represents the logo. You have to respond with a single word."
\end{verbatim}

\subsubsection{IN Shape Recognition Visual CoT}

\textbf{Stage 1 - Generation Prompt:}

\begin{verbatim}
Create a visualization that highlights how the background elements
contribute to forming the shape. Draw outlines, arrows, and annotations
showing which parts of the scene create the shape's silhouette or
features.
\end{verbatim}

\textbf{Stage 2 - Question Prompt:}

\begin{verbatim}
shape_string = "Airplane, Bicycle, Bird, Bottle, Car, Cat, Dog,
Dolphin, Fork, Guitar, Mug, Panda, Paper_clip, Sailboat, Scooter,
Teapot"
scene_string = "Ocean, Origami, Forest, Cloud, Sand_dune,
Medieval_Village, City, Underwater_ruins, Museum, Bazaar_market,
Time_square"

pre_prompt: "You are given the original image and a visualization
highlighting how background elements form the shape. Use both to
identify the shape.\n\n"

prompt = f"This image contains an icon integrated into a background,
where elements of the background contribute to forming the icon.
Identify the shape that is represented in the image by choosing
exclusively among the following options:{shape_string},{scene_string}
Provide your response by stating only the single, most accurate class
name that represents the icon. You have to respond with a single word."
\end{verbatim}

\subsection{VisualPuzzles Tasks}

\subsubsection{Algorithmic Visual CoT}

\textbf{Stage 1 - Generation Prompt:}

\begin{verbatim}
Create a visualization that breaks down the algorithmic pattern or
sequence. Draw step-by-step diagrams, number sequences, or flowcharts
showing the logical progression and rules governing the pattern.
\end{verbatim}

\textbf{Stage 2 - Question Prompt:}

\begin{verbatim}
pre_prompt: "You are given the original puzzle image(s) and a
visualization breaking down the algorithmic pattern. Use both to
identify the underlying algorithm or sequence rule.\n\n"

MULTI_CHOICE_DIRECT_PROMPT = "Answer the question with the option's
letter from the given choices directly."

question = "Question: " + doc["question"].strip()
options = doc["options"]
if options is not None:
    question += "\nOptions:\n(A) " + options[0] + "\n(B) " + options[1]
    + "\n(C) " + options[2] + "\n(D) " + options[3]
else:
    question += "\nOptions: Choose from (A) (B) (C) (D) in the image."

question += "\n" + MULTI_CHOICE_DIRECT_PROMPT
\end{verbatim}

\subsubsection{Analogical Visual CoT}

\textbf{Stage 1 - Generation Prompt:}

\begin{verbatim}
You are given an analogical reasoning puzzle (A is to B as C is to ?).

{question}

Your task:
1. Identify the transformation or relationship from A to B
2. Create a diagram that clearly shows:
   - The key features that change from A to B
   - The transformation rule (rotation, color change, shape
     modification, etc.)
   - How the same rule applies to C
   - Annotations or arrows showing the transformation steps
3. Make the transformation pattern visually obvious

Generate a diagram that reveals the transformation pattern between
pairs.
\end{verbatim}

\textbf{Stage 2 - Question Prompt:}

\begin{verbatim}
{question}

You are given TWO images:
1) ORIGINAL PUZZLE: The analogical reasoning puzzle
2) AUXILIARY DIAGRAM: A visualization showing the transformation
   relationship

Use the auxiliary diagram to understand how A transforms to B, apply
the same rule to C, then select the correct answer.
Answer with the option letter (A, B, C, or D) directly.
\end{verbatim}

\subsubsection{Deductive Visual CoT}

\textbf{Stage 1 - Generation Prompt:}

\begin{verbatim}
You are given a deductive reasoning puzzle that requires logical
inference.

{question}

Your task:
1. Identify the given premises, rules, or constraints in the puzzle
2. Create a diagram that clearly shows:
   - All given conditions/rules listed clearly
   - A logical flowchart or inference chain
   - Step-by-step deduction from premises to conclusion
   - Elimination of incorrect possibilities
   - The logical path leading to the answer
3. Use arrows to show the deduction flow

Generate a logical inference diagram that traces the reasoning path.
\end{verbatim}

\textbf{Stage 2 - Question Prompt:}

\begin{verbatim}
{question}

You are given TWO images:
1) ORIGINAL PUZZLE: The deductive reasoning puzzle
2) AUXILIARY DIAGRAM: A logical inference diagram showing the deduction
   steps

Follow the logical chain in the auxiliary diagram to reach the
conclusion, then select the correct answer.
Answer with the option letter (A, B, C, or D) directly.
\end{verbatim}

\subsubsection{Inductive Visual CoT}

\textbf{Stage 1 - Generation Prompt:}

\begin{verbatim}
You are given an inductive reasoning puzzle that requires pattern
recognition.

{question}

Your task:
1. Observe the sequence of examples and identify the underlying pattern
2. Create a diagram that clearly shows:
   - The repeating elements or motifs highlighted/circled
   - The progression rule (what changes from one step to the next)
   - Annotations showing the pattern cycle or growth rule
   - A prediction of what comes next based on the pattern
   - Color-coding or numbering to show pattern repetition
3. Make the inductive pattern visually obvious

Generate a diagram that highlights the repeating pattern and predicts
the next element.
\end{verbatim}

\textbf{Stage 2 - Question Prompt:}

\begin{verbatim}
{question}

You are given TWO images:
1) ORIGINAL PUZZLE: The inductive reasoning puzzle
2) AUXILIARY DIAGRAM: A visualization highlighting the pattern and its
   progression

Use the auxiliary diagram to identify the pattern, then select the
correct answer.
Answer with the option letter (A, B, C, or D) directly.
\end{verbatim}

\subsubsection{Spatial Visual CoT}

\textbf{Stage 1 - Generation Prompt:}

\begin{verbatim}
You are given a spatial reasoning puzzle involving 3D visualization or
transformations.

{question}

Your task:
1. Analyze the spatial transformation required (rotation, folding,
   unfolding, different viewpoint)
2. Create a diagram that clearly shows:
   - The object from multiple angles if rotation is involved
   - Step-by-step folding/unfolding process if applicable
   - Arrows indicating rotation direction and degree
   - Reference points or markers to track orientation
   - The resulting shape after transformation
3. Add axis lines or reference frames to clarify spatial orientation

Generate a multi-view or step-by-step transformation diagram.
\end{verbatim}

\textbf{Stage 2 - Question Prompt:}

\begin{verbatim}
{question}

You are given TWO images:
1) ORIGINAL PUZZLE: The spatial reasoning puzzle
2) AUXILIARY DIAGRAM: A visualization showing the spatial
   transformation from multiple views

Use the auxiliary diagram to mentally trace the transformation, then
select the correct answer.
Answer with the option letter (A, B, C, or D) directly.
\end{verbatim}

\subsection{ChartQA Tasks}

\subsubsection{chartqa100\_visual\_cot}

\textbf{Stage 1 - Generation Prompt:}

\begin{verbatim}
Based on this chart and the question being asked, generate an annotated
version of the chart that highlights the relevant data points, bars,
lines, or regions needed to answer the question. Mark or circle the key
values and label them clearly.
\end{verbatim}

\textbf{Stage 2 - Question Prompt:}

\begin{verbatim}
In addition to the original chart, you are also given an annotated
version that highlights the relevant data points for the question.

{question}

Answer the question with a single word.
\end{verbatim}

\subsection{Geometry Tasks}

\subsubsection{geometry3k\_visual\_cot}

\textbf{Stage 1 - Generation Prompt:}

\begin{verbatim}
Based on this geometry problem, generate an auxiliary diagram that adds
helpful constructions to solve the problem.

Your task:
1. Analyze what geometric relationships need to be proven or calculated
2. Add auxiliary lines, points, or constructions that reveal these
   relationships
3. Label any new points, angles, or segments clearly
4. Highlight key geometric properties (parallel lines, congruent
   segments, equal angles, etc.)

Generate a diagram with auxiliary constructions that make the solution
path clearer.
\end{verbatim}

\textbf{Stage 2 - Question Prompt:}

\begin{verbatim}
You are given TWO images:
1) ORIGINAL DIAGRAM: The geometry problem diagram
2) AUXILIARY DIAGRAM: The same diagram with helpful auxiliary
   constructions added

Use both diagrams to solve the problem. The auxiliary constructions
should help you identify key relationships.

{question}
\end{verbatim}

\subsubsection{auxsolidmath\_easy\_visual\_cot}

\textbf{Stage 1 - Generation Prompt:}

\begin{verbatim}
You are given a solid geometry problem with a 3D diagram. Analyze the
problem and create an enhanced version of the SAME diagram with
auxiliary constructions added.

Problem: {question}

Instructions:
1. KEEP all original elements exactly as they are (all points, edges,
   faces, labels)
2. Analyze what auxiliary constructions would help solve this problem
3. ADD auxiliary lines in a different color (e.g., red or dashed
   lines). Common auxiliary constructions include:
   - Perpendiculars from a point to a plane or line
   - Line segments connecting specific points
   - Midpoints of edges with connections
   - Extended lines to find intersections
   - Parallel lines through specific points
   - Cross-sections of the solid
4. Label any new points you add (use letters not already in the
   diagram)
5. The final diagram should look like the original 3D figure with extra
   auxiliary lines drawn on top

Generate an enhanced 3D diagram that preserves the original and adds
helpful auxiliary constructions.
\end{verbatim}

\textbf{Stage 2 - Question Prompt:}

\begin{verbatim}
You are given a solid geometry problem.

Problem: {question}

You are given TWO images:
1) ORIGINAL DIAGRAM: The 3D solid geometry figure as given
2) AUXILIARY DIAGRAM: The same figure with auxiliary constructions
   (extra lines) added to help solve the problem

Instructions:
1. Look at the auxiliary diagram to see what constructions were added
   (perpendiculars, connecting segments, midpoints, etc.)
2. Identify the geometric relationships established by these auxiliary
   lines
3. Use these constructions to set up your solution approach
4. Apply relevant theorems (Pythagorean theorem in 3D, properties of
   perpendiculars, volume formulas, etc.)
5. Show your step-by-step solution with clear calculations
6. State the final numerical answer

Solve this problem step by step using the auxiliary constructions.
\end{verbatim}

\subsection{Physics Tasks}

\subsubsection{phyx\_optics100\_visual\_cot}

\textbf{Stage 1 - Generation Prompt:}

\begin{verbatim}
Based on this optics problem, draw a light ray diagram that helps solve
the problem. Show the path of light rays through the optical system,
including incident rays, reflected rays, refracted rays, focal points,
and any relevant angles or focal points as needed by the problem.
\end{verbatim}

\textbf{Stage 2 - Question Prompt:}

\begin{verbatim}
In addition to the original image, you are also given an auxiliary
light ray diagram to help you solve the problem.

{question}
\end{verbatim}

\subsubsection{phyx\_mechanics100\_visual\_cot}

\textbf{Stage 1 - Generation Prompt:}

\begin{verbatim}
Based on this mechanics problem, draw a free body diagram (force
analysis diagram) that helps solve the problem. Show all the forces
acting on the object(s), including gravity, normal force, friction,
tension, applied forces, etc., with arrows indicating direction and
relative magnitude.
\end{verbatim}

\textbf{Stage 2 - Question Prompt:}

\begin{verbatim}
In addition to the original image, you are also given an auxiliary free
body diagram (force analysis diagram) to help you solve the problem.

{question}
\end{verbatim}

\subsection{RealUnify GEU Tasks}

\subsubsection{realunify\_mental\_tracking\_visual\_cot}

\textbf{Stage 1 - Generation Prompt:}

\begin{verbatim}
Here is the question: {question}
Please apply the corresponding transformations and modifications to the
contents of the image according to the question.
\end{verbatim}

\textbf{Stage 2 - Question Prompt:}

\begin{verbatim}
In addition to the original image, you are also given a transformed
version of the image with the modifications applied according to the
question.

{prompt}
\end{verbatim}

\subsubsection{realunify\_mental\_reconstruction\_visual\_cot}

\textbf{Stage 1 - Generation Prompt:}

\begin{verbatim}
Please restore the image that has been shuffled by patches, without
adding extra content or altering the original image.
\end{verbatim}

\textbf{Stage 2 - Question Prompt:}

\begin{verbatim}
In addition to the original image, you are also given a restored
version of the shuffled image to help you answer the question.

{prompt}
\end{verbatim}

\subsubsection{realunify\_attentional\_focusing\_visual\_cot}

\textbf{Stage 1 - Generation Prompt:}

\begin{verbatim}
Here is the question: {question}
Please highlight the regions of the image that are relevant to the
question.
\end{verbatim}

\textbf{Stage 2 - Question Prompt:}

\begin{verbatim}
In addition to the original image, you are also given a visualization
that highlights the regions relevant to the question.

{prompt}
\end{verbatim}

\subsection{VSP (Visual-Spatial-Planning) Tasks}

\subsubsection{vsp\_google\_map\_visual\_cot}

\textbf{Stage 1 - Generation Prompt:}

\begin{verbatim}
This image shows a map with a starting point marked 'S' (blue) and a
goal marked 'G' (red). Your task: Generate a clear visualization that
highlights the optimal path from S to G. Draw a simplified map showing
the route with arrows indicating directions (North, South, East, West)
and mark the number of crossroads to pass in each direction.
\end{verbatim}

\textbf{Stage 2 - Question Prompt:}

\begin{verbatim}
In addition to the original images, you are also given an auxiliary
visualization image showing the path analysis.

As a professional pathfinder, your task is to analyze a map and find a
route from the starting location to the goal. Since coding is not
within your skill set, your approach relies on logical reasoning of the
map.

## Game Setup
- The game presents a fully observable map.
- The starting location is marked with blue "S", and the goal is marked
  with red "G".
- Your goal is to find a path from the starting location to the goal.

## Moving Rules
- The action plan involves moves in four directions: 'W' (West), 'E'
  (east), 'N' (north), or 'S' (south).
- Each move is along with distances. Distances are measured by how many
  crossroads passed.
We provide an example to further illustrate the rules.

[Example Image]

In this provided example:
- You are now at the southwest of the goal.
- If you move north by 1 crossroad, you will be at the west of the
  goal.
- If you move east by 4 crossroads, you will be at the goal.
- IMPORTANT: Please ignore the name of the street and avenue. The
  numbers in the name cannot be used to compute how many crossroads
  need to be passed.

## Procedure and Output
Now you will solve the given maze. To analyze the relative spatial
relation between the starting point and the goal (for example,
southwest). Then, output a path using the format <Direction>: <Number
of crossroads passed>.
For example:
<Output>
1. North: 1
2. East: 4
means move north by 1 crossroad, and move east by 4 crossroads.
<Output>
1. South: 1
means move south by 1 crossroad.
Do not output any extra content after the above aggregated output.

Please output path for the following map:

[Test Image]
\end{verbatim}

\subsubsection{vsp\_collision\_visual\_cot}

\textbf{Stage 1 - Generation Prompt:}

\begin{verbatim}
This image shows a map with a car, a person, and a goal. Your task:
Generate a clear visualization that analyzes the distances and paths.
Draw a simplified diagram showing the grid distances from the car to
the goal and from the person to the goal, with arrows indicating their
movement directions.
\end{verbatim}

\textbf{Stage 2 - Question Prompt:}

\begin{verbatim}
In addition to the original images, you are also given an auxiliary
visualization image showing the distance analysis.

As a professional navigation agent, your task is to analyze a map and
determine the time needed for the car and the person passing the goal.

## Game Setup
- The game presents a fully observable map. There is a person, a car,
  and a goal on the map.
- The game further specifies the moving direction of the person and car
  ("up", "down", "left", "right").
- Your goal is to determine the time needed for the car and the person
  passing the goal.
The following figure shows how the player, the car, and the goals look
like.

[Icon Image]

We provide an example to further illustrate the rules.

[Example Image]

The car is moving left with speed 1.0 grid per second, and the person
is moving up with speed 0.5 grid per second.

In this provided example:
- The car is 2 grid away from the goal. Given it's time as 1.0 grid per
  second, the time needed is 2 / 1.0 = 2 seconds.
- The person is 1 grid away from the goal. Given it's time as 0.5 grid
  per second, the time needed is 1 / 0.5 = 2 seconds.

## Procedure and Output
Now you will answer for the following given map. To solve it, analyze
the car and the person separately. Then, answer for them separately.
For example:
Car: 2.0
Person: 2.0
means car and the person will need 2.0 seconds to pass the goal
respectively.
Do not output any extra content after the above aggregated output.

Please analyze and determine the time needed for the car and the person
passing the goal:

[Test Image]

The car is moving {car_dir} with speed {car_speed}, and the person is
moving {person_dir} with speed {person_speed}.
\end{verbatim}

\subsection{BabyVision Tasks}

\subsubsection{babyvision\_fine\_grained\_visual\_cot}

\textbf{Stage 1 - Generation Prompt:}

\begin{verbatim}
This image contains subtle visual details that need careful
examination. Your task: Generate a visualization that highlights and
emphasizes the fine-grained details, subtle differences, and
distinguishing features in the image. Make the key discriminative
elements more prominent and easier to identify.
\end{verbatim}

\textbf{Stage 2 - Question Prompt:}

\begin{verbatim}
In addition to the original image, you are also given an auxiliary
visualization image that highlights key visual elements.

{question}

Think step by step and give your final answer in \boxed{Answer} format.
\end{verbatim}

\subsubsection{babyvision\_visual\_tracking\_visual\_cot}

\textbf{Stage 1 - Generation Prompt:}

\begin{verbatim}
This image involves tracking objects or their movements. Your task:
Generate a visualization that highlights the objects of interest, their
positions, trajectories, or movement patterns. Make it easier to follow
and track the relevant visual elements.
\end{verbatim}

\textbf{Stage 2 - Question Prompt:}

\begin{verbatim}
In addition to the original image, you are also given an auxiliary
visualization image that highlights key visual elements.

{question}

Think step by step and give your final answer in \boxed{Answer} format.
\end{verbatim}

\subsection{Uni-MMMU Tasks}

\subsubsection{uni\_mmmu\_jigsaw100\_visual\_cot}

\textbf{Full Prompt:}

\begin{verbatim}
You are a unified vision-language model. You will be given:
(1) a 2x2 reference image with the bottom-right cell hidden, and
(2) two candidate patch images ("Candidate 0" and "Candidate 1").

Your job:
- For each candidate, synthesize a completed 2x2 image by placing that
  candidate EXACTLY into the bottom-right cell. Keep the other three
  cells pixel-identical to the reference (no filtering, no
  re-rendering). If sizes differ, only scale the candidate to fit that
  quadrant; do NOT rotate, mirror, or alter colors.
- Compare the two completed results and decide which candidate yields
  the correct completion.

Output EXACTLY the following, in order:

1) A single image with Candidate 0 placed in the bottom-right cell

2) A single image with Candidate 1 placed in the bottom-right cell


3) analysis comparing seam continuity, color/texture gradient,
   structural alignment, and global semantics

4) One strict JSON object with your decision, wrapped as:
<FINAL_ANSWER_JSON>
{"choice": 0 or 1, "rationale": "<=30 words decisive cue"}
</FINAL_ANSWER_JSON>

Hard constraints:
- Deterministic, single outcome. No hedging, no multiple possibilities.
- No meta talk about prompts, models, or pipelines.
- Do not restate the task as the answer; reason from visual evidence.
- The only edits allowed are pasting the candidate into the bottom-
  right cell and necessary size matching for that cell. All other
  pixels must remain identical to the reference.

Inputs:
\end{verbatim}

\subsubsection{uni\_mmmu\_maze100\_visual\_cot}

\textbf{Full Prompt:}

\begin{verbatim}
You are a precise maze solver.

SEMANTICS (for all mazes)
- Black squares: walls (impassable)
- White squares: path (walkable)
- Blue dot: start (the agent)
- Green rectangular frame: goal (reaching any white cell inside the
  green frame counts as success)
- Legal moves: up, down, left, right only. One cell per step; no
  diagonals, no jumps; never cross walls.

OUTPUT FORMAT (STRICT)
1) MULTI-IMAGE MODE - generate a SEQUENCE OF SEPARATE IMAGES, one per
   move:
   - Each output image must depict the maze state AFTER applying
     exactly one legal move.
   - Do NOT include the initial (pre-move) state.
   - Keep palette/layout/scale identical to the input; only the blue
     dot moves.
   - The number of returned images MUST equal the number of moves in
     the final answer (see step 2).
   - Absolutely FORBIDDEN: any collage/montage/spritesheet/grid/
     multi-panel/side-by-side/stacked images; no arrows, captions, or
     overlays; no GIFs/animations/video.

2) After all step images, emit EXACTLY ONE LINE containing ONLY the
   final move list as a JSON array of lowercase strings, wrapped as:
   <ANSWER_JSON>["right","down","left"]</ANSWER_JSON>


NO EXTRAS
- No tools, no OCR, no explanations, and no text other than the single
  <ANSWER_JSON>.....</ANSWER_JSON> line.
- Do not restate the instructions or the condition.

REMINDERS
- Decide the full path first, then emit the image sequence (one image
  per move), then the single <ANSWER_JSON> line.
- One move per image; images must be separate files/parts, not stitched
  together in any way.
\end{verbatim}

\subsubsection{uni\_mmmu\_sliding54\_visual\_cot}

\textbf{Full Prompt:}

\begin{verbatim}
You are a precise sliding puzzle solver.

SEMANTICS
- The puzzle is a 3x3 grid with 8 colored tiles and one empty space
  (shown as red).
- A "move" slides an adjacent tile INTO the empty space.
- Moves are named by the direction the COLORED TILE moves (not the
  empty space).
- Legal moves: up, down, left, right only. One tile per step.

OUTPUT FORMAT (STRICT)
1) MULTI-IMAGE MODE - generate a SEQUENCE OF SEPARATE IMAGES, one per
   move:
   - Each output image must depict the puzzle state AFTER applying
     exactly one move.
   - Do NOT include the initial (pre-move) state.
   - Keep tile colors identical; only positions change.
   - The number of returned images MUST equal the number of moves in
     the final answer.
   - Absolutely FORBIDDEN: any collage/montage/spritesheet/grid/
     multi-panel/side-by-side/stacked images.

2) After all step images, emit EXACTLY ONE LINE containing ONLY the
   final move list as a JSON array of lowercase strings, wrapped as:
   <ANSWER_JSON>["down","right","up"]</ANSWER_JSON>

NO EXTRAS
- No explanations beyond the single <ANSWER_JSON>.....</ANSWER_JSON>
  line.
- Do not restate the instructions.
\end{verbatim}


\section{RA and AL Evaluation Pipeline and Full Prompt Specification}
\label{app:ra-al}

\subsection{Purpose}
This appendix documents the complete evaluation protocol used for Visual Chain-of-Thought (Visual CoT) quality assessment with two judge metrics:
\begin{itemize}
\item \textbf{RA} (Reasoning-to-Visual Alignment)
\item \textbf{AL} (Answer-to-Visual Alignment)
\end{itemize}
The implementation is fully prompt-based and uses a multimodal LLM judge.

\subsection{Per-sample Inputs and Metadata}
For each sample \(i\), the evaluator reads:
\begin{itemize}
\item Original image(s): \(I_{\text{orig}}^{(i)}\)
\item Generated auxiliary image(s): \(I_{\text{aux}}^{(i)}\)
\item Generation prompt: \(p_{\text{gen}}^{(i)}\)
\item Original question: \(q^{(i)}\)
\item Final Stage-2 answer: \(a^{(i)}\)
\item Task name: \(t^{(i)}\)
\end{itemize}

The metadata file contains:
\begin{verbatim}
{
  "doc_id": ...,
  "task": ...,
  "generation_prompt": ...,
  "generated_images": [...],
  "question": ...,
  "stage2_answer": ...
}
\end{verbatim}

\subsection{Metric Definitions}
\paragraph{RA.}
\[
\mathrm{RA}_i = \mathcal{J}_{\mathrm{RA}}\!\left(I_{\text{orig}}^{(i)}, I_{\text{aux}}^{(i)},
p_{\text{gen}}^{(i)}, q^{(i)}, t^{(i)}\right)\in\{1,2,3,4,5\}
\]
RA evaluates whether generated auxiliary visualization follows the generation instruction and is useful/relevant.

\paragraph{AL.}
\[
\mathrm{AL}_i = \mathcal{J}_{\mathrm{AL}}\!\left(I_{\text{orig}}^{(i)}, I_{\text{aux}}^{(i)},
q^{(i)}, a^{(i)}, t^{(i)}\right)\in\{1,2,3,4,5\}
\]
AL evaluates whether the final answer is consistent with visual evidence and question intent.

\paragraph{Dataset-level averages.}
For \(N\) samples:
\[
\overline{\mathrm{RA}}=\frac{1}{N}\sum_{i=1}^N \mathrm{RA}_i,\qquad
\overline{\mathrm{AL}}=\frac{1}{N}\sum_{i=1}^N \mathrm{AL}_i.
\]

\subsection{End-to-end Evaluation Flow}
For each sample:
\begin{enumerate}
\item Load metadata JSON.
\item Resolve \texttt{generated\_images} (relative paths are optionally resolved via \texttt{base\_dir}).
\item Construct RA prompt: \texttt{Base RA Prompt + Category-specific RA Addition}.
\item Call multimodal judge with images \([I_{\text{orig}}, I_{\text{aux}}]\), parse \texttt{reasoning\_visual\_score}.
\item Construct AL prompt: \texttt{Base AL Prompt + Category-specific AL Addition}.
\item Call multimodal judge with images \([I_{\text{orig}}, I_{\text{aux}}]\), parse \texttt{answer\_visual\_score}.
\item Save output fields:
\texttt{ra\_score}, \texttt{ra\_reasoning}, \texttt{al\_score}, \texttt{al\_reasoning}.
\end{enumerate}

\subsection{Judge Call Settings}
\begin{itemize}
\item Multimodal judge: GPT-4o compatible endpoint (Azure OpenAI interface).
\item Temperature: \(0.0\).
\item Max response tokens: \(3000\).
\item Retry strategy: metric-level retries and API-level retries with backoff.
\item Image encoding: JPEG base64 payload.
\end{itemize}

\subsection{Prompt Composition Rule}
Prompt construction uses this priority:
\begin{enumerate}
\item If \texttt{task\_category} is explicitly set, append that category's criteria.
\item Else, apply legacy task-name mapping:
\begin{itemize}
\item \texttt{illusionbench} \(\rightarrow\) \texttt{perception}
\item \texttt{chartqa} \(\rightarrow\) \texttt{chart\_table}
\item \texttt{mathvista} \(\rightarrow\) \texttt{mathematical}
\end{itemize}
\item Else, use base prompt only.
\end{enumerate}

\subsection{Complete Base Prompt: RA}
\begin{verbatim}
You are a professional AI evaluation specialist for Visual Chain-of-Thought tasks.

## Task Context
- **Task**: {task}
- **Original Question**: {question}

## Your Objective
Evaluate if the **generated visualization images** correctly follow the
**generation prompt**.

## Images Provided
- **Image 1**: Original question image (the input)
- **Image 2+**: Generated visualization image(s) (the auxiliary output created
  to help answer the question)

## Generation Prompt (What model was instructed to generate)
"{generation_prompt}"

## Evaluation Criteria

### 1. Instruction Following (40%)
- Does the generated image follow the generation prompt instructions?
- Are all requested visual elements present?
- Does it match the specified visualization type?

### 2. Visual Quality (30%)
- Is the generated image clear and well-formed?
- Are visual elements properly rendered?
- Is it helpful as an auxiliary visualization?

### 3. Relevance to Task (30%)
- Does the visualization help answer the original question?
- Is it relevant to the task context?
- Does it provide useful visual cues?

## Scoring Scale (1-5)

- **5 Perfect**: Generated image(s) perfectly match generation prompt;
  all instructions followed; high quality; highly relevant
- **4 Good**: Generated image(s) mostly match prompt (80-90%);
  minor issues; good quality; relevant
- **3 Adequate**: Generated image(s) partially match prompt (60-70%);
  noticeable issues; acceptable quality
- **2 Poor**: Generated image(s) barely match prompt (30-50%);
  major issues; poor quality or relevance
- **1 Failed**: Generated image(s) completely fail to match prompt;
  unusable; irrelevant

## Important Notes
- Focus on whether the **generation prompt** was correctly executed
- Do NOT evaluate if the final answer is correct (that's a separate metric)
- Evaluate the **visualization quality** and **instruction adherence**

## Output Format
{
  "reasoning_visual_score": X,
  "reasoning": "Detailed evaluation of how well generated images match the
  generation prompt"
}
\end{verbatim}

\subsection{Complete Base Prompt: AL}
\begin{verbatim}
You are a professional AI evaluation specialist for Visual Chain-of-Thought tasks.

## Task Context
- **Task**: {task}
- **Original Question**: {question}

## Your Objective
Evaluate if the **final answer** is consistent with the **generated visualization
images** and the **original question**.

## Images Provided
- **Image 1**: Original question image (the input)
- **Image 2+**: Generated visualization image(s) (auxiliary output used to
  derive the answer)

## Final Answer
"{answer}"

## Evaluation Criteria

### 1. Visual-Answer Consistency (50%)
- Is the answer supported by visual evidence in generated images?
- Can you trace the answer back to visual elements?
- Are there contradictions between answer and visuals?

### 2. Question-Answer Alignment (30%)
- Does the answer correctly address the original question?
- Is the answer format appropriate?
- Is it a valid response to the question?

### 3. Reasoning Coherence (20%)
- Does the reasoning flow make sense (original → generated visual → answer)?
- Is the answer logically derived from the visualization?
- Is the overall chain-of-thought coherent?

## Scoring Scale (1-5)

- **5 Perfect**: Answer perfectly matches generated visuals and question;
  fully consistent; logically sound; correct reasoning
- **4 Good**: Answer mostly consistent (80-90%);
  minor inconsistencies; generally correct reasoning
- **3 Adequate**: Answer partially consistent (60-70%);
  some contradictions; reasoning has gaps
- **2 Poor**: Answer barely consistent (30-50%);
  major contradictions; flawed reasoning
- **1 Failed**: Answer completely inconsistent with visuals or question;
  contradictory; nonsensical

## Important Notes
- Focus on **consistency** between answer and generated visuals
- Check if answer logically follows from the visualization
- Verify answer addresses the original question
- Do NOT evaluate if generation prompt was followed (that's RA metric)

## Output Format
{
  "answer_visual_score": X,
  "reasoning": "Detailed evaluation of answer consistency with visuals and question"
}
\end{verbatim}

\subsection{Full Category-specific Additions}

\subsubsection*{A. Real-world Applications (\texttt{real\_world})}
\paragraph{RA addition}
\begin{verbatim}
## Real-world Applications - RA Criteria
- **Safety & Feasibility**: Does visualization show safe, practical steps?
- **Contextual Understanding**: Does it consider real-world constraints
  (time, resources, safety)?
- **Actionable Guidance**: Are visual steps clear enough to follow in practice?
- **Ethical Considerations**: Does visualization respect safety protocols and
  ethical guidelines?
\end{verbatim}
\paragraph{AL addition}
\begin{verbatim}
## Real-world Applications - AL Criteria
- **Practical Validity**: Is the answer practically feasible and safe?
- **Evidence-Based**: Is answer supported by visual evidence showing real-world execution?
- **Completeness**: Does answer address all critical real-world aspects
  (safety, resources)?
- **Ethical Soundness**: Is the answer ethically appropriate and safe?
\end{verbatim}

\subsubsection*{B. Geometry Reasoning (\texttt{mathematical})}
\paragraph{RA addition}
\begin{verbatim}
## Mathematical Reasoning - RA Criteria
- **Formula Visualization**: Are mathematical formulas and equations
  properly visualized?
- **Step-by-Step Clarity**: Does visualization show clear calculation steps?
- **Numerical Accuracy**: Are all numerical values and calculations visually correct?
- **Mathematical Notation**: Is proper mathematical notation used?
\end{verbatim}
\paragraph{AL addition}
\begin{verbatim}
## Mathematical Reasoning - AL Criteria
- **Calculation Correctness**: Is the answer mathematically accurate?
- **Formula Application**: Does answer correctly apply formulas shown in visualization?
- **Logical Consistency**: Is the mathematical logic coherent from visual → answer?
- **Unit Consistency**: Are units and scales properly maintained?
\end{verbatim}

\subsubsection*{C. Physics (\texttt{stem})}
\paragraph{RA addition}
\begin{verbatim}
## Physics - RA Criteria
- **Scientific Accuracy**: Does visualization follow scientific principles and laws?
- **Technical Precision**: Are technical details and processes accurately depicted?
- **Domain Knowledge**: Does it demonstrate correct Physics domain knowledge?
- **Experimental/Process Visualization**: Are processes or experiments
  properly illustrated?
\end{verbatim}
\paragraph{AL addition}
\begin{verbatim}
## Physics - AL Criteria
- **Scientific Validity**: Is answer scientifically correct?
- **Technical Accuracy**: Does answer align with technical/scientific principles shown?
- **Evidence-Based**: Is answer supported by scientific evidence in visualization?
- **Domain Consistency**: Does answer demonstrate proper Physics domain understanding?
\end{verbatim}

\subsubsection*{D. Puzzles and Games (\texttt{puzzles})}
\paragraph{RA addition}
\begin{verbatim}
## Puzzles and Games - RA Criteria
- **Rule Adherence**: Does visualization follow game/puzzle rules correctly?
- **Strategic Clarity**: Are strategic steps or moves clearly visualized?
- **Valid Moves**: Are all shown moves/steps valid according to rules?
- **Solution Path**: Does visualization show a valid path to solution?
\end{verbatim}
\paragraph{AL addition}
\begin{verbatim}
## Puzzles and Games - AL Criteria
- **Rule Compliance**: Does answer follow all game/puzzle rules?
- **Solution Validity**: Is the answer a valid solution to the puzzle?
- **Move Correctness**: Do final moves/steps match what's shown in visualization?
- **Win Condition**: Does answer achieve the stated goal/win condition?
\end{verbatim}

\subsubsection*{E. Chart \& Table Reasoning (\texttt{chart\_table})}
\paragraph{RA addition}
\begin{verbatim}
## Chart & Table Reasoning - RA Criteria
- **Data Highlighting**: Are relevant data points properly highlighted?
- **Trend Visualization**: Are trends, patterns, or key values emphasized?
- **Label Clarity**: Are axes, labels, and legends clearly marked?
- **Data Accuracy**: Is extracted/highlighted data numerically accurate?
\end{verbatim}
\paragraph{AL addition}
\begin{verbatim}
## Chart & Table Reasoning - AL Criteria
- **Data Accuracy**: Does answer match actual data values in chart/table?
- **Trend Correctness**: If describing trends, is interpretation correct?
- **Numerical Precision**: Are numerical answers precise and accurate?
- **No Hallucination**: Is answer derived from actual data, not hallucinated?
\end{verbatim}

\subsubsection*{F. Spatial Intelligence (\texttt{spatial})}
\paragraph{RA addition}
\begin{verbatim}
## Spatial Intelligence - RA Criteria
- **3D Understanding**: Does visualization correctly represent 3D spatial relationships?
- **Transformation Accuracy**: Are rotations, translations, or transformations correct?
- **Perspective Consistency**: Is spatial perspective maintained correctly?
- **Geometric Precision**: Are geometric relationships accurately visualized?
\end{verbatim}
\paragraph{AL addition}
\begin{verbatim}
## Spatial Intelligence - AL Criteria
- **Spatial Correctness**: Is answer spatially accurate?
- **Transformation Validity**: Do described transformations match visualization?
- **Geometric Consistency**: Is answer geometrically consistent with shown
  spatial relationships?
- **Orientation Accuracy**: Are orientations and positions correctly described?
\end{verbatim}

\subsubsection*{G. Perception Reasoning (\texttt{perception})}
\paragraph{RA addition}
\begin{verbatim}
## Perception Reasoning - RA Criteria
- **Object Highlighting**: Are relevant objects properly highlighted or marked?
- **Scene Understanding**: Does visualization show correct scene interpretation?
- **Feature Emphasis**: Are key perceptual features emphasized?
- **Visual Clarity**: Is the visualization clear for perception tasks?
\end{verbatim}
\paragraph{AL addition}
\begin{verbatim}
## Perception Reasoning - AL Criteria
- **Object Identification**: Are identified objects correct?
- **Scene Accuracy**: Does answer accurately describe the scene?
- **Attribute Correctness**: Are object attributes (color, shape, position) correct?
- **Visual Evidence**: Is answer supported by visible evidence in images?
\end{verbatim}

\subsection{Scoring Outputs and Schema}
Per-sample output:
\begin{verbatim}
{
  "doc_id": "...",
  "task": "...",
  "json_path": "...",
  "ra_score": 1..5 or null,
  "ra_reasoning": "...",
  "al_score": 1..5 or null,
  "al_reasoning": "..."
}
\end{verbatim}

\subsection{Notes on Robustness}
\begin{itemize}
\item If JSON parsing of judge response fails, regex extraction is used as fallback.
\item If all retries fail, score is returned as \texttt{null}.
\item Multi-image tasks are supported by concatenating all original images before auxiliary images.
\end{itemize}

\end{document}